\documentclass[11pt]{article}

\PassOptionsToPackage{table}{xcolor}
\usepackage[final]{acl}

\usepackage{times}
\usepackage{latexsym}

\usepackage[T1]{fontenc}
\usepackage[utf8]{inputenc}

\usepackage{microtype}
\usepackage{inconsolata}
\usepackage{upquote}

\usepackage{graphicx}
\input{supp-pdf.mkii}
\usepackage{epstopdf}
\usepackage{booktabs}
\usepackage{tabularx}
\usepackage{multirow}
\usepackage{tipa}
\usepackage[table]{xcolor}
\usepackage{adjustbox}
\usepackage{ragged2e}
\usepackage{array}
\usepackage{caption}

\usepackage{comment}
\usepackage{cuted}
\usepackage[most]{tcolorbox}
\usepackage{soul}
\usepackage{tipa}

\usepackage{enumitem}

\usepackage{arabtex}
\usepackage{utf8}
\setcode{utf8}

\newcommand{\cardamom}{\textsc{Cardamom}}

\newcommand{\chip}[2]{%
  \begingroup
  \setlength{\fboxsep}{1.1pt}%
  \colorbox{#1!16}{\raisebox{0pt}[1.6ex][0.6ex]{\textsf{\scriptsize #2}}}%
  \endgroup
}

\newcommand{\EGY}{\chip{orange!85!black}{EGY}}
\newcommand{\KSA}{\chip{green!70!black}{KSA}}
\newcommand{\PAL}{\chip{teal!80!black}{PAL}}
\newcommand{\JOR}{\chip{olive!70!black}{JOR}}
\newcommand{\LEB}{\chip{red!75!black}{LEB}}
\newcommand{\MAU}{\chip{cyan!75!black}{MAU}}

\newcommand{\Egypt}{\colorbox{orange!20}{Egypt}}
\newcommand{\SaudiArabia}{\colorbox{green!15}{Saudi Arabia}}
\newcommand{\Palestine}{\colorbox{teal!20}{Palestine}}
\newcommand{\Jordan}{\colorbox{olive!30}{Jordan}}
\newcommand{\Lebanon}{\colorbox{red!15}{Lebanon}}
\newcommand{\Mauritania}{\colorbox{cyan!7}{Mauritania}}

\title{\texorpdfstring{\raisebox{-0.29\height}{\includegraphics[scale=0.12]{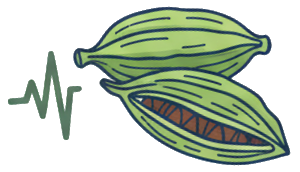}} }{}\cardamom{}: A Micro-Dialectal Arabic Speech Dataset for ASR}

\author{
  \parbox{\linewidth}{\centering \normalfont
    Bashar Talafha\textsuperscript{1},
    Samar M. Magdy\textsuperscript{1},
    Aisha Alansari\textsuperscript{6},
    Alaa Alkhawaldeh\textsuperscript{16},
    Abdurrahman Juma\textsuperscript{3},\\
    Sharaf Makahleh\textsuperscript{11},
    Nour Gamal\textsuperscript{12},
    Omar Attia\textsuperscript{12},
    Hanaa Kurdi\textsuperscript{7},
    Najwa Rizk\textsuperscript{12},\\
    Maysa Anaya,
    Hessah Altimyat\textsuperscript{10},
    Layal Alhazmi\textsuperscript{7},
    Shumukh Alotaibi\textsuperscript{7},
    Hajar Alhadaris\textsuperscript{11},\\
    Rayan Alomari\textsuperscript{11},
    Rahaf Almalaq\textsuperscript{7},
    Malak Alkhorasani\textsuperscript{9},
    Sara alghamdi\textsuperscript{7},
    Rahaf Alshamrani\textsuperscript{9},\\
    Nsrin Ashraf\textsuperscript{13},
    Ibrahim Jaradat\textsuperscript{11},
    Nada Qardahji\textsuperscript{11},
    Yasmin Zaraket\textsuperscript{14},
    Elmoukhtar Brahim\textsuperscript{8},\\
    Sidi Ebeidy\textsuperscript{8},
    Oumoulmouminin Mahmoud\textsuperscript{8},
    Yahjeb Bouha Khatraty\textsuperscript{8},
    Meya Haroune\textsuperscript{8},
    Mohammad Ghaddar\textsuperscript{15},
    Mohamad Eldirany\textsuperscript{15},
    Rashed Alamoush\textsuperscript{11},
    Tala Chhaytle\textsuperscript{4},
    Nuha Albadi\textsuperscript{6},
    Yahya El Hadj\textsuperscript{8},
    Hamzah Luqman\textsuperscript{6},
    Fadi A. Zaraket\textsuperscript{4,5},
    Mustafa Jarrar\textsuperscript{2,3},\\
    Muhammad Abdul-Mageed\textsuperscript{1, $\alpha$}
  }
  \vspace{0.7em} \\
  \parbox{\linewidth}{\centering \normalfont \small
    \textsuperscript{1}The University of British Columbia,
    \textsuperscript{$\alpha$}Canada Research Chair in NLP and ML,
    \textsuperscript{2}Hamad Bin Khalifa University,
    \textsuperscript{3}Birzeit University,
    \textsuperscript{4}American University of Beirut, 
    \textsuperscript{5}Arab Center for Research and Policy Studies, 
    \textsuperscript{6}King Fahd University of Petroleum and Minerals,
    \textsuperscript{7}Taibah University,
    \textsuperscript{8}Institut Supérieur du Numérique,
    \textsuperscript{9}Imam Abdulrahman Bin Faisal University,
    \textsuperscript{10}Northern Border University,
    \textsuperscript{11}Jordan University of Science and Technology,
    \textsuperscript{12}Badr University in Cairo,
    \textsuperscript{13}El Sewedy University of Technology
    \textsuperscript{14}Imperial College London,
    \textsuperscript{15}Lebanese University,
    \textsuperscript{16}Al al-Bayt University
  }\\
  \small \texttt{\{btalafha@mail.ubc.ca, muhammad.mageed@ubc.ca}\}
}

\usepackage{xspace}
\newcommand{\Ar}[1]{{\scriptsize \<#1>\xspace}}

\makeatletter
\def\@fileswithoptions#1{%
  \@ifnextchar[
    {\@fileswith@ptions#1}%
    {\@fileswith@ptions#1[]}%
}
\makeatother

\providecommand{\DIFdel}[1]{}

\providecommand{\DIFdelFL}[1]{}

\begin{document}
\maketitle

\nolinenumbers
\begin{strip}
\vspace{4.75em}
\centering

\includegraphics[width=0.9\textwidth]{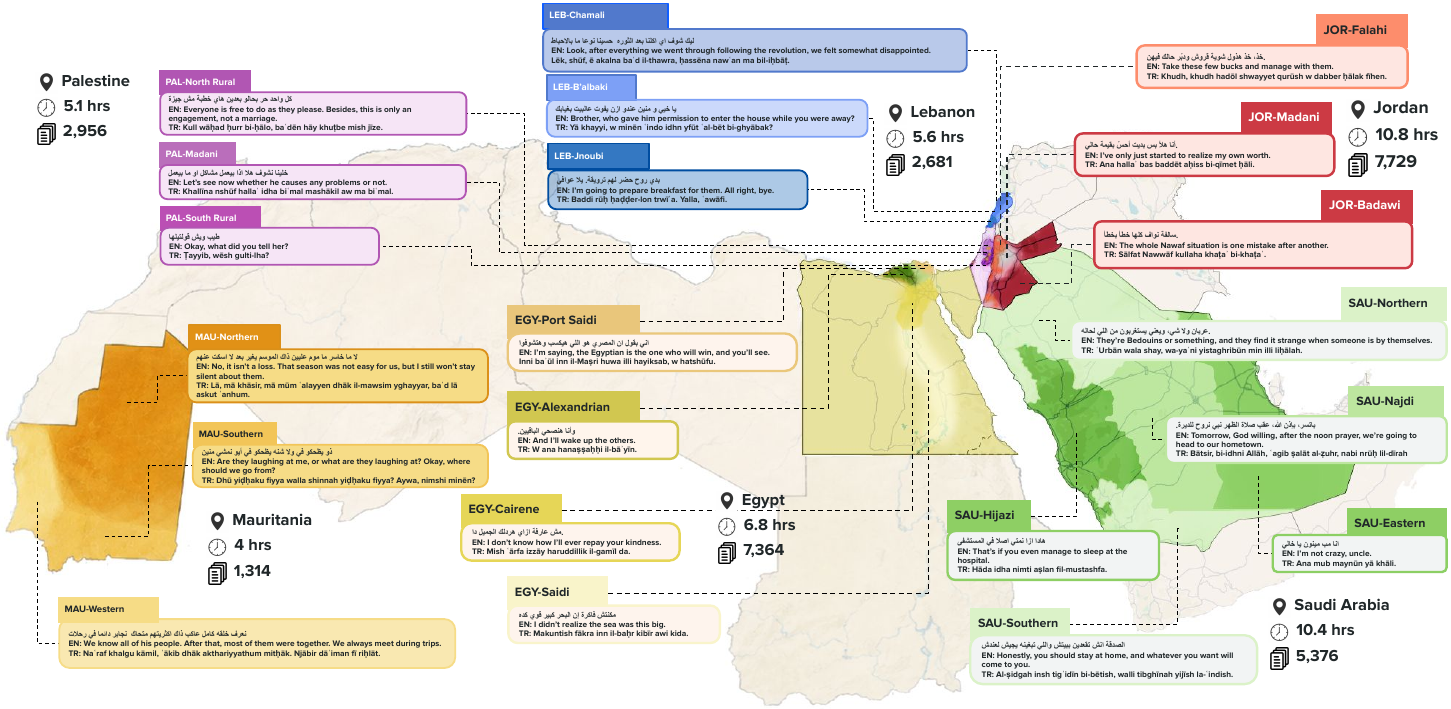}

\captionof{figure}{
Geographic coverage of \cardamom{} across six countries: Egypt, Jordan, Lebanon, Mauritania, Palestine, and Saudi Arabia. The corpus covers 21 micro-dialects spanning urban, rural, and regional varieties. 
}
\label{fig:geographic-coverage}

\vspace{-0.75em}
\end{strip}
\linenumbers

\begin{abstract}
We present \cardamom{}, a micro-dialectal Arabic speech dataset designed to support fine-grained evaluation and adaptation of automatic speech recognition (ASR) systems. Community-curated by native speakers familiar with the represented varieties, \cardamom{} contains approximately 40 hours of transcribed YouTube speech spanning 21 micro-dialects across Egypt, Jordan, Lebanon, Mauritania, Palestine, and Saudi Arabia. Each segment is annotated with one or more operational micro-dialect labels, code-switching information, and utterance-level perceived gender, enabling analysis of sub-country variation that is obscured by conventional country-level labels. We describe the collection and annotation process, motivate the micro-dialect inventory linguistically, and benchmark four multilingual ASR systems in zero-shot and adapted settings. The strongest zero-shot system obtains 43.47\% aggregate WER, with particularly high error rates on Mauritanian and Lebanese varieties; adaptation on \cardamom{} reduces its WER to 35.21\%. Audio-based identification experiments further highlight the difficulty of fine-grained micro-dialect recognition: across three training seeds, a dedicated classifier obtains 46.85\% accuracy and 31.95\% F1. \cardamom{} provides a resource for studying localized dialectal variation and developing Arabic speech systems with broader regional coverage.\footnote{\cardamom{} anonymized repository: \url{https://github.com/cardamom-paper/Cardamom-data}.}

\end{abstract}

\section{Introduction}
Arabic speech technologies have advanced rapidly, yet performance remains uneven across Modern Standard Arabic (MSA) and the regional dialects that dominate everyday communication~\cite{sullivan2026arab,alqadasi2025arabic,alsayadi2022deep}. Although MSA is widely used in education, official communication, and formal media, spoken interaction across the Arab world is primarily dialectal. These dialects encode historical, cultural, social, lexical, phonetic, and prosodic patterns, creating substantial variability for automatic speech recognition (ASR) and related speech processing tasks~\cite{talafha2025context,djanibekov-etal-2025-dialectal}. 

Recent multidialectal corpora, such as Casablanca~\cite{talafha2024casablanca} and SADA~\cite{alharbi2024sada}, have expanded coverage across major varieties. However, most existing resources still represent dialects at relatively coarse regional or national levels, leaving fine-grained sub-country variation under-represented. In this work, we employ the term \textit{micro-dialect} to refer to a localized sub-country variety associated with specific speech communities and reflected in systematic phonetic, lexical, or prosodic patterns. This granularity is important since intra-country variation can be substantial and hence country-level labels may obscure speech continua shaped by geography, contact, migration, and local identity~\cite{AlWer2002b}. 
This gap motivates the development of Arabic speech resources that move beyond broad dialectal labels and make localized variation observable. Such resources are essential for testing whether speech models generalize across nearby but distinct communities, identifying varieties that remain underserved, and developing ASR systems that are robust to Arabic as it is actually spoken. 

We address this need with \cardamom{}, a community-driven Arabic speech corpus designed for fine-grained analysis of micro-dialectal variation. The corpus spans localized varieties in \textit{Egypt}, \textit{Jordan}, \textit{Lebanon}, \textit{Mauritania}, \textit{Palestine}, \textit{Saudi Arabia}. It is curated and annotated by native speakers of the relevant local varieties. The data are drawn from publicly available YouTube videos and transcribed to support ASR evaluation in real-world speech conditions. Overall, \cardamom{} contains $\sim$40 hours of transcribed speech across 21 micro-dialects, together with utterance-level perceived gender. This design enables evaluation of ASR and dialect identification at both country and micro-dialect levels, revealing which localized Arabic varieties current models still serve poorly.

The contributions of this study are: (i) \textit{We introduce \cardamom{}, a micro-level multidialectal Arabic speech dataset} containing 40 hours of transcribed speech from localized varieties across six countries. (ii) \textit{We provide high-granularity micro-dialect annotations} covering 21 local varieties, enabling evaluation beyond coarse country-level dialect categories. (iii) \textit{We benchmark ASR and micro-dialect identification, including adaptation and a source-video-disjoint identification split, and report inter-annotator agreement on a reannotated subset.}.

\section{Related Work}
\textbf{Arabic Speech and ASR Resources.}
Arabic ASR has benefited from large broadcast and web-derived corpora, including NIST EARS \cite{pallett2003look}, DARPA GALE \cite{soltau2009advances}, and the MGB Challenge series \cite{ali2016mgb,ali2017speech,ali2019mgb}, which supported multi-dialect broadcast ASR, Egyptian Arabic YouTube speech, and dialect identification across Arabic-speaking countries. Later resources such as QASR \cite{mubarak2021qasr}, MASC \cite{10022652}, Aswat \cite{alkanhal-etal-2023-aswat}, and SADA \cite{alharbi2024sada} further expanded Arabic speech data.


\textbf{Dialectal, Multidialectal, and Fine-Grained Arabic ASR.}
Recent resources and shared tasks have expanded dialectal Arabic speech processing, including Casablanca \cite{talafha2024casablanca}, NADI 2025 \cite{talafha2025context}, ADI-20 \cite{elleuch2025adi}, and focused low-resource corpora such as TuniFra \cite{choux-etal-2025-tunifra}. At the same time, work on text-based Arabic dialect identification has shown that variation can be modeled below the country level~\cite{abdul-mageed-etal-2018-tweet,abdul-mageed-etal-2020-toward}. However, spoken resources remain mostly organized around countries or broad regional varieties \cite{ali2019mgb,talafha2024casablanca,talafha2025context,elleuch2025adi}, leaving fine-grained micro-dialectal supervision underrepresented in Arabic ASR.



\textbf{Code-Switching, Orthography, and Transcription Challenges.}
Arabic ASR technology also need to handle code-switching, mixed orthographies, and non-standardized dialectal spelling. Mixat \cite{al-ali-aldarmaki-2024-mixat} and ZAEBUC-Spoken \cite{hamed-etal-2024-zaebuc} address bilingual Arabic-English speech, Casablanca includes code-switching annotations in a multidialectal ASR setting \cite{talafha2024casablanca}, and CAFE targets spontaneous code-switching in North African Arabic \cite{lachemat2025cafe}. PolyWER \cite{kadaoui-etal-2024-polywer} and Arabic ASR diacritization work \cite{aldarmaki2023diacritic} further show that transcription and evaluation choices are critical under script, transliteration, and dialectal variation.

\textbf{Modeling and Benchmarking.}
Large multilingual models such as Whisper \cite{radford2023robust}, MMS \cite{pratap2024scaling}, and SeamlessM4T \cite{seamless2025joint} are common ASR baselines, but Arabic-specific work shows that dialectal variation, unseen dialects, code-switching, and data imbalance remain challenging \cite{talafha2023n,djanibekov-etal-2025-dialectal}. Recent work addresses these issues through Whisper adaptation for multidialectal Arabic ASR \cite{ozyilmaz2025overcoming}, Arabic-focused models such as VoxArabica \cite{waheed2023voxarabica}, Habibi \cite{chen2026habibi}, ArTST \cite{toyin-etal-2023-artst}, and shared evaluation through the Open Universal Arabic ASR Leaderboard \cite{wang2024open}.

\cardamom{} complements these resources with micro-dialect labels; Appendix~\ref{appendix:related-work} gives an extended discussion.
\section{Dataset Description}
\textbf{Data Source and Selection}
We collect naturally occurring speech from publicly available YouTube videos covering localized varieties in \textit{Egypt}, \textit{Jordan}, \textit{Lebanon}, \textit{Mauritania}, \textit{Palestine}, and \textit{Saudi Arabia}. We prioritize non-scripted speech (interviews, vlogs, informal discussions) to capture conversational dialect use, including naturally occurring code-switching. To construct the micro-dialect inventory for each country, we first elicited candidate local varieties from native speaker annotators based on their linguistic and geographic knowledge.\footnote{We use \textit{micro-dialect} to denote a fine-grained sub-variety of a broader dialect \cite{abdul-mageed-etal-2020-toward}.} Highly similar or hard-to-distinguish candidates were consolidated into a broader category to reduce label redundancy. Annotators then curated source metadata by assigning YouTube URLs to the corresponding micro-dialect categories, using the dominant variety (the one used for most of the speech content) when a video contained more than one. Final micro-dialect labels were assigned per segment and could be multiple. The inventory broadly reflects geography: larger countries (\textit{Saudi Arabia}, \textit{Egypt}, \textit{Mauritania}) have more dispersed micro-dialect regions than the smaller Levantine ones.

\textbf{Release Format.} We release \cardamom{} as standoff annotations rather than redistributed audio: for each utterance we release the source YouTube video ID together with our human-produced transcription, micro-dialect label(s), perceived gender, code-switching status, and start/end timestamps, with code to re-fetch the audio from the original videos. Further release and licensing details are provided in the Ethics and Data Statement.

\textbf{Segmentation}
Recordings were split into annotation units of at most 20 minutes for manageable tasks and stable playback, preserving the source-level micro-dialect label.

\textbf{Team Members}
All annotators are co-authors and native speakers fluent in their country-level dialect and at least one of its micro-dialects. In total, the team is comprised of $31$ members: $5$ for Egypt, $8$ for Jordan, $2$ for Lebanon, $4$ for Mauritania, $3$ for Palestine, and $9$ for Saudi Arabia. Each member mainly annotated their own micro-dialect, and others from the same country when needed. Table~\ref{tab:annotators-per-microdialect} (Appendix~\ref{sec:annotation-schema}) breaks these totals down by micro-dialect. Per-dialect counts can exceed country totals, as some members cover several micro-dialects.

\section{Annotation Tasks and Guidelines}
\subsection{Overview}
All annotation tasks were uploaded to Label Studio\footnote{\url{https://labelstud.io/}}~\cite{tkachenko2020label} and organized by country. Annotators were provided with written guidelines\footnote{Available for review at this anonymized repository: \url{https://github.com/cardamom-paper/Cardamom-data}.} with labeling criteria and examples, refined through weekly team meetings.

We annotate each speech segment using a unified segment-level schema designed to support ASR, Arabic dialect identification (ADI), micro-dialect analysis, code-switching analysis, and aggregate analysis by perceived gender.

\subsection{Annotation Protocol}
As illustrated in Figure~\ref{fig:annotation} (Appendix~\ref{sec:annotation-schema}), annotators provide, for each segment: \textit{(i)} transcription reflecting the spoken form as it is typically written in everyday communication; \textit{(ii)} country-level dialect label and one or more micro-dialect labels; \textit{(iii)} perceived speaker gender; \textit{(iv)} a code-switching indicator; \textit{(v)} start and end timestamps.

\subsubsection{Task 1: Segment Selection}
Annotators select candidate segments from long-form videos using a pre-defined set of labels, including micro-dialect labels for dialect-specific content (e.g., in Egypt: Alexandrian, Cairene, Port Saidi, and Saidi) and MSA. Each selected segment must: \textit{(i)} contain speech from a single speaker with no overlap; \textit{(ii)} be audibly clear and transcribable, allowing for mild background noise; \textit{(iii)} contain at least one word and preferably be no longer than 30 seconds; and \textit{(iv)} capture a complete utterance with intact boundaries, including the full phonetic material of the first and last words.

Each segment is assigned at least one micro-dialect label when it contains dialectal speech, and more than one when it plausibly belongs to multiple micro-dialects. For example, the Jordanian utterance \Ar{ماشي ولا يهمك} (\textit{māshi walā yihimmak}, `okay, no worries') can be realized in both Falahi and Madani varieties; in such cases, both labels are applied.

\subsubsection{Task 2: Transcription}
Arabic dialects lack a single standardized writing system, so annotators transcribe the audio using a natural written form reflecting how the relevant dialect is commonly written in everyday communication, prioritizing fidelity to the speech signal over normalization to MSA. Annotators are encouraged to use standard Arabic orthographic markers when appropriate (e.g., hamza and tanwin) and to write numbers alphabetically to preserve inflectional and contextual realizations.

For code-switching (CS), annotators provide two transcript variants: \textit{(i)} a version in which foreign words are rendered in Arabic script, and \textit{(ii)} a version in which foreign words are rendered in Latin script.

\subsubsection{Task 3: Speaker Gender}
Annotators label speaker gender using the available label set \textit{\{male, female\}}. Labels reflect annotators' perception, serve only aggregate analysis, and make no claim about speakers' self-identified gender.

\subsection{Quality Control}
We manually inspected completed annotations for transcription completeness, segment boundary quality, single-speaker constraints, and consistency between country-level and micro-dialect labels, resolving ambiguous cases through discussion with annotators familiar with the relevant local varieties.

\subsection{Inter-annotator agreement}
\label{sec:agreement-audit}
Most segments were labeled once, so we measured agreement on a reannotated subset. We asked each country team to have 500 utterances relabeled by at least two new annotators; completed coverage varied and fell short of this target for several countries, so we report results per country without pooling (Table~\ref{tab:agreement-audit}). We compare pairs of new annotators only, not the original label, and, because a segment may carry several micro-dialect labels, report whether two label sets share any label, match exactly, or overlap by Jaccard similarity.

\begin{table}[t]
\centering
\small
\setlength{\tabcolsep}{3pt}
\begin{tabular}{@{}lccccc@{}}
\toprule
& & & \multicolumn{3}{c}{\textbf{Micro-dialect}} \\
\cmidrule(lr){4-6}
\textbf{Country} & \textbf{Ann.} & \textbf{Gender} & \textbf{Any} & \textbf{Exact} & \textbf{Jaccard} \\
\midrule
Egypt & 3 & 95.76 & 83.05 & 75.42 & 79.10 \\
Jordan & 3 & 97.20 & 88.11 & 65.73 & 76.22 \\
Lebanon & 3 & 99.31 & 84.72 & 50.69 & 64.93 \\
Mauritania & 4 & 97.17 & 80.72 & 31.15 & 51.09 \\
Palestine & 2 & 98.39 & 96.77 & 96.77 & 96.77 \\
Saudi Arabia & 6 & 98.81 & 95.26 & 86.40 & 90.44 \\
\bottomrule
\end{tabular}
\caption{Pairwise agreement (\%) between new annotators on the reannotated subset. \textit{Ann.}: number of reannotators; \textit{Gender}: identical perceived gender; \textit{Any}: label sets share at least one label (e.g., \{Northern, Southern\} vs.\ \{Northern\}); \textit{Exact}: identical sets; \textit{Jaccard}: mean intersection over union.}
\label{tab:agreement-audit}
\end{table}

Perceived-gender agreement is 95.8--99.3\%; some disagreements appear to involve young or childlike voices. Micro-dialect agreement is high under any overlap (80.7--96.8\%) but lowest under exact match, which requires annotators to select identical label sets. This difference is most pronounced for Mauritania, where native speakers describe the regional varieties as closely related and frequently applicable to the same utterance. Annotators therefore often agree on at least one plausible variety while differing in the additional labels selected, reflecting genuine boundary ambiguity rather than complete disagreement about the speech variety.



\section{Micro-Dialects Overview}
Arabic dialects exhibit substantial regional and social variation, and broad national labels often obscure the localized varieties that speakers use in everyday communication~\cite{JZHNW23, holes2004modern, prochazka2021arabic, zaidan2014arabic}. For this reason, \cardamom{} is designed to represent fine-grained micro-dialectal variation  across \textit{Egypt}, \textit{Jordan}, \textit{Lebanon}, \textit{Mauritania}, \textit{Palestine}, and  \textit{Saudi Arabia}. We use \textit{micro-dialect} as an operational label for localized sub-country varieties that are recognizable to speakers and reflected in systematic differences in pronunciation, morphology, lexical choice, prosody, and grammatical patterns \cite{holes2004modern, prochazka2021arabic}. Capturing this level of variation is important for evaluating Arabic speech technologies beyond coarse country-level labels, which can mask substantial variation within the same national dialect category~\cite{salameh-etal-2018-fine, abdul-mageed-etal-2020-toward}.

Our taxonomy is informed by prior dialectological literature but is not intended to mechanically reproduce any single linguistic classification. Instead, it adapts established descriptions into categories suitable for computational speech evaluation, acknowledging that dialect boundaries are often gradual and shaped by regional, social, and historical contact. The labels therefore serve as operational annotation categories rather than a definitive linguistic classification; Appendix~\ref{appendix:microdialects-overview} discusses this distinction further, including a case where our own cited sources support finer subdivisions than the granularity we annotate.

Accordingly, \cardamom{} represents Egyptian speech through \textit{Alexandrian}, \textit{Cairene}, \textit{Port Saidi}, and \textit{Saidi} groups; Jordanian speech through \textit{Madani}, \textit{Falahi}, and \textit{Badawi}; Lebanese speech through \textit{Chamali}, \textit{Jnoubi}, and \textit{B'albaki}; Mauritanian \textit{\d{H}ass\=aniyya} through \textit{Western}, \textit{Northern}, and \textit{Southern/Eastern} groupings; Palestinian speech through \textit{Madani urban}, \textit{Northern Fallahi}, and \textit{Southern Badawi}; and Saudi speech through broad regional categories including \textit{Najdi}, \textit{Hijazi}, \textit{Eastern}, \textit{Northern}, and \textit{Southern}. These categories are not intended to exhaustively represent all local varieties within each country; rather, they define annotation units that are recognizable to native-speaker annotators, sufficiently frequent in publicly available speech, and useful for testing whether ASR systems generalize beyond broad country-level dialect labels. Appendix~\ref{appendix:microdialects-overview} provides a fuller discussion of the dialect classifications reported in the literature for the countries covered in \cardamom{}.


\subsection{Phonological Motivation for Micro-Dialect Annotation}
\cardamom{} captures several well-known axes of Arabic micro-dialectal variation, including reflexes of /q/, interdental preservation or merger, velar palatalization, and affrication of /\textipa{\t{dZ}}/, each realized differently across the Egyptian, Jordanian, Palestinian, Saudi, Lebanese, and Mauritanian varieties in the corpus (e.g., MSA \Ar{قلب} \textit{qalb} ``heart'' surfaces as \Ar{ألب} \textit{\textglotstop{}alb} in Cairene, \Ar{كلب} \textit{kalb} in Palestinian North Rural, and \Ar{گلب} \textit{galb} in Hassaniya). These patterns motivate fine-grained annotation beyond country-level dialect labels; Table~\ref{tab:cross_dialect_phonology} (Appendix~\ref{appendix:microdialects-overview}) gives the full set of phoneme-by-phoneme correspondences with lexical examples and citations.
\section{Corpus Statistics and Analysis}
\label{sec:corpus_stats}

\begin{table}[t!]
\centering
\small
\setlength{\tabcolsep}{2.5pt}
\resizebox{\columnwidth}{!}{%
\begin{tabular}{llccccc}
\hline

\textbf{Country} & \textbf{Micro-dialect} & \textbf{Segments} & \textbf{Total Hrs} & \textbf{Mean} & \textbf{CS Hrs} & \textbf{F/M} \\
\hline

\multirow{5}{*}{\textbf{\Egypt}}
& Cairene & 3,871 & 3.482 & 3.238 & 0.264 & 33.4/66.6 \\
& Saidi & 1,735 & 1.771 & 3.675 & 0.008 & 37.9/62.1 \\
& Port Saidi & 1,124 & 1.021 & 3.269 & 0.039 & 12.2/87.8 \\
& Alexandrian & 848 & 0.769 & 3.265 & 0.020 & 40.7/59.3 \\
& \textbf{\textit{TOTAL}} & \textbf{7,364} & \textbf{6.889} & \textbf{3.368} & \textbf{0.327} & \textbf{32.1/67.9} \\
\hline

\multirow{4}{*}{\textbf{\Jordan}}
& Falahi & 3,276 & 4.540 & 4.989 & 0.119 & 18.6/81.4 \\
& Badawi & 2,899 & 3.968 & 4.928 & 0.005 & 26.6/73.4 \\
& Madani & 1,909 & 2.755 & 5.196 & 0.309 & 38.2/61.8 \\
& \textbf{\textit{TOTAL}} & \textbf{7,729} & \textbf{10.806} & \textbf{5.033} & \textbf{0.407} & \textbf{26.5/73.5} \\
\hline

\multirow{4}{*}{\textbf{\Lebanon}}
& Chamali & 1,105 & 2.071 & 6.746 & 0.560 & 13.6/86.4 \\
& Jnoubi & 1,464 & 3.123 & 7.679 & 0.212 & 19.4/80.6 \\
& B'albaki & 258 & 0.737 & 10.281 & 0.185 & 28.4/71.6 \\
& \textbf{\textit{TOTAL}} & \textbf{2,681} & \textbf{5.628} & \textbf{7.557} & \textbf{0.902} & \textbf{18.0/82.0} \\
\hline

\multirow{4}{*}{\textbf{\Mauritania}}
& Western & 1,210 & 3.780 & 11.247 & 0.464 & 8.3/91.7 \\
& Northern & 1,205 & 3.809 & 11.380 & 0.470 & 8.1/91.9 \\
& Southern & 1,126 & 3.666 & 11.720 & 0.467 & 8.7/91.3 \\
& \textbf{\textit{TOTAL}} & \textbf{1,314} & \textbf{3.998} & \textbf{10.953} & \textbf{0.473} & \textbf{7.7/92.3} \\
\hline

\multirow{4}{*}{\textbf{\Palestine}}
& Falahi & 1,564 & 2.720 & 6.260 & 0.028 & 9.0/91.0 \\
& Badawi & 864 & 1.461 & 6.088 & 0.001 & 3.1/96.9 \\
& Madani & 537 & 0.997 & 6.686 & 0.001 & 32.3/67.7 \\
& \textbf{\textit{TOTAL}} & \textbf{2,956} & \textbf{5.159} & \textbf{6.283} & \textbf{0.031} & \textbf{11.5/88.5} \\
\hline

\multirow{6}{*}{\textbf{\SaudiArabia}}
& Southern & 1,423 & 2.550 & 6.452 & 0.088 & 40.5/59.5 \\
& Hijazi & 1,413 & 2.661 & 6.779 & 0.250 & 95.0/5.0 \\
& Najdi & 1,089 & 2.234 & 7.384 & 0.156 & 3.4/96.6 \\
& Eastern & 956 & 2.023 & 7.618 & 0.044 & 17.9/82.1 \\
& Northern & 514 & 1.009 & 7.064 & 0.003 & 0.4/99.6 \\
& \textbf{\textit{TOTAL}} & \textbf{5,376} & \textbf{10.425} & \textbf{6.981} & \textbf{0.538} & \textbf{39.6/60.4} \\
\hline
\end{tabular}%
}
\caption{\textsc{Cardamom} statistics by country and micro-dialect. \textit{Total Hrs} gives speech duration; \textit{Mean} is average segment duration in seconds; \textit{CS Hrs} gives code-switched speech duration; \textit{F/M} reports the perceived speaker-gender split (see \textit{Gender distribution bias} in Limitations). Country-level totals (\textbf{bold}) count each segment once; per-dialect rows count label memberships and can exceed the total when segments carry multiple micro-dialect labels, most visibly for Mauritania (Table~\ref{tab:dialect_intersections}). An extended version with segment-duration range, transcript length, vocabulary, and speaking-rate statistics is in Table~\ref{tab:micro-dialect-detailed-full} (Appendix~\ref{appendix:microdialectal-overlap}).}
\label{tab:micro-dialect-detailed}
\end{table}


\subsection{Micro-dialect and Transcript Statistics}
As Table~\ref{tab:micro-dialect-detailed} shows, the corpus comprises {27{,}420} segments totaling {$\sim$$40$} hours. While some varieties, such as Egyptian Cairene and Jordanian dialects, are well represented, others, such as Lebanese B'albaki and Saudi Northern, are less represented but add unique linguistic diversity.
 
Average utterance duration varies notably across dialects, with Egyptian micro-dialects having shorter segments and Mauritanian the longest; Lebanese varieties fall in a mid-to-upper range (6.7--10.3 seconds). CS prevalence is unevenly distributed across micro-dialects, with Mauritanian, Jordanian Madani, Lebanese, and Egyptian varieties contributing the largest amounts, while Palestinian, Saudi, and Jordanian Bedouin subsets contain very little CS speech. Appendix~\ref{appendix:codeswitch-examples} provides representative CS examples. Gender splits also vary strongly by micro-dialect.

A similar pattern is observed in the transcript's length and lexical diversity (Table~\ref{tab:micro-dialect-detailed-full}, Appendix~\ref{appendix:microdialectal-overlap}). Lebanese and Mauritanian subsets exhibit the longest transcripts and some of the highest vocabulary rates. Using words-per-second (WPS) and characters-per-second (CPS) as coarse proxies for speaking rate, the Egyptian and some Saudi dialects exhibit the fastest speech, while Palestinian and Lebanese are comparatively slower, consistent with prior findings \cite{talafha2024casablanca}.
The above findings are influenced by the domain of the underlying YouTube sources (e.g., interviews vs. vlogs, topic, recording conditions, and conversational formality) and by the segmentation strategy.\footnote{For example, podcast-style recordings often contain longer, more monologic turns than daily conversations. Thus, differences in duration, code-switching prevalence, lexical statistics, and speech-rate metrics may partially reflect source-domain composition rather than linguistic variation.}

\subsection{Quantitative Micro-dialectal Overlap}
\label{subsec:quant_overlap}

To complement the aggregate corpus statistics, we examine the frequency and composition of multi-label assignments. These assignments identify utterances that annotators judged compatible with more than one micro-dialect and therefore provide a corpus-level view of where the operational label boundaries overlap. 

The extent of co-assignment varies substantially across the sampled regions. As shown in Table~\ref{tab:dialect_intersections} (Appendix~\ref{appendix:microdialectal-overlap}), Mauritania has the largest pairwise overlap counts, with extensive co-assignment among the three Hassaniya-oriented regional labels.
This pattern should be interpreted with caution.
Localized Mauritanian material was scarce, so much of the data is broadly shared Hassaniya speech compatible with several regional labels. We therefore treat Mauritania as a distinct multi-label case in our evaluations (Section~\ref{sec:dialect-id}).

Outside Mauritania, Jordan has the largest  overlap counts: Badawi--Falahi and Falahi--Madani occur in 178 and 171 multi-label segments, respectively, compared with 23 for Badawi--Madani. Falahi thus shares features with both, without this establishing it as an intermediate variety. In Egypt, co-assignment is concentrated around Cairene, which shares substantial overlap with the coastal varieties  Alexandrian (130) and Port Saidi (67), whereas overlap involving Saidi and between other pairs of regional varieties is sparse (2--10). Saudi Arabia shows much less co-assignment, with most observed intersections containing only a few segments (e.g., one Eastern--Southern segment). Low counts are not evidence of categorical separation.

Table~\ref{tab:multi_label_examples} illustrates the kinds of utterances receiving multiple labels across countries with different overlap profiles. In Jordan, \Ar{يلا بخاطرك} (`alright, see you!') (Badawi, Falahi) and \Ar{اه والله مزبوط} (`yes, that is correct') (Falahi, Madani) contain widely shared lexical items and discourse expressions that provide limited evidence for assignment to a single variety. Similarly, the Egyptian examples \Ar{استنى يابا} (`wait, Dad') (Alexandrian, Port Saidi) and \Ar{فلوس باباكي متلزمناش} (`we don't need your father's money') (Alexandrian, Cairene) illustrate how common everyday expressions may be compatible with multiple neighboring varieties. In Saudi Arabia, where co-assignment is less frequent, \Ar{تماما ينطبق نفس الشي على الرجال} (`exactly, the same thing applies to men') (Hijazi, Najdi) illustrates how relatively formal or broadly shared phrasing may provide few variety-specific cues. Overall, the examples suggest that multi-label assignments can arise from shared lexical and discourse material, limited variety-specific evidence within short utterances, and genuine overlap among localized varieties. Their distribution should therefore be understood as a property of both the sampled speech and the annotation framework.


Co-assignment counts may also reflect source domain and annotators' differing readiness to apply multiple labels (relevant to the low Palestinian counts). As YouTube speakers may recur, Appendix~\ref{appendix:speaker-clustering} estimates speaker diversity via speaker-embedding clustering.

\section{Zero-Shot Benchmarking}


\begin{table}[t!]
\centering
\setlength{\tabcolsep}{1pt}
\small
\resizebox{0.4\textwidth}{!}{%
\begin{tabular}{llrrrrrrrr}
\toprule
\textbf{Cnt.} & \textbf{M-Dia} & \multicolumn{2}{c}{\textbf{\textsc{O}-CTC}} & \multicolumn{2}{c}{\textbf{\textsc{O}-LLM}} & \multicolumn{2}{c}{\textbf{Sea-v2}} & \multicolumn{2}{c}{\textbf{W-v3}} \\
\cmidrule(lr){3-4} \cmidrule(lr){5-6} \cmidrule(lr){7-8} \cmidrule(lr){9-10}
 & & \scriptsize{WER} $\downarrow$  & \scriptsize{CER} $\downarrow$ & \scriptsize{WER} $\downarrow$ & \scriptsize{CER} $\downarrow$ & \scriptsize{WER} $\downarrow$ & \scriptsize{CER} $\downarrow$ & \scriptsize{WER} $\downarrow$ & \scriptsize{CER} $\downarrow$ \\
\hline
\multirow{4}{*}{\textbf{\EGY}} & Alex. & 49.0 & 20.6 & \textbf{41.4} & \textbf{20.3} & 46.7 & 21.0 & 61.9 & 33.5 \\
 & Cairene & 44.9 & 19.7 & \textbf{36.1} & \textbf{17.9} & 38.3 & 18.2 & 59.8 & 32.0 \\
 & Port Said & 56.4 & 25.6 & \textbf{46.1} & 25.2 & 50.5 & \textbf{24.1} & 71.5 & 36.6 \\
 & Saidi & 53.0 & 22.9 & \textbf{44.1} & \textbf{20.8} & 49.1 & 21.8 & 69.0 & 35.4 \\
\hline
\multirow{3}{*}{\textbf{\JOR}} & Badawi & 44.1 & 16.6 & \textbf{33.6} & \textbf{12.2} & 46.0 & 20.2 & 45.8 & 25.1 \\
 & Falahi & 44.6 & 18.9 & \textbf{33.0} & \textbf{13.7} & 41.5 & 18.5 & 41.1 & 19.5 \\
 & Madani & 49.2 & 26.6 & \textbf{28.9} & \textbf{11.9} & 40.7 & 22.1 & 41.6 & 22.2 \\
\midrule
\multirow{3}{*}{\textbf{\LEB}} & B'albaki & 92.0 & 79.7 & 64.1 & 40.0 & 75.2 & 50.7 & \textbf{62.2} & \textbf{34.1} \\
 & Chamali & 91.5 & 79.8 & \textbf{38.9} & \textbf{21.6} & 75.3 & 53.0 & 77.6 & 45.3 \\
 & Jnoubi & 72.9 & 50.8 & \textbf{45.2} & \textbf{23.0} & 60.9 & 33.5 & 75.4 & 51.8 \\
\midrule
\multirow{3}{*}{\textbf{\MAU}} & Northern & 87.2 & 69.1 & \textbf{76.0} & \textbf{48.5} & 84.0 & 63.1 & 100.9 & 72.5 \\
 & Southern & 87.2 & 69.4 & \textbf{75.9} & \textbf{48.5} & 84.1 & 63.4 & 101.2 & 72.9 \\
 & Western & 87.2 & 69.4 & \textbf{76.2} & \textbf{49.2} & 84.4 & 63.4 & 99.5 & 72.1 \\
\midrule
\multirow{3}{*}{\textbf{\PAL}} & Badawi & 49.3 & 20.2 & \textbf{45.0} & 21.1 & 45.0 & \textbf{19.5} & 47.6 & 21.1 \\
 & Falahi & 51.9 & 23.4 & \textbf{46.4} & \textbf{22.9} & 51.5 & 23.9 & 62.1 & 34.9 \\
 & Madani & 49.5 & \textbf{24.1} & \textbf{44.5} & 26.6 & 56.9 & 44.1 & 47.9 & 28.6 \\
\midrule
\multirow{5}{*}{\textbf{\KSA}} & Eastern & 57.3 & 31.6 & \textbf{35.9} & \textbf{17.1} & 43.3 & 20.4 & 48.6 & 29.2 \\
 & Hijazi & 44.1 & 16.2 & \textbf{36.0} & \textbf{12.9} & 40.1 & 16.1 & 40.9 & 14.4 \\
 & Najdi & 61.7 & 39.6 & \textbf{39.4} & \textbf{21.1} & 52.5 & 31.6 & 49.7 & 30.9 \\
 & Northern & 80.3 & 54.8 & \textbf{67.5} & \textbf{44.9} & 87.5 & 52.8 & 107.2 & 64.3 \\
 & Southern & 56.3 & 35.5 & \textbf{36.5} & \textbf{15.6} & 57.9 & 39.6 & 59.3 & 41.5 \\
\bottomrule
\end{tabular}
}
\caption{WER and CER, in \%, per model across countries and micro-dialects (test set). \textit{Cnt}: Country, \textit{M-Dia}: Micro-dialect, \textit{O-}: Omni, \textit{Sea-v2}: Seamless-v2, and \textit{W-v3}: Whisper-v3. Best result per dialect is in \textbf{bold}. Per-dialect test-set sizes are listed in Table~\ref{tab:micro-split-counts} (Appendix~\ref{sec:data-splits}); Table~\ref{tab:asr-zeroshot-ci} (Appendix~\ref{sec:asr-adaptation-details}) reports 95\% bootstrap confidence intervals for WER per micro-dialect.}
\label{tab:wer-cer-results}
\end{table}

We evaluate \cardamom{} primarily in a zero-shot ASR setting, then show that its train/development split supports adaptation (Section~\ref{sec:asr-adaptation}) and that the micro-dialect labels support identification as a downstream task (Section~\ref{sec:dialect-id}), and analyze transcript-level dialectness to contextualize ASR performance (Appendix~\ref{app:dialectness}). We split the corpus into train, development, and test sets at the utterance level, balancing micro-dialect memberships within each country; MDI additionally uses a source-video-disjoint split (Section~\ref{sec:dialect-id}). Details are in Appendix~\ref{sec:data-splits}.


\subsection{Automatic Speech Recognition (ASR)}
\label{sec:asr-benchmark}

We evaluate zero-shot ASR on the test and validation splits using four recent multilingual speech recognition systems, without fine-tuning on \cardamom{}. Performance is reported with word error rate (WER) and character error rate (CER), after applying the same text-normalization pipeline to all systems (Appendix~\ref{appendix:preprocessing}); because dialectal Arabic lacks standardized orthography, these values reflect both acoustic recognition errors and residual orthographic mismatch after normalization. For each micro-dialect, we evaluate all test segments annotated with that label, including segments that overlap with other micro-dialects. MSA-only clips are excluded.

\paragraph{Models.}
\textsc{Whisper-v3}, a large multilingual encoder--decoder model, is a strong Arabic ASR baseline~\cite{radford2023robust}. \textsc{Seamless-v2} is a multilingual speech model supporting transcription and translation~\cite{barrault2023seamless}. \textsc{OmniASR-LLM-7B-V2} and \textsc{OmniASR-CTC-7B-V2} are omnilingual ASR models differing in decoding strategy, generative LLM-based versus CTC-based~\cite{keren2025omnilingual}.
Table~\ref{tab:wer-cer-results} reports WER and CER by country and micro-dialect. Overall, \textsc{OmniASR-LLM-7B-V2} is the strongest system, obtaining the best WER for 20 of 21 micro-dialects and the best CER for 17 of 21, with especially clear gains for Jordanian dialects (WER 28.9--33.6) and Saudi Eastern, Najdi, and Southern. The only WER exception is Lebanese B'albaki, where Whisper-v3 is best (62.2), though OmniASR-LLM remains competitive.
Results vary substantially across regions: Jordanian micro-dialects are easiest overall; Egyptian and Palestinian dialects fall in a middle range, with errors remaining high for Port Said, Saidi, and Palestinian varieties; and Lebanese and Mauritanian dialects are considerably more challenging. In Lebanon, all models struggle on B'albaki, while OmniASR-LLM is much stronger on Chamali and Jnoubi. Mauritanian varieties are hardest overall (best model still $\sim$76 WER, 48--49 CER), suggesting a major mismatch with the pretraining data distribution. Within most countries, the mainstream capital-associated variety yields the lowest WER (e.g., Cairene, Madani), likely reflecting their prominence in web-scale pretraining data. Saudi Arabia is the exception: despite Najdi being mainstream, Eastern yields the lowest WER, likely due to proximity to the broader Gulf variety.

\subsection{Adaptation}
\label{sec:asr-adaptation}
 
To demonstrate the intended use of its train/development/test partitions and give a more cautious, uncertainty-aware picture of the per-dialect comparisons in Table~\ref{tab:wer-cer-results}, we additionally fine-tune all four systems on the \cardamom{} training split (Appendix~\ref{sec:asr-adaptation-details} reports checkpoints, hyperparameters, and compute). Whisper-v3 and Seamless-v2 are adapted with LoRA on the query/value projections; both \textsc{OmniASR} variants are fully fine-tuned. Validation uses the development split; the test split is never used during adaptation. We report 95\% confidence intervals from 2,000 bootstrap resamples over the test set, including for the smallest Lebanese subsets (B'albaki: 81, Chamali: 362, Jnoubi: 478 test utterances), capturing test-sample uncertainty rather than training-seed variance, since each adapted model was trained once.

Table~\ref{tab:asr-adaptation-overall} (Appendix~\ref{sec:asr-adaptation-details}) compares aggregate WER/CER under the zero-shot and adapted settings. Adaptation reduces both error rates for Whisper-v3 and the two \textsc{OmniASR} variants. The adapted \textsc{OmniASR-LLM} model achieves the lowest overall WER and CER, reaching 35.21\% WER and 16.04\% CER, compared with  43.47\% and 20.67\%, respectively in the zero-shot setting. For Seamless-v2, adaptation yields only a small reduction in WER (53.13\% to 52.48\%), although its CER decreases more clearly (28.84\% to 25.38\%). The overall pattern of relative difficulty across micro-dialects remains qualitatively similar after adaptation, but differences involving small test subsets, such as Lebanese B'albaki and Saudi Northern, should not be interpreted as conclusive. As an oracle diagnostic of dialect conditioning, prompting zero-shot Whisper-v3 with each clip's gold country and micro-dialect label(s) increased WER from 55.38\% to 61.43\% (Appendix~\ref{sec:asr-adaptation-details}); we did not test dialect conditioning for the other systems.

On a stricter source-video-disjoint split, adapting \textsc{OmniASR-LLM} reduces WER from 43.51\% to 35.24\% and CER from 21.84\% to 15.83\%; Appendix~\ref{sec:asr-source-disjoint} gives the protocol, source-clustered confidence intervals, and per-micro-dialect results.


\subsection{Micro-Dialect Identification}
\label{sec:dialect-id}
 


The micro-dialect labels 
also support micro-dialect identification (MDI) from audio. We define three tasks: \textit{(i)} country identification (6-way), \textit{(ii)} global micro-dialect 
identification over all label names, and \textit{(iii)} country-conditioned identification 
given the gold country. Because a clip may carry several gold labels $G_i$ while the classifier predicts one label $\hat{y}_i$, we report any-match accuracy, $\mathrm{Acc}_{\mathrm{any}}=\frac{1}{N}\sum_{i=1}^{N}\mathbf{1}[\hat{y}_i\in G_i]$, and macro-F1 over one-vs-rest label decisions.

Since the ASR split places segments from the same YouTube video, and hence the same speakers and recording conditions, in different partitions, MDI uses complete videos as units in an 80/10/10 split (no shared video). We train a \textsc{Whisper-large-v3} classifier with a shared encoder and one head per task, using three seeds (Appendix~\ref{sec:dialect-id-appendix}).

\begin{table}[t]
\centering
\footnotesize
\setlength{\tabcolsep}{2pt}
\renewcommand{\arraystretch}{0.85}
\begin{tabular}{@{}llcccc@{}}
\toprule
\textbf{Task} & & \textbf{Maj.} & \textbf{Whisper} & \textbf{w/o MAU} & \textit{\textbf{Utt.}} \\
\midrule
Country & Acc. & 24.20 & $72.97\pm2.72$ & $74.54\pm3.18$ & \textit{95.67} \\
        & F1   & 6.49  & $64.56\pm3.46$ & $69.68\pm3.93$ & \textit{95.72} \\
\midrule
Global  & Acc. & 15.77 & $46.85\pm1.14$ & $47.00\pm1.66$ & \textit{85.57} \\
        & F1   & 1.51  & $31.95\pm1.23$ & $34.85\pm0.67$ & \textit{73.76} \\
\midrule
Cond.   & Acc. & 50.26 & $59.60\pm2.69$ & $57.55\pm2.86$ & \textit{88.20} \\
        & F1   & 19.05 & $44.45\pm0.82$ & $44.08\pm1.07$ & \textit{77.11} \\
\bottomrule
\end{tabular}
\caption{MDI accuracy and macro-F1 (\%) on the source-disjoint test set (3,062 clips, 46 videos): training-majority baseline and mean $\pm$ sample SD over three seeds, with and without the 162 Mauritanian clips. \textit{Utt.}: the same classifier on the utterance-level ASR split, shown only as a leakage-sensitive comparison.}
\label{tab:mdi-main}
\end{table}

The classifier exceeds the majority baseline on every task (Table~\ref{tab:mdi-main}), but global accuracy below 50\% shows that fine-grained identification on unseen sources remains difficult. Retraining with different seeds changes scores by only 0.8--3.5 points, whereas resampling the 46 held-out test videos yields a 95\%  interval of 30.98--68.68 for global accuracy; results therefore depend more on which videos are held out than on training randomness. On the utterance-level ASR split, the same classifier scores much higher (\textit{Utt.}) because speakers and videos are shared between training and test, so those scores overstate performance on unseen speakers. Mauritania is a distinct case: 151 of its 162 test clips carry all three regional labels, so conditioned any-match accuracy is permissive (96.30\%) while exact-set accuracy is 3.09\% (Appendix~\ref{sec:mauritania-mdi}). For context, the released ADI-20~\cite{elleuch2025adi} and VoxLect~\cite{feng2026voxlect} checkpoints reach 53.14\% (restricted to our six countries) and 38.96\% (four regions) accuracy on the same clips; given their different training data and coarser labels, these are not controlled comparisons (Appendix~\ref{sec:external-baselines}).

\section{Conclusion}


We introduced \textsc{Cardamom}, a micro-dialectal Arabic speech corpus covering 21 localized varieties across Egypt, Jordan, Lebanon, Mauritania, Palestine, and Saudi Arabia. With $\sim$40 hours of community-curated, transcribed YouTube speech and utterance-level perceived-gender labels, \textsc{Cardamom} enables fine-grained ASR evaluation beyond coarse country-level dialect labels. Zero-shot benchmarking of four SOTA multilingual ASR systems confirms that micro-dialectal variation poses uneven challenges, with Mauritanian and Lebanese varieties hardest and \textsc{OmniASR-LLM} the strongest baseline overall; adapting these systems on \cardamom{}'s training split narrows, but does not close, these gaps. Micro-dialect identification on held-out source videos remains difficult (46.85\% global accuracy) and far below utterance-level results, underscoring the need for source-disjoint evaluation. Because subset size, gender balance, and label distinctness vary, \cardamom{} is best used as a diagnostic and adaptation benchmark rather than for precise dialect rankings. In future work, we plan to expand \textsc{Cardamom} to additional Arabic countries and micro-dialects, and to enrich the annotation with phoneme transcriptions, emotion labels, and translation.

\section*{Acknowledgments}
Muhammad Abdul-Mageed acknowledges support from Canada Research Chairs (CRC), the Natural Sciences and Engineering Research Council of Canada (NSERC; RGPIN-2026-07098), the Social Sciences and Humanities Research Council of Canada (SSHRC; 895-2020-1004), Canadian Foundation for Innovation (CFI; 37771), Digital Research Alliance of Canada, and UBC ARC-Sockeye.
\section*{Limitations}
\paragraph{Incomplete micro-dialect coverage.}
Although \cardamom{} spans 21 micro-dialects across six countries, the coverage within each country is uneven. Several micro-dialects, such as Lebanese B'albaki and Saudi Northern, contain substantially fewer segments than their better-represented counterparts (Table~\ref{tab:wer-cer-results}). This imbalance reflects the limited availability of suitable public speech material for certain localized varieties rather than an editorial choice, and it constrains the conclusions that can be drawn about these subsets. We mitigate this limitation by reporting results at the micro-dialect level, documenting the size of each subset, and interpreting low-resource varieties as diagnostic evaluation cases rather than as balanced training partitions. More broadly, \cardamom{} does not cover all Arabic-speaking countries or all sub-national varieties within the six countries it includes; dialects from the Maghreb, Gulf, or Nile Valley periphery, for instance, are absent. Future extensions should expand coverage to additional countries and underrepresented local varieties, prioritizing micro-dialects for which current ASR systems show high error rates or limited robustness.

\paragraph{Domain and genre effects.}
The corpus is drawn entirely from YouTube, and the distribution of genres, including podcasts, vlogs, interviews, old television serials, and informal conversations, is uneven across micro-dialects. As noted in Section~\ref{sec:corpus_stats}, aggregate statistics such as average segment duration, code-switching prevalence, speaking rate, and lexical diversity are influenced by the domain composition of the underlying sources, not only by linguistic properties of the variety. We do not have a reliable way to retroactively assign a genre label (interview, vlog, podcast, broadcast, etc.) to each of the 27{,}420 segments, since annotators were free to pull from any non-scripted, naturally occurring content they could find for their variety, and many of these are casual personal videos with no clean genre metadata attached at collection time; for several of our lowest-resource micro-dialects, suitable public speech was already scarce, so our priority during collection was finding any usable naturalistic speech for that variety at all, rather than sampling within specific genre categories. Our two content requirements were unscripted, non-read-out speech, to capture naturalistic dialectal usage, and segment-level audibility and transcribability, regardless of recording setting; we did not filter or track by recording environment (e.g., studio vs.\ phone-recorded), since intelligibility rather than production quality was our inclusion bar. \cardamom{}'s source material is naturalistic and unscripted by design, but genre and recording-condition distribution across the corpus is uncontrolled and unlabeled. We mitigate this limitation by reporting duration, code-switching, and speaking-rate statistics descriptively rather than treating them as definitive dialectal properties, and by interpreting cross-variety differences in light of possible source-domain effects. Future extensions should increase genre balance within each micro-dialect and include more controlled sampling across conversational settings, recording conditions, and registers.

\paragraph{Gender distribution bias.}
The perceived gender distribution in \cardamom{} varies substantially across micro-dialects and is highly skewed in some subsets (e.g., Saudi Hijazi is 95\% female, while Saudi Northern is 99.6\% male). These imbalances reflect the availability of suitable publicly accessible speech from each variety rather than a principled sampling strategy: for several of our lower-resource micro-dialects, publicly available YouTube content covering that specific variety was already scarce, which left little room to also balance for speaker gender within that subset, since annotators could only work with the videos that existed for a given variety rather than curate toward a target gender distribution. This is a known pattern in YouTube-sourced dialectal speech corpora more broadly; Casablanca~\cite{talafha2024casablanca}, for instance, reports similarly skewed gender distributions across some of its dialects for the same underlying reason. We mitigate this limitation by reporting gender distributions explicitly next to the corpus statistics (Table~\ref{tab:micro-dialect-detailed}) and by treating gender metadata as suitable for aggregate descriptive analysis rather than balanced subgroup comparison within every micro-dialect. Model evaluations stratified by gender should therefore account for the underlying data skew, and future extensions should target more balanced speaker coverage where suitable public speech is available.


\paragraph{Linguistic proximity across borders.}
Some micro-dialects that straddle national boundaries share strong phonological and lexical features, making their boundaries difficult to distinguish computationally and even perceptually. For example, Jordanian Badawi and Saudi Northern both exhibit Bedouin-type features, including the realization of /q/ as [g] and conservative vowel systems, that are more similar to each other than to the urban varieties of their respective countries. We mitigate this issue by treating such cases as expected boundary phenomena rather than annotation failures, and by using multi-label annotation where an utterance plausibly belongs to more than one local variety.

\paragraph{Orthographic inconsistency.}
Arabic dialects lack a standardized writing system, and transcription conventions vary across annotators, regions, and even within the same speech community. A single spoken form may therefore have multiple valid written realizations; for example, \Ar{شفتو} and \Ar{شفته} ('I saw him') are both attested across Egyptian varieties. We mitigate this limitation by providing annotators with shared transcription guidelines and by applying the same normalization pipeline across all ASR systems before computing WER and CER. Nevertheless, residual orthographic variation is unavoidable and may inflate error rates beyond purely acoustic recognition errors. This is a known challenge for dialectal Arabic ASR evaluation and further motivates future work with multi-reference evaluation metrics~\cite {ali2015multi}.

\paragraph{Annotation agreement.}
Because the corpus was collected through community-based annotation in which annotators primarily worked on varieties they know well, most segments were labeled once. We therefore report agreement on a reannotated subset (Section~\ref{sec:agreement-audit}) rather than for the full corpus. Reannotation coverage differs by country, and exact-set agreement is low for Mauritania (31.1\%) and Lebanon (50.7\%), even though the annotators usually share at least one label. Fine-grained labels in these subsets should be interpreted with corresponding caution. Future extensions should prioritize sources with clearer regional provenance for closely related micro-dialects, such as the Mauritanian varieties, so that segments can be assigned to a single variety more reliably.

\paragraph{Evaluation design.}
The ASR split is balanced at the utterance level and is not source-disjoint, so adapted ASR results may overestimate performance on unseen speakers and sources. Each adapted ASR system was trained once, so its confidence intervals reflect test-sample uncertainty only; repeated training runs are reported for the MDI classifier. The source-disjoint MDI split prevents same-video overlap, but the same person could still appear in different videos, and several labels are concentrated in few held-out videos (Table~\ref{tab:mdi-source-split}).


\paragraph{Source availability over time.} As with any YouTube-sourced corpus, some source videos may become unavailable or be removed over time, whether by the uploader, by YouTube, or through our own takedown process. This is a known and unavoidable property of building speech resources from public video platforms, and it affects prior work in the same space, including the MGB Challenge series~\cite{ali2017speech,ali2019mgb}, GigaSpeech~\cite{chen2021gigaspeech}, VoxLingua107~\cite{valk2021voxlingua107}, Casablanca~\cite{talafha2024casablanca}, and AV-Speech~\cite{ephrat2018looking}. Because we release \cardamom{} as standoff annotations tied to source video IDs (Ethics and Data Statement), a share of the corpus may become unreconstructable for later users as source videos disappear; we mitigate this by releasing the dataset as early as possible and by not depending on any single mirror of the source material.

\paragraph{Data scale.}
With approximately 40 hours of transcribed speech, \cardamom{} is designed as an evaluation and adaptation resource rather than a large-scale training corpus. Training a full end-to-end ASR model from scratch on \cardamom{} alone would not usually be feasible, as modern large-scale speech models require orders of magnitude more data. The intended use cases are zero-shot and few-shot evaluation, fine-tuning or adapter-based adaptation of existing multilingual models, and micro-dialect identification; Sections~\ref{sec:asr-adaptation} and \ref{sec:dialect-id} report adaptation and micro-dialect identification results that make this intended use concrete. We therefore view \cardamom{} primarily as a low-resource diagnostic benchmark for micro-dialectal robustness rather than a training-scale ASR corpus.

\section*{Ethics and Data Statement}

\cardamom{} is constructed from publicly available YouTube videos containing public online speech and is intended for research on Arabic speech technology, especially ASR and micro-dialectal robustness. The dataset includes transcriptions, micro-dialect labels, code-switching labels, timestamps, source identifiers, and utterance-level perceived gender for aggregate analysis only. Gender labels are based on annotators' perception from the speech signal and should not be interpreted as speakers' self-identified gender or used for individual-level inference. Users should respect the terms of use of the original platforms and should not use the dataset for speaker identification, surveillance, biometric profiling, or inference of sensitive personal attributes.

\textbf{Utterance-level, non-speaker-linked annotation.} \cardamom{} does not release persistent speaker identities or aggregated speaker profiles, and we do not conduct speaker-recognition experiments. For auditing only, Appendix~\ref{appendix:speaker-clustering} uses temporary, video-local speaker-embedding clusters to estimate speaker recurrence and split leakage; these clusters are automatic estimates, not verified identities, and are not released. Perceived gender is annotated per utterance, exactly like the micro-dialect label, and is never linked across utterances to a speaker identity. Our labels describe properties of an utterance (what dialect is spoken in it, what the perceived speaker sounds like in terms of gender), not properties of a person or their location: a micro-dialect label reflects the variety spoken in a given utterance, not where the speaker currently lives, was born, or can be found. A speaker of Jordanian Badawi, for instance, may live anywhere, and our annotation makes no claim otherwise. This distinction is intended to head off any reading of \cardamom{} as enabling speaker-level geographic inference.

\textbf{Release format and license.} We release \cardamom{} as standoff annotations rather than redistributed audio. For each utterance, we release the source YouTube video ID together with our own human-produced annotations, transcription, micro-dialect label(s), perceived gender, code-switching status, and start/end timestamps, along with code to reconstruct the corpus by fetching the corresponding audio segments from the original YouTube IDs. We do not redistribute audio, source video content, or any automatically generated transcription; every transcript and label released is our own human annotation. This follows the standoff-release approach used by other YouTube-sourced speech corpora, including the MGB Challenge series~\cite{ali2017speech,ali2019mgb}, GigaSpeech~\cite{chen2021gigaspeech}, VoxLingua107~\cite{valk2021voxlingua107}, and Casablanca~\cite{talafha2024casablanca}. Public availability of the source videos is the basis for access during annotation; it does not by itself govern what we redistribute, and removal rights for any source video remain with the original uploader and with YouTube at all times. The team-created annotations and metadata (transcripts, labels, and timestamps are released under the Creative Commons Attribution--NonCommercial 4.0 International License (CC BY-NC 4.0). This license does not apply to the source audio or video, which remain governed by their rights holders and the platform's terms. The annotations, guidelines, and reconstruction code are available at \url{https://github.com/cardamom-paper/Cardamom-data}.


\bibliography{mybib,custom}
\clearpage
\appendix
\section*{Appendix}
\label{sec:appendix}

\begin{itemize}
    \item Related Work: \S\ref{appendix:related-work}
    \item Micro-Dialects Overview: \S\ref{appendix:microdialects-overview}
    \item Micro-Dialectal Overlap: \S\ref{appendix:microdialectal-overlap}
    \item Speaker-Controlled Overlap Analysis: \S\ref{appendix:speaker-clustering}
    \item Annotation Schema: \S\ref{sec:annotation-schema}
    \item Representative Multi-label Dialect Segments: \S\ref{appendix:multi-label-examples}
    \item Code-switching Examples: \S\ref{appendix:codeswitch-examples}
    \item Train/Development/Test Splits: \S\ref{sec:data-splits}
    \item Text Preprocessing: \S\ref{appendix:preprocessing}
    \item ASR Adaptation Details: \S\ref{sec:asr-adaptation-details}
    \item Micro-Dialect Identification Details: \S\ref{sec:dialect-id-appendix}
    \item Dialectness Result: \S\ref{app:dialectness}
\end{itemize}

\section{Related Work} \label{appendix:related-work}
\textbf{Arabic Speech and ASR Resources.}
Arabic ASR research has historically relied on large broadcast and web-derived corpora. Early programs such as NIST EARS \cite{pallett2003look} and DARPA GALE \cite{soltau2009advances} provided large-scale Arabic speech data, but were not designed around systematic fine-grained dialectal annotation. The MGB Challenge series \cite{ali2016mgb, ali2017speech, ali2019mgb} played a central role in Arabic ASR benchmarking: MGB-2 introduced a 1,200-hour Arabic multi-dialect broadcast recognition benchmark, MGB-3 shifted attention toward Egyptian Arabic YouTube speech in more realistic ``in-the-wild'' conditions, and MGB-5 expanded dialect identification to 17 Arabic-speaking countries while also targeting Moroccan Arabic speech recognition. QASR \cite{mubarak2021qasr} later provided a 2,000-hour multi-dialect Al Jazeera speech corpus with lightly supervised transcriptions, segmentation, punctuation, and speaker information. More recent large-scale Arabic speech resources, such as MASC \cite{10022652}, Aswat \cite{alkanhal-etal-2023-aswat}, and SADA \cite{alharbi2024sada}, further increased the amount and diversity of Arabic speech available for ASR training and evaluation. These resources substantially advanced Arabic ASR, but they were primarily designed for broadcast, web, or country-level dialect settings rather than fine-grained micro-dialectal modeling.

\textbf{Dialectal and Multidialectal Arabic ASR.}
Recent work has increasingly emphasized dialectal coverage and multidialectal supervision. Casablanca \cite{talafha2024casablanca} is especially relevant, as it introduces a fully supervised multidialectal Arabic speech dataset covering eight Arabic dialects and provides annotations for transcription, dialect, gender, and code-switching. NADI 2025 \cite{talafha2025context} further establishes spoken Arabic dialect processing as a shared-task setting, with subtasks for spoken dialect identification, multidialectal ASR, and diacritic restoration. ADI-20 \cite{elleuch2025adi} extends spoken Arabic dialect identification to 19 Arabic dialects in addition to MSA, providing large-scale supervision for dialect classification. Smaller but more focused resources, such as TuniFra \cite{choux-etal-2025-tunifra}, also highlight the growing need for carefully transcribed speech corpora targeting individual low-resource Arabic varieties. Together, these works show that the field is moving toward dialect-aware Arabic speech processing; however, most existing resources still operate at broad dialectal, regional, or country-level granularity.

\textbf{Fine-Grained Dialect and Micro-Dialect Modeling.}
The motivation for micro-dialectal Arabic speech resources is further supported by prior work in Arabic dialect identification. Text-based studies have shown that Arabic dialect variation can be modeled at a much finer level than national dialect categories. \cite{abdul-mageed-etal-2018-tweet} introduced a city-level dataset of Arabic dialects, demonstrating that locality-level variation is computationally meaningful. Later work proposed Micro-Dialect Identification (MDI) in diaglossic and code-switched environments \cite{abdul-mageed-etal-2020-toward}, further showing that fine-grained Arabic dialect modeling is an important problem in Arabic NLP. In speech, however, this level of granularity remains underrepresented. Existing spoken resources such as MGB-5 \cite{ali2019mgb}, Casablanca \cite{talafha2024casablanca}, NADI 2025 \cite{talafha2025context}, and ADI-20 \cite{elleuch2025adi} substantially improve dialectal coverage, but their labels are still primarily organized around countries or broad regional varieties rather than micro-dialects. This creates a gap between fine-grained dialect modeling in Arabic NLP and available supervision for Arabic ASR.

\textbf{Code-Switching, Orthography, and Transcription Challenges.}
Arabic speech recognition is also complicated by code-switching, mixed orthographies, and the lack of standardized spelling conventions for dialectal Arabic. Code-switching corpora such as Mixat \cite{al-ali-aldarmaki-2024-mixat} and ZAEBUC-Spoken \cite{hamed-etal-2024-zaebuc} address bilingual Arabic-English speech, while Casablanca \cite{talafha2024casablanca} includes code-switching annotations within a multidialectal Arabic ASR resource. CAFE \cite{lachemat2025cafe} further reflects the need to model spontaneous code-switching in North African Arabic contexts. These resources are important because dialectal Arabic ASR systems must often handle not only dialectal phonology and lexicon, but also English or French insertions, MSA alternation, and inconsistent orthographic choices. Evaluation work such as PolyWER \cite{kadaoui-etal-2024-polywer} shows that standard WER can be overly strict for code-switched speech, since valid transcriptions may differ by script, transliteration, or language-mixing convention. Similarly, work on Arabic ASR diacritization \cite{aldarmaki2023diacritic} highlights the importance of explicit transcription and evaluation choices in Arabic speech processing. These challenges motivate transcription guidelines that preserve spoken dialectal forms while making evaluation consistent across dialectal and code-switched segments.

\textbf{Modeling and Benchmarking.}
Recent modeling work has focused on building Arabic ASR systems that are more robust to dialectal diversity. Large multilingual models such as Whisper \cite{radford2023robust}, MMS \cite{pratap2024scaling}, and SeamlessM4T \cite{seamless2025joint} have become common baselines for multilingual and low-resource ASR. However, Arabic-specific studies show that these models still struggle under dialectal variation. \cite{talafha2023n} benchmarked Whisper on diverse Arabic speech under zero-shot, few-shot, and full-shot settings, showing that performance degrades for unseen dialects. More recent work on Whisper fine-tuning for multidialectal Arabic ASR \cite{ozyilmaz2025overcoming} explores dialect-specific and dialect-pooled adaptation strategies under data scarcity. Arabic-centered modeling efforts such as VoxArabica \cite{waheed2023voxarabica} combine dialect identification with ASR, with Habibi introducing an open-source unified-dialectal Arabic TTS framework and benchmark built partly by repurposing ASR corpora \cite{chen2026habibi}, while ArTST \cite{toyin-etal-2023-artst} introduces an Arabic text-and-speech transformer for Arabic speech technology and ArVoice \cite{toyin2025arvoice} further reflect growing interest in Arabic speech modeling, although ArVoice targets MSA speech synthesis rather than dialectal ASR. Work on dialectal coverage and generalization \cite{djanibekov-etal-2025-dialectal} further shows that robust Arabic ASR requires explicit coverage of MSA, dialectal speech, and code-switching, and that systems continue to struggle with data imbalance and unseen dialects. Evaluation has also become more systematic through the Open Universal Arabic ASR Leaderboard \cite{wang2024open}, which provides a common benchmark for open-source multidialectal Arabic ASR models.

Despite these advances, most Arabic speech resources still treat dialects as broad regional or national categories. This overlooks micro-level variation within the same country or macro-dialect, including localized differences in pronunciation, lexical choice, morphology, prosody, and code-switching behavior.  \cardamom{} addresses this gap by introducing a micro-level multidialectal Arabic speech resource with 40 hours of transcribed speech and high-granularity labels for 21 micro-dialects. To the best of our knowledge, \cardamom{} is the first Arabic ASR resource designed explicitly around micro-dialect-level supervision across multiple localized Arabic varieties. By enabling evaluation of both within-dialect variation and cross-micro-dialect generalization, \cardamom{} complements existing broad-coverage resources by targeting a level of dialectal granularity that current Arabic ASR benchmarks do not capture.

\section{\texorpdfstring{Micro-dialects Overview}{Micro-dialects Overview}}\label{appendix:microdialects-overview}

\subsection{Operational Status of the Taxonomy}
The micro-dialect labels serve as operational annotation categories used to organize the dataset and structure evaluation, informed by prior dialectological literature and refined through the knowledge and judgments of native-speaker annotators. Our categorization is intended to reflect real-world linguistic communities as recognized by native speakers, so that the data represent the local varieties associated with these labels; although Table~\ref{tab:cross_dialect_phonology} and this appendix document clear linguistic differences among several of these micro-dialects, we do not claim that the proposed categories constitute a definitive linguistic classification, since we have not conducted the kind of comprehensive linguistic analysis required to establish theoretical boundaries among these varieties or to characterize their precise status within broader levels of language variation. Linguists themselves do not always agree on a single correct granularity, and varieties we treat as single categories can be further subdivided: Jordanian Falahi is a clear example, as our own cited sources on Jordanian dialectology \cite{Herin2010,AlWerHerin2011,HerinAlWer2013,AlWerEtAl2015} study individual localities such as Salt at a finer level than our three-way Madani/Falahi/Badawi split, and other regional surveys further divide rural Jordanian speech into named town- or tribe-specific varieties (e.g., Salti, Bani Hassan, Karaki, Tafilawi) rather than treating it as one category. We chose a level of granularity that is recognizable to annotators, sufficiently attested in the speech we collected, and useful for evaluating ASR systems beyond country-level labels, while acknowledging that finer subdivisions remain possible and are supported by prior work.

\subsection{Cross-Dialect Phonological Variation}
Table~\ref{tab:cross_dialect_phonology} summarizes representative cross-variety phonological patterns referenced in \S5.1 and throughout this appendix. Each row identifies an MSA phoneme by its IPA symbol and Arabic letter; the second column lists its attested realizations in specific Egyptian, Jordanian, Palestinian, Saudi, Lebanese, and Mauritanian varieties, with lexical examples illustrating the correspondence and arrows indicating the change from the MSA form to the variety-specific form.

\begin{table}[!t]
\centering
\small
\renewcommand{\arraystretch}{1.12}
\setlength{\tabcolsep}{3.5pt}

\begin{adjustbox}{width=\columnwidth, max totalheight=0.82\textheight, keepaspectratio}

\begin{tabular}{@{} p{0.23\columnwidth} >{\RaggedRight\arraybackslash}p{\columnwidth} @{}}

\toprule
\textbf{MSA Phoneme} & \textbf{Variety-specific realizations \& lexical example} \\
\midrule

Uvular Stop /\textipa{q}/ (\<ق>) &
\EGY\ Cairene: realized as a glottal stop [\textipa{?}] (e.g., \<قلب> $\rightarrow$ \<ألب>).
\JOR\ \PAL\ Madani: commonly [\textipa{?}] as well.
\EGY\ Saidi: shifts to [\textipa{g}].
\PAL\ North Rural: shifts to [\textipa{k}] (e.g., \<كلب>).
\PAL\ South Rural: shifts to the affricate [\textipa{\t{tS}}] (e.g., \<چلب>).
\cite{holes2004modern,watson2002phonology,AlWerHerin2011,Youssef2021}
\LEB\ Chamali and Jnoubi (sedentary): collapse to [\textipa{?}] (e.g., \<قلب> $\rightarrow$ \<ألب>).
\LEB\ B'albaki: retains [\textipa{q}] under a Bedouin substrate (e.g., \<قلب>), with Zahle optionally aligning with the sedentary [\textipa{?}].
\cite{agbaht2023akkar,lentin2018}
\MAU\ Hassaniya: shifts to [\textipa{q}] represented by \<گ> (e.g., \<قلب> $\rightarrow$ \<گلب>).
\\
Interdental /\textipa{T}/ (\<ث>) &
\KSA\ Najdi and \JOR\ Falahi: preserved as [\textipa{T}].
\KSA\ Hijazi and \JOR\ \PAL\ Madani: often merge to the alveolar stop [\textipa{t}] (e.g., \<ثلاثة> $\rightarrow$ \<تلاتة>).
\EGY\ Egyptian varieties: may merge to the sibilant [\textipa{s}] (e.g., \<ثم> $\rightarrow$ \<سم>).
\cite{holes2004modern,AlWer2004Interdentals,Youssef2021}
\LEB\ Chamali, Jnoubi, and Zahle B'albaki: merge to the alveolar stop [\textipa{t}] (e.g., \<ثلاثة> $\rightarrow$ \<تلاتة>).

\LEB\ Rural Akkari and Hermeli B'albaki: partial retention as [\textipa{T}].
\\

Interdental /\textipa{D}/ (\<ذ>) &
\KSA\ Northern and Southern KSA: preserved as [\textipa{D}].
\KSA\ Hijazi, \EGY\ Cairene, and \PAL\ Madani: shift to [\textipa{d}] (e.g., \<ذهب> $\rightarrow$ \<دهب>).
\PAL\ Falahi and Badawi: shifts to the emphatic [\textipa{D\super Q}] (e.g., \<هذا> $\rightarrow$ \<هاظا>).
\cite{holes2004modern,AlWer2004Interdentals,Youssef2021}
\LEB\ merges to [\textipa{d}] (e.g., \<هذا> $\rightarrow$ \<هاد>, with \<هيدا> as the dominant demonstrative).
\LEB\ Hermeli B'albaki: optional retention as [\textipa{D}].
\\

Velar Stop /\textipa{k}/ (\<ك>) &
\JOR\ Falahi and \PAL\ North Rural: undergo palatalization to [\textipa{\t{tS}}]
(e.g., \<كيف> $\rightarrow$ \<تشيف> [\textipa{\t{tS}i:f}]).
\EGY\ Egyptian and \JOR\ \PAL\ Madani: typically retain [\textipa{k}].
\cite{DickinsWatson2001,Youssef2021}

\LEB\ All Lebanese micro-dialects: retain [\textipa{k}] uniformly, with no kashkasha-style affrication anywhere in the country (e.g., \<كذاب> stays \<كذاب>).
\\

Palato-alveolar /\textipa{\t{dZ}}/ (\<ج>) &
\EGY\ Cairene and Port-Saidi: de-affricate to a voiced velar stop [\textipa{g}] (e.g., \<جميل> [\textipa{g}ami:l]).
\EGY\ Saidi: can shift to an alveolar stop [\textipa{d}] in specific lexical environments (e.g., \<الجيش> $\rightarrow$ \<الديش>).
\cite{watson2002phonology,holes2004modern,Youssef2021}

\LEB\ Chamali, Jnoubi, and Zahle B'albaki: de-affricate to [\textipa{Z}] (e.g., \<جبل> [\textipa{Zabal}]).
\LEB\ Rural Akkari and Hermeli B'albaki: variably retain [\textipa{\t{dZ}}].
\cite{lentin2018}
\MAU\ Sharg: shifts to [\textipa{S}] in \<المجتمع> $\rightarrow$ \<المشتمع>.
\\

Emphatic Coronal /\textipa{t\super Q}/ (\<ط>) &
Emphasis/pharyngealization is typically retained as [\textipa{t\super Q}]; in \EGY\ Saidi, it may lose emphasis to [\textipa{t}]
(e.g., \<طار> $\rightarrow$ \<تار> [\textipa{t}a:r]).
\cite{watson2002phonology,Davis1995} \\

\bottomrule
\end{tabular}
\end{adjustbox}
\caption{Key cross-variety phonological variation across Egyptian, Jordanian, Palestinian, Saudi, Lebanese, and Mauritanian varieties compared to MSA.}
\label{tab:cross_dialect_phonology}

\end{table}

\subsection{Country and Micro-dialect Coverage}
Arabic dialects exhibit substantial regional and social variation, and broad national labels often obscure the localized varieties that speakers use in everyday communication~\cite{JZHNW23, holes2004modern, prochazka2021arabic, zaidan2014arabic}. For this reason, \cardamom{} is designed to represent fine-grained micro-dialectal variation  across \textit{Egypt}, \textit{Jordan}, \textit{Lebanon}, \textit{Mauritania}, \textit{Palestine}, and  \textit{Saudi Arabia}. We use \textit{micro-dialect} as an operational label for localized sub-country varieties that are recognizable to speakers and reflected in systematic differences in pronunciation, morphology, lexical choice, prosody, and grammatical patterns \cite{holes2004modern, prochazka2021arabic}. Capturing this level of variation is important for evaluating Arabic speech technologies beyond coarse country-level labels, which can mask substantial variation within the same national dialect category~\cite{salameh-etal-2018-fine, abdul-mageed-etal-2020-toward}.

Our micro-dialect inventory is grounded in prior dialectological work, but it is not intended to reproduce any single taxonomy mechanically. Instead, we use existing descriptions as a linguistic foundation and adapt them into annotation categories that are practical for speech data collection, recognizable to native-speaker annotators, and sufficiently consistent for computational modeling. This design choice reflects the fact that dialect boundaries are often gradual rather than discrete, and that speech communities may share features across regional, social, and national boundaries.

In Egypt, \cardamom{} represents three practical micro-dialect groups: \textit{Cairene}, \textit{Coastal (Alexandrian and Port Saidi)}, and \textit{Saidi (Upper Egyptian)}. This grouping reflects Egypt’s regional diversity, where pronunciation, lexicon, rhythm, and expression vary across regions, even though Cairo Arabic is widely understood through media exposure~\cite{fashwan2025computational, abdel1979comprehensive,el1987egyptian, gadalla2000comparative}. This grouping is also consistent with broader atlas-based descriptions of Egyptian dialect areas, including the Delta, Nile Valley, Upper Egypt, Western Oases, and Bedouin varieties \cite{woidich1996rural,miller2005between}.

In Jordan, \cardamom{} follows the widely used distinction between \textit{Madani}, \textit{Falahi}, and \textit{Badawi} varieties. These categories capture major points along Jordan's urban, rural, and Bedouin dialect continuum, including the role of the Ammani urban koiné, sedentary rural dialects, and Bedouin speech varieties~\cite{Palva1984,AlWerEtAl2015,AlWer2007,Palva2008,HerinEtAl2022}.

In Lebanon, \cardamom{} includes three practical micro-dialect groups: \textit{Chamali (Northern)}, \textit{Jnoubi (Southern)}, and \textit{B'albaki (Bekaa/Baalbek)}. These categories reflect localized variations relevant to the collected speech data while remaining practical for native-speaker annotation and downstream modeling. They are grounded in prior descriptions of Lebanese Arabic, where regional speech differs across northern, southern, Bekaa, Beiruti, and Mount Lebanon areas, alongside a media-driven dominant Lebanese variety associated with Beirut and central Mount Lebanon \cite{EJHZ22,lentin2018,germanos2009variation,stone2008rahbani}. 
Chamali captures northern varieties such as Tripoli, Zgharta, and Akkar speech \cite{joukhadar2023zgharta,agbaht2023akkar}, while Jnoubi reflects southern coastal and inland varieties, and B'albaki represents the Bekaa/Baalbek-Hermel area \cite{EJHZ22,lentin2018,germanos2009variation}.

In \textit{Mauritania}, \cardamom{} adopt three practical \textit{\d{H}ass\=aniyya} micro-dialect groups: \textit{Western (G\textschwa bla)}, \textit{Northern (S\=a\d{h}il)}, and \textit{Southern/Eastern (Sharg)}. This grouping reflects Mauritania’s regional and sociolinguistic diversity, where \textit{\d{H}ass\=aniyya} is the dominant spoken Arabic variety but coexists with Pulaar, Sonink\'e, Wolof, Zenaga, French, and other languages across different regions and urban settings \cite{naciriazzouz2023overview,taine2020hassaniyya,taine2018hassaniyya}. As with the other countries, the goal is not to impose a rigid taxonomy, but to define annotation categories that preserve salient localized variation in the available speech data, especially differences in pronunciation, lexicon, and usage patterns \cite{francis1979mauritanian,hanchey1979mauritanian,tainecheikh:halshs-00563853}.

In Palestine, we adopt a three-way classification consisting of \textit{Madani urban}, \textit{Northern Fallahi}, and \textit{Southern Badawi} speech. This structure avoids reducing Palestinian Arabic to a simple city--village contrast and instead captures broader bundles of phonological, lexical, and grammatical features associated with urban, northern rural, and southern rural or Bedouin-adjacent varieties~\cite{EJHZ22,ulbah2024adverb,AlWerHerin2011}.

In Saudi Arabia, \cardamom{} covers major regional dialect groups, including \textit{Najdi}, \textit{Hijazi}, \textit{Eastern}, \textit{Northern}, and \textit{Southern} varieties. These categories reflect broad divisions commonly used in sociolinguistic and phonetic research and capture variation across central, western, eastern, northern, and southern Saudi speech communities~\cite{ingham1994najdi,Prochazka1988,sambas2016hijazi,asiri2009,alqahtani2015tihami,north_Dialect}. Given the size and internal diversity of Saudi Arabia, these labels should be understood as high-level micro-dialectal groupings rather than exhaustive representations of all local speech varieties.



Our final inventory is grounded in this prior work, but adapted to the practical requirements of dataset curation: the categories must be meaningful to native speaker annotators, applicable to naturally occurring speech, and useful for evaluating model behavior beyond country-level labels.  
This allows \cardamom{} to preserve fine-grained regional and social variation that would otherwise collapse under broad national dialect categories.

\subsection{Phonological Variation in Arabic: A Micro-Dialectal Analysis}
\label{app:phonology}

Spoken Arabic is characterized by a high degree of phonological variation \cite{watson2002phonology,khan2011semitic}. An analysis of phonetic data across various micro-dialects reveals systematic axes of phonological divergence from MSA. The primary differences can be observed as follows:


\textbf{The Reflexes of the Uvular Stop /\textipa{q}/ (\<ق>)}.
The voiceless uvular stop /\textipa{q}/ is one of the most salient
sociolinguistic variables in Arabic dialectology
\cite{watson2002phonology,holes2004modern}. The data show a shift from
/\textipa{q}/ to a glottal stop [\textipa{?}] in \textit{Cairene} and
\textit{Port-Saidi}, as well as in the \textit{Madani} dialects of both
Jordan and Palestine. In these micro-dialects, MSA /\textipa{q}alb/
(\Ar{قلب}, `heart') is realized as [\textipa{?}alb] (\Ar{ألب}).
Alexandrian shows a mixed pattern in the reported examples, retaining
/\textipa{q}/ in \Ar{قلب} while showing glottalization in
\Ar{تقبل} $\rightarrow$ \Ar{تأبل}. Other micro-dialects exhibit different
reflexes. \textit{Saidi} realizes \Ar{قلب} as [\textipa{g}alb]
(\Ar{جلب}), while the Mauritanian Hassaniya varieties realize qaf with a
voiced velar stop reflex, represented orthographically as \Ar{گ}, as in
\Ar{قلب} $\rightarrow$ \Ar{گلب}. In the Palestinian data, North Rural
\textit{Falahi} shifts /\textipa{q}/ to [\textipa{k}]
(\Ar{كلب}), while South Rural \textit{Badawi} shifts it to the affricate
[\textipa{\t{tS}}] (\Ar{چلب}) \cite{AlWerHerin2011,Youssef2021}. The
Jordanian \textit{Falahi} and \textit{Badawi} micro-dialects retain qaf in
the reported examples \cite{AlWer2003}. Lebanese \textit{Chamali} and
\textit{Jnoubi} show glottalization, while \textit{B'albaki} varies between
retention and glottalization \cite{agbaht2023akkar,lentin2018}.

\vspace{0mm}
\noindent
\textbf{Interdental Fricatives /\textipa{T}/ (\<ث>), /\textipa{D}/ (\<ذ>), and /\textipa{D\super Q}/ (\<ظ>).}
MSA maintains the interdental fricatives, a feature that is preserved in some
micro-dialects but merged with stops, sibilants, or other reflexes in others
\cite{watson2002phonology,al2022arabic,AlWer2004Interdentals}. The
interdental articulation is retained in the \textit{Southern},
\textit{Northern}, \textit{Najdi}, and \textit{Eastern} dialects of Saudi
Arabia, as well as in the \textit{Falahi} and \textit{Badawi} dialects of
Jordan and the North Rural \textit{Falahi} and South Rural \textit{Badawi}
dialects of Palestine. For example, MSA /\textipa{T}ala:Ta/
(\Ar{ثلاثة}, `three') and /\textipa{ha:Da}/ (\Ar{هذا}, `this') remain
relatively unchanged in these varieties. Hassaniya also preserves interdental
contrasts in many forms, although regional and lexical variation is attested
\cite{taine2018hassaniyya,tainecheikh:halshs-00563853}.

In contrast, interdentals merge with alveolar stops or alveolar sibilants in
\textit{Cairene}, \textit{Port-Saidi}, \textit{Alexandrian},
\textit{Saidi}, \textit{Hijazi}, and the \textit{Madani} dialects of Jordan
and Palestine. For instance, /\textipa{D}/ merges with [\textipa{d}] in
\textit{Hijazi}, \textit{Cairene}, and \textit{Madani} varieties, as in
\Ar{ذهب} $\rightarrow$ \Ar{دهب}, while /\textipa{T}/ merges with the
sibilant [\textipa{s}] in the Egyptian varieties, as in
\Ar{ثم} $\rightarrow$ \Ar{سم}. In \textit{Hijazi} and the
\textit{Madani} dialects of Jordan and Palestine, /\textipa{T}/ commonly
shifts to [\textipa{t}], as in \Ar{ثلاثة} $\rightarrow$ \Ar{تلاتة}. In
Lebanese \textit{Chamali} and \textit{Jnoubi}, /\textipa{T}/ and
/\textipa{D}/ also show stop or sibilant reflexes, while \textit{B'albaki}
shows more variable retention and merger \cite{lentin2018}.

\vspace{0mm}
\noindent
\textbf{Velar Palatalization.}
Another distinguishing feature in Arabic dialectology is the conditional
palatalization of the voiceless velar stop /\textipa{k}/ (\<ك>) to the
palato-alveolar affricate [\textipa{\t{tS}}], a process often referred to
in traditional Arabic philology as kashkasha
\cite{cantineau1960cours,holes2004modern,DickinsWatson2001}. This shift is
active in the Jordanian \textit{Falahi} and \textit{Badawi} micro-dialects,
as well as the Palestinian North Rural \textit{Falahi} and South Rural
\textit{Badawi} micro-dialects. In these varieties, MSA /\textipa{kajfa}/
(\Ar{كيف}, `how') is realized as [\textipa{\t{tS}i:f}]
(\Ar{تشيف}). In the Palestinian data, \Ar{كذاب} becomes
\Ar{تشذاب} in North Rural \textit{Falahi}, while South Rural
\textit{Badawi} retains \Ar{كذاب}. This indicates that palatalization is
micro-dialect-specific and may also be lexically conditioned. This contrasts
with \textit{Cairene}, \textit{Port-Saidi}, \textit{Alexandrian},
\textit{Saidi}, Lebanese varieties, and the \textit{Madani} dialects of
Jordan and Palestine, which retain [\textipa{k}] in the corresponding
examples.

The Mauritanian Hassaniya varieties add further patterns not captured in the
earlier four-country comparison. Besides the realization of /q/ as [\textipa{g}],
they show variation involving \Ar{غ} /ghayn /\textipa{Gh}/, as in
\Ar{لغة} $\rightarrow$ \Ar{لقة} in the Northern variety and
\Ar{صغير} $\rightarrow$ \Ar{سقير} in Northern and Southern/Eastern.
The same examples show de-emphasis of /\textipa{s\super Q}/ to
/\textipa{s}/, as in \Ar{صغير} $\rightarrow$ \Ar{سقير}/\Ar{سغير}.
Additional examples include initial hamza reduction
(\Ar{الأول} $\rightarrow$ \Ar{لول}) and a Western/Gabla emphatic reflex in
\Ar{التراب} $\rightarrow$ \Ar{اطراب}. These patterns are consistent with
descriptions of Hassaniya as a regional continuum with both shared and
localized phonological features \cite{taine2018hassaniyya,taine2020hassaniyya,francis1979mauritanian}.
Mapping the specific micro-dialectal data against MSA highlights distinct, systematic phonological rules. Whether through the lenition of /q/, the treatment of interdentals, or the presence of velar palatalization, each micro-dialect demonstrates a unique phonological profile that diverges from standard phonetic boundaries.

\subsection{Egyptian Micro Dialects}
Egyptian dialects are characterized by substantial regional diversity, reflecting geography as well as social, cultural, and historical factors (including internal and external migrations). Pronunciation, lexicon, rhythm, and styles of expression vary across regions. Modern media (cinema and television) have contributed to the spread of a widely understood variety, but it has not erased local distinctions \cite{fashwan2025computational,gadalla2000comparative}.  Within Egypt, fine-grained (“micro”) dialect work is facilitated by the fact that the country’s dialect map is described as “relatively well established” due to the \textit{Atlas of Egyptian Rural Arabic} \cite{woidich1996rural}, while earlier linguistic descriptions disproportionately focused on Cairo Arabic and provided comparatively scarcer coverage of regional varieties \cite{serreli2019perceptual}. Accordingly, microdialectal analysis in Egypt is best framed as \textit{(i)} adopting the atlas-based geographic skeleton and \textit{(ii)} grounding micro-variation in diagnostic isoglosses at multiple linguistic levels. 

Following the atlas-based perspective, Egyptian Arabic can be organized by first distinguishing the dialects of the “true Bedouins,” described as groups who arrived relatively recently and live on Egypt’s margins (e.g., Sinai and the Western/Eastern deserts) and are not fully sedentarized \cite{miller2005between}. The atlas then divides the “rural dialects” (sedentary rural population) into three main geographic areas—\textit{Nile Delta}, \textit{Nile Valley}, and \textit{Western Oases}—each with subgroups. In the \textit{Nile Delta}, the subgroups include \textit{Western Dialects (WD 1, 2, 3, 4)}, \textit{North-Eastern Dialects}, \textit{Central Dialects}, and \textit{North Eastern Dialects (NED 1 and 2}. In the \textit{Nile Valley}, the atlas distinguishes \textit{Middle Egypt dialects} (\textit{NME 1–2, SME}) and \textit{Upper Egypt} dialects (\textit{UE 1, 2, 3, 4}). \textit{Western Oases} dialects include four oasis varieties: \textit{al-Bahariyya, al-Farafra, ad-Dakhla, al-Kharga}. 

Across these regions, reported diagnostic differences include phonological and morphological isoglosses \cite{miller2005between,woidich1996rural}. A widely used macro-isogloss pair is: Cairo / central Delta: \textipa{/j/}~\(\rightarrow\)~\textipa{/g/} and \textipa{/q/}~\(\rightarrow\)~\textipa{/?/} (e.g., \textit{gamal} ‘camel’ \Ar{جمل}, \textipa{?alb} ‘heart’ \Ar{ألب}); versus Sinai Bedouin: \textipa{/j/}~\(\rightarrow\)~\textipa{/dZ/} or \textipa{/Z/} and \textipa{/q/}~\(\rightarrow\)~\textipa{/g/} (e.g., \textipa{dZamal} \Ar{چمل}, \textit{galb} \Ar{جلب}). Another diagnostic concerns the interdentals (\textipa{T}, \textipa{D}): Sinai Bedouin is described as retaining interdentals (e.g., \textipa{Tala:Ta} ‘three’ \Ar{ثلاثة}, \textipa{De:l} ‘tail’ \Ar{ذيل}), whereas Central Delta merges them to \textipa{/t/} and \textipa{/d/} (e.g., \textit{tala:ta} \Ar{تلاتة}, \textit{de:l} \Ar{ديل}). Gender distinctions in plural verbal agreement are reported as retained in some areas but lost/neutralized in others (e.g., Sinai retains masc./fem. plural contrast, whereas central Delta patterns are described as neutralizing it). Morphologically, a diagnostic reported for Qena and farther south, and also in Bahariyya and Farafra \cite{mohamed2024distinctive}, is the “Western imperfect paradigm.” The atlas also points to internal microvariation within Upper Egypt (e.g., UE1 vs.\ UE2). 

In our \cardamom{} construction from YouTube speech, we operationalize a complementary, data-driven grouping that aligns with common high-level regional labels while preserving micro-dialectal coverage in our sampling. For Egypt, we group the collected speech into three practical categories: \textit{Cairene} (Cairo), \textit{Coastal/Delta urban varieties} (including Alexandrian and Port Saidi speech), and \textit{Saidi} (Upper Egyptian). This grouping is not intended to replace the more detailed atlas-based classifications of Egyptian Arabic, but rather to provide annotation categories that are recognizable to native speakers and suitable for computational modeling. Each category still exhibits internal variation shaped by city--village differences, regional history, and patterns of contact.
\subsubsection{Cairene Micro-dialect}
The Cairo dialect is widely used and prominent, corresponding to Cairo's political, cultural, and media importance; cinema and television have helped popularize and increase comprehension of this dialect \cite{abdel1979comprehensive, el1987egyptian, gadalla2000comparative}. It is described as flexible and easy to understand, with a strong ability to absorb new vocabulary from local dialects and other languages, making it a common variety for communication among speakers from different regions.
\subsubsection{Saidi Micro-dialect}
Saidi Egyptian refers to dialects spoken in southern Egypt, particularly in Beni Suef and extending through Minya, Asyut, Sohag, Qena, to Aswan. A key characteristic is the realization of /\Ar{ق}/, which is variable in Minya but is described as consistently realized as \Ar{ج} in more southern variants \cite{woidich1996rural}. Moving southward, vowel lengthening increases, the pronunciation of emphatic consonants strengthens, and speech patterns become slower and more stable \cite{mohamed-eida-habash-2025-beyond}. Despite differences, these dialects share a common phonological framework and are described as showing internally structured regional variation rather than sharply bounded linguistic divisions: from Minya to Aswan, systematic phonological variation forms a geographic continuum.

\subsubsection{Coastal Micro-dialects}
Coastal dialects include those spoken in seashore cities such as Alexandria, Port Said, Damietta, and Marsa Matrouh \cite{woidich1996rural}. They are characterized by historical and cultural influences associated with coastal cities’ openness to trade and contact. The \textit{Alexandrian micro-dialect} is described as differing from Cairene in rhythm and pronunciation and using local vocabulary related to sea life and trade; it is characterized by a fast rhythm and a relatively sharp tone with clear pronunciation (e.g., a tendency for~\mbox{\Ar{ش} and \Ar{س}} to be pronounced clearly without softening) \cite{behnstedt2018formation}. The \textit{Port Saidi micro-dialect} is described as softer in tone with a slightly slower rhythm, including slight elongation in some parts (especially in informal contexts), the use of specific local idioms, and effects from internal migrations reflecting the city’s commercial character and history as an important port.

\subsection{Jordanian Micro-dialects}
Jordanian Arabic forms a continuum of mutually intelligible dialects across the Kingdom of Jordan, with an emerging common variety alongside significant regional and social sub-varieties that differ in phonological and morphological features \cite{Palva1984,AlWerEtAl2015}. Traditional accounts divide Jordan’s vernaculars into \emph{sedentary} vs.\ \emph{Bedouin} types \cite{Palva1984}: sedentary varieties further split into rural (\textit{Falā\d{h}ī} \Ar{فلاحي}) and urban (\textit{Madani} \Ar{مدني}) dialects, while (\textit{Badawi} \Ar{بدوي}) dialects are associated with nomadic or semi-nomadic tribal origins. In recent decades, strong dialect contact in urban centers, especially Amman, has also produced a mixed koiné that functions as a prestige urban variety \cite{AlWer2007}; below, we overview the three primary micro-dialect groups: Madani, Falahi, and Badawi.
 
\subsubsection{Madani – Urban Koiné Dialect}
The urban dialect of Amman, the capital and largest city, is often described as a new \emph{koiné} that emerged from dialect mixing \cite{AlWer2007,AlWer2002b}. Amman was a small town in the early 20th century, but its designation as the capital brought major migration from northern Jordan (e.g., Salt and Irbid), southern Jordan (Karak, Ma'an), Bedouin communities, and waves of Palestinian refugees after 1948 and 1967 \cite{AlWer2007}. As a result, Ammani speech formed as a hybrid combining features from these sources, with particularly strong influence from northern Jordan (Horani) input \cite{AlWer2007,Palva2008}. Over a few decades, these initially mixed inputs converged into a stable local variety by the third generation of native Ammani speakers \cite{AlWer2007}.

The modern urban Jordanian koiné (exemplified by Ammani speech) is characterized by reduction or leveling of some marked rural or Bedouin features. For example, northern Jordanian and rural dialects may affricate /k/ to [\textipa{\t{tS}}] before front vowels (e.g., /keef/ “how” $\rightarrow$ [\textipa{\t{tS}}i:f]) \cite{Herin2010,HerinAlWer2013,AlHawamdeh2015}. Although this feature was brought to Amman by early migrants (e.g., from Salt), it rapidly receded in the capital: first-generation Ammanis already shifted to [k] in all contexts \cite{AlWer2007}. In Amman, affrication became stigmatized as a rural marker, though it continues in cities such as Irbid and Ramtha, especially in semi-urban and rural--urban transitional communities \cite{HerinAlWer2013}.

Another case of leveling concerns the second-person plural system. Traditional sedentary Jordanian (\textit{Falā\d{h}ī} \Ar{فلاحي}) varieties distinguish masculine /-ku/ vs.\ feminine /-ken/ (e.g., \textipa{laQindku} \Ar{لَعِندكو} vs.\ \textipa{laQindken} \Ar{لَعِندكَن}), and in the north /k/ may affricate, yielding forms such as \textipa{laQind\t{tS}en} \Ar{لَعِندتشن} \cite{HerinAlWer2013,herin2014dialect,Palva2008}. In Amman, the urban koiné shows functional neutralization to a single plural form /-kum/ (e.g., \textipa{laQindkum} \Ar{لَعِندكُم}) for both genders in most contexts \cite{AlWer2007}.

Phonologically, urban Jordanian shares traits with other Levantine city dialects, including variation in the reflex of \textit{qāf} (\Ar{ق}) between [\textipa{P}] and [g]. Studies report that [\textipa{P}] is favored by women while men more often use [g], and both variants are socially meaningful in Amman \cite{AlWerHerin2011,AbdelJawad1986}: [\textipa{P}] is associated with an urban or Palestinian-oriented identity, while [g] indexes a more “tribal” or East Bank identity, yielding stable sociolinguistic variation in (q).

Overall, the Madani koiné of Amman has become the prestigious urban variety of the capital and a key source of linguistic influence. At the same time, Falā\d{h}ī and Badawi dialects remain widespread and locally dominant elsewhere, and media usage varies by genre (MSA in news, colloquial in entertainment).

\subsubsection{\textit{Falā\d{h}ī} – Sedentary Rural Dialects}
The term \textit{Falā\d{h}ī} (“peasant” or rural) refers to the sedentary dialects spoken in Jordan’s villages and smaller towns by non-nomadic communities. These rural dialects are part of the broader Southern Levantine Arabic continuum and share many features with Palestinian rural speech. In northern and central Jordan (historically \textit{\d{H}awrān}, from Amman up to southern Syria), the local rural vernacular is often called the Horani dialect (\textit{\d{h}ōranī} \Ar{حوراني}) \cite{Palva1984}. Horani varieties are characterized by classic Levantine rural features, including /q/ realized as [g] (e.g., \textit{qalb} $\rightarrow$ \textit{galb}) and conditional palatalization (affrication) of /k/ to [t\textipa{S}] before front vowels (e.g., \textit{kīf} $\rightarrow$ \textit{chīf}) \cite{Herin2010,HerinAlWer2013,AlHawamdeh2015}. These features are widespread in northwest Jordan (Irbid, Ajloun, Balqa) and are well documented, though they are often stigmatized, and younger rural speakers frequently adopt urbanized variants in casual speech \cite{AlWer2007}.

Beyond phonetics, \textit{Falā\d{h}ī} dialects preserve older grammatical distinctions, including gender-specific forms in plural addresses and verb conjugations \cite{AlWer2007}, and they use the proclitic \textit{b-} to mark present-tense verbs (e.g., \textit{b-niktub} \Ar{بنكتب} “we are writing”), unlike some Bedouin dialects that lack \textit{b-} \cite{AlWerHerin2011}. Lexically, rural Jordanian Arabic overlaps substantially with neighboring Palestinian dialects, alongside localized vocabulary tied to agriculture and village life. Sedentary dialects are not monolithic: there is notable sub-regional variation (e.g., far north vs.\ the Karak plateau), including regional realizations of the feminine ending “-a” (tā\textsuperscript{'} marbū\d{t}a), raised to [e] in southern rural Jordan (\textit{maktabe} \Ar{مكتبي}) and often realized as a more central [æ] in northern dialects (\textit{maktabeh} \Ar{مكتبا}), in contrast to Bedouin [a] \cite{Palva2008}. Nevertheless, \textit{Falā\d{h}ī} varieties remain mutually intelligible within the broader sedentary Levantine type, and many rural speakers are bidialectal, using the local dialect at home but shifting towards urban speech in formal settings or when interacting with outsiders.

\subsubsection{Badawi – Bedouin Dialects}
The Bedouin dialects of Jordan are associated with Bedouin tribes, traditionally pastoral nomads of the eastern deserts and southern steppe. Linguistically, they fall under Arabian Bedouin Arabic and are often classified as “Northwest Arabian” Arabic \cite{Palva2008}. In Palva’s classification, the Bedouin-type dialects spoken in Jordan (and neighboring desert regions of Syria and Arabia) form an eastern subgroup of Northwest Arabian dialects \cite{Palva2008,HerinEtAl2022} and show strong affinities with the dialects of the Negev, Sinai, and the Najd rather than with sedentary Levantine city and village dialects \cite{Palva2008,HerinEtAl2022}. Within Jordan, Bedouin varieties further differ along tribal lines and historical origins: southern Jordan dialects (e.g., \d{H}uway\d{t}āt, Banī 'A\d{t}iyyah, the Bdul of Petra, and the 'Nmāt) share features with Sinai and Negev Bedouin \cite{Palva2008}, while major tribes in central and northern Jordan (e.g., Banī \d{S}axr, Banī \d{H}asan, Sir\d{h}ān) speak North Arabian (Najdi)–type dialects associated with the 'Anaza and Shammār confederations \cite{HerinEtAl2022}. Some north-eastern Bedouin communities (e.g., Sir\d{h}ān of the \d{H}amād) speak \textit{šāwī} Arabic, similar to Syrian–Iraqi Bedouin dialects and often distinguished by plural verb forms ending in -\textit{ūn} (e.g., \textit{yīgūlūn} \Ar{يِگولون} “they say”) \cite{HerinEtAl2022}.

Despite local differences, Jordanian Bedouin dialects share defining features. Phonetically, /q/ is typically realized as [g \Ar{گ}] (e.g., \textit{gabīl} for \textit{qabīl} “before”), and /k/ generally remains [k \Ar{ك}] (with /k/-palatalization described as rare and largely limited to certain words or older speakers). They also tend to preserve classical vowel features reduced in sedentary speech; for instance, /aw/ and /ay/ may be retained as diphthongs or only partially monophthongized, whereas urban Jordanian often realizes them as [\textipa{o:}] and [\textipa{e:}] \cite{Palva2008}. Grammatically, Bedouin varieties often lack (or optionally use) the habitual \textit{b-} prefix on present tense verbs, using forms such as \textit{aktub} or an auxiliary (e.g., \textit{gā\textipa{P}id aktub} \Ar{گاعد أكتب} “I am writing”) \cite{AlWerHerin2011}, and they preserve lexical items and idioms tied to desert life. Although many Bedouin groups have become settled and dialect contact has led to gradual convergence in some communities \cite{HerinEtAl2022}, core Bedouin features remain salient identity markers (e.g., [g] vs.\ [\textipa{P}] for /q/), and many speakers are bidialectal, using Badawi at home and shifting to the urban koine in mixed settings.

\subsection{Lebanese Micro-dialects}
Lebanese micro-dialects vary around five provincial clusters and one
media-driven dominant dialect
\citep{EJHZ22,lentin2018,germanos2009variation}. The
\textit{dominant Lebanese} is anchored around Beirut in central Mount
Lebanon and was projected nationally and pan-Arab by the Rahbani musical
theater, the Fairuz repertoire and the folkloric serials broadcast on the
major Lebanese media and TVs from the 1960s onward~\citep{stone2008rahbani}
reproduced in formal and mediated contexts. 
\textit{Beiruti} clusters around the
historical city neighborhoods (Achrafieh, Basta, Ras Beirut) and behave as an
urban prestige form that absorbs migrant inputs and re-allocates them through
contact-driven leveling
\citep{germanos2009variation}. 
\subsubsection{Northern}
The \textit{Northern} (Chamali) cluster runs from a brisk, Syrian-leaning Tripoli speech
through the Maronite-marked Zgharta variety with its non-verbal negation patterns
\citeyearpar{joukhadar2023zgharta}, into the
rural Akkar varieties~\citep{agbaht2023akkar}. 
\subsubsection{Southern}
The \textit{Southern} cluster splits between the coastal Sidon and Tyre varieties,
which align with Palestinian coastal Arabic in vowel
quality and selected morphology, and the inland Jizzine and Nabatieh villages, 
which retain more conservative traits~\citep{EJHZ22}. 
The \textit{Bekaa} divides between the urban Zahle and the Baalbek-Hermel
cluster of the northern valley, which carries a marked bedouin substrate. 
Eastern Bekaa shifts more southern dialects.  
The \textit{Jabal} cluster covers the
Mount Lebanon villages outside the Northern Matn belt, where conservative phonological
reflexes (notably the realization of \textit{q\=af}) and Aramaic-substrate
lexicon are best preserved \citep{lentin2018,germanos2009variation}.

Despite this microdialectal richness, computational coverage of Lebanese
remains thin. The Lebanese-side portion of Curras+Baladi \citep{EJHZ22} is to
date the only morphologically annotated Lebanese corpus, complemented on the
Syrian side of the Levantine continuum by N\^abra \citep{ANMFTM23}. 
\cardamom{}'s explicit
microdialectal split for Lebanese and coverage of its Southern, Chamali and Bikaa variants is meant to help close this gap. 

\subsection{Mauritanian Micro-dialects}
\textit{\d{H}ass\=aniyya} Arabic, also known as Mauritanian Arabic or \textit{kl\=am \d{H}ass\=an}, is the dominant spoken Arabic variety in Mauritania and the mother tongue of the Bi\d{d}\=an and \d{H}ar\=a\d{t}\=in. It also extends beyond Mauritania into the wider western Sahara. Mauritania itself is a multilingual and multiethnic country, where \textit{\d{H}ass\=aniyya} coexists with Pulaar, Sonink\'e, Wolof, Zenaga, French, and other languages, and where linguistic practices vary by region, neighborhood, social class, age, and urban setting \cite{naciriazzouz2023overview}. Historically, \textit{\d{H}ass\=aniyya} is associated with the Ban\=u \d{H}ass\=an Arab tribes and is generally classified as a Bedouin or Hil\=alian-type Arabic variety. Its development in the western Sahara has also been shaped by prolonged contact with \textit{Zenaga Berber} and, in southern areas, with West African languages such as \textit{Pulaar}, \textit{Sonink\'e}, and \textit{Wolof}, which helps explain why it differs substantially from both MSA and many mainstream North African Arabic varieties \cite{taine2020hassaniyya}.

For Mauritania specifically, the literature supports treating \textit{\d{H}ass\=aniyya} as a regional and sociolinguistic continuum rather than a single homogeneous dialect \cite{francis1979mauritanian,
cohen1963dialecte, taine2020hassaniyya, heath2004hassaniya}. This is consistent with descriptions of Mauritania as a highly diverse linguistic space: the north tends toward a \d{H}ass\=anophone majority, while multilingualism is more common in the south and the Senegal River Valley; large cities such as Nouakchott are especially heterogeneous, with \textit{\d{H}ass\=aniyya}, Pulaar, Wolof, French, and other languages functioning differently across contexts \cite{naciriazzouz2023overview}. Pedagogical descriptions of Mauritanian \textit{\d{H}ass\=aniyya} \cite{francis1979mauritanian} recognize four broad sub-dialects — Northern, Southern, Western, and Eastern Hassaniya — and further note variation between urban, rural, and nomadic speakers, while characterizing these local differences as primarily lexical and phonological rather than deeply structural. In \cardamom{}, we therefore adopt three geo-social annotation categories that are practical for the collected speech data and recognizable to native-speaker annotators: \textit{Western}, \textit{Northern}, and \textit{Southern/Eastern}.

\subsubsection{Western}
The Western variety, \Ar{الگبلة}/\textit{G\textschwa bla}, refers to the western and southwestern parts of Mauritania, including areas near the Senegal River. It is characterized by influence from Wolof and neighboring Sub-Saharan languages, along with local phonological and lexical features. Compared with the more desert-oriented eastern regions, this variety is also associated with a relatively more urban or commercial profile.

\subsubsection{Northern}
The Northern variety, \Ar{الساحل}/\textit{S\=a\d{h}il}, covers northern and northwestern Mauritania, including the Atlantic coastal strip. It is associated with stronger Bedouin and maritime influences, relatively conservative \textit{\d{H}ass\=aniyya} features, and lexical items connected to nomadic and coastal life.

\subsubsection{Southern}
The Southern/Eastern variety, \Ar{الشَّرگ}/\textit{Sharg}, covers eastern and southeastern Mauritania. It is characterized by a highly homogeneous \textit{\d{H}ass\=aniyya} profile, strong tribal and nomadic continuity, and comparatively conservative phonological and lexical features.

These categories should be understood as geo-social annotation categories rather than rigid dialect boundaries. As noted above, regional variation in \textit{\d{H}ass\=aniyya} is not deeply structural; it is more visible in pronunciation, lexicon, and certain pragmatic or usage-related features. Phonologically, however, \textit{\d{H}ass\=aniyya} differs from MSA in several important ways \cite{watson2002phonology,holes2004modern}.

Phonologically, \textit{\d{H}ass\=aniyya} shows several features that are important for micro-dialectal speech modeling. One of its conservative traits is the preservation of many Classical Arabic consonantal contrasts that have merged in several urban Arabic dialects, especially the interdental fricatives \textipa{/T/} \Ar{ث}, \textipa{/D/} \Ar{ذ}, and \textipa{/D\super Q/} \Ar{ظ}. This is important for ASR because these sounds may remain phonetically salient in \textit{\d{H}ass\=aniyya}, whereas other dialects often map them to \textipa{/t, d, d\super Q/} or \textipa{/s, z, z\super Q/} \cite{tainecheikh:halshs-00563853}.

Another central variable is the reflex of Classical Arabic \textipa{/q/}. As in many Bedouin-type dialects, \textipa{/q/} is often realized as a voiced velar stop \textipa{/g/} in \textit{\d{H}ass\=aniyya}. However, some descriptions also report alternations involving \textipa{/q/} and \textipa{/G/}, depending on the lexical item, register, and degree of classicization. The Peace Corps grammar notes that \textit{\d{H}ass\=aniyya} pronunciation shows local peculiarities compared with Classical Arabic and explicitly highlights alternations involving \textit{q\=af}, \textipa{/q/} \Ar{ق}, and \textit{ghayn}, \textipa{/G/} \Ar{غ}. For computational transcription, this means that Arabic orthography may hide substantial phonetic divergence from MSA \cite{hanchey1979mauritanian}.

\textit{\d{H}ass\=aniyya} also includes phonemes or marginal phonemes associated with contact and borrowing, especially \textipa{/v/}. This sound is absent from the traditional Classical Arabic phonemic inventory but is common in \textit{\d{H}ass\=aniyya}, particularly in words of French, Berber, or other contact origin. The Peace Corps material identifies \textipa{/v/} as a distinctive \textit{\d{H}ass\=aniyya} feature not represented in the Classical Arabic alphabet \cite{hanchey1979mauritanian}. This also aligns with Taine-Cheikh's broader account of contact-induced change in \textit{\d{H}ass\=aniyya}, where contact with Berber and other languages has enriched both the phonological and morphological systems \cite{taine2018hassaniyya}.

Beyond consonantal variation, \textit{\d{H}ass\=aniyya} differs from MSA in syllable structure, vowel realization, stress, and reduction. The \textit{\=Am\textrevglotstop ay\d{t}} book \cite{hayballah2024amait} and its review \cite{ouldalamir_amait_review} note that \textit{\d{H}ass\=aniyya} tends to shorten syllables and introduce consonant clustering or \textit{suk\=un}-like effects relative to the Classical/Standard system. These patterns directly affect how dialectal words should be written and transcribed. For ASR, this is especially important because a transcription policy normalized toward MSA may erase the phonological features needed for micro-dialect modeling.

The scholarly tradition on \textit{\d{H}ass\=aniyya} is substantial but uneven. Early works were largely practical and administrative, providing grammatical sketches, vocabulary lists, dialogues, and common expressions for communication with local populations. Representative examples include \cite{faidherbe1887langues} \textit{Langues s\'en\'egalaises}, \cite{mariebernard1893arabe} Boisroux's two-volume \textit{M\'ethode d'arabe parl\'e: idiome du S\'en\'egal}, and \cite{reynier1909methode}'s \textit{M\'ethode pour l'\'etude du dialecte maure}. Later work became more systematic, especially \cite{cohen1963dialecte}'s study of the \textit{G\textschwa bla} variety, which is described in the \textit{\=Am\textrevglotstop ay\d{t}} book review \cite{ouldalamir_amait_review} as a rigorous contribution to \textit{\d{H}ass\=aniyya} phonetics and morphology. Catherine \cite{tainecheikh:halshs-00563853}'s work is particularly central because it covers \textit{\d{H}ass\=aniyya} and Zenaga across phonetics, morphology, syntax, lexicon, prosody, and rhetoric. \cite{heath2004hassaniya,heath2003hassaniya}'s work on \textit{\d{H}ass\=aniyya} Arabic in Mali is also relevant for comparison, as it documents closely related Saharan varieties outside Mauritania.

\subsection{Palestinian Micro-dialects}

Palestinian Arabic forms a continuum of dialects across Palestine, with regional and social variations shaped by urbanization, village ties, and the dialect contact produced by forced displacement after the 1948 Nakba \cite{Shahin2008,CotterHoresh2015}. Traditional classifications often divide Palestinian colloquial speech into \emph{urban} vs.\ \emph{rural} types based on salient phonological features. However, this binary captures only part of the variation, as many speakers exhibit feature-mixing across local and social varieties, giving rise to several micro-dialects, including varieties associated with the Galilee, Carmel, the Nazareth region, the Jenin, Tulkarm and Nablus regions, Nablus city, the Ramallah and Jerusalem regions, the cities of Jerusalem and Jaffa, the Bethlehem and Hebron regions, Hebron city, the Gaza region, and the Negev region \cite{EJHZ22,JHRAZ17}. 
For analytical clarity, and given the limited linguistic resources available for each individual variety, this paper adopts a three-part classification: urban dialect, northern rural dialect, and southern rural dialect. This classification is not intended as a rigid geographic taxonomy; rather, it reflects recurring patterns of linguistic co-occurrence across phonological, morphosyntactic, and lexical features.

\subsubsection{Madani -- Urban Palestinian}
The urban dialect encompasses the major Palestinian cities, particularly the coastal cities stretching from Akko to Gaza, as well as the inland centers of Hebron, Jerusalem, and Nablus. The speech of these cities reflects a linguistic ecology shaped by long-standing urbanization, social mobility, and trade networks \cite{Shahin2008}. 

The main phonological feature of this variety is the realization of the letter (\textit{qāf} \Ar{قاف}), which is realized as ([\textipa{P}] \Ar{أ}) instead of ([Q] \Ar{ق}). For example, the word (\textit{qalb} \Ar{قلب}, heart) is pronounced (\textit{\textglotstop{}alb} \Ar{ألب}) \cite{JHRAZ17,ANMFTM23}. Other distinctive 
phonological features in the urban dialect include the pronunciation of \Ar{ث} (\textit{th\=a}) as \Ar{ت} (\textit{t\=a}) and \Ar{ذ} (\textit{dh\=a}) as \Ar{د} (\textit{d\=a}). For example, the word (\textit{kth\=ir} \Ar{كثير}, many) is realized as  (\textit{kt\=ir} \Ar{كتير}), and the word (\textit{hatha} \Ar{هذا}, this) is realized as (\textit{hada}) \cite{EJHZ22}.

The 1948 Nakba and the 1967 war triggered large-scale forced displacement that significantly altered the dialectal composition of Palestinian cities. For example, refugees from Jaffa and Lod settled in Ramallah and Jerusalem; Hebronite families relocated to Jerusalem; and refugees from Jaffa, Asdod, Asqalan, and Majdal were displaced into Gaza \cite{Shahin2008}. The Ramallah dialect currently represents the most salient outcome of this process — a leveled urban koine, sometimes referred to informally as the ``white'' dialect, that bears notable similarity to the Amman urban koine.

\subsubsection{\textit{Falā\d{h}ī} - Northern Rural Palestinian}

The Northern Rural Palestinian dialect is distributed across the northern West Bank and Galilee, extending from the Ramallah region through Salfit, Qalqilya, Tulkarm, and Jenin, and further north into Nazareth and the Galilee. These varieties exhibit distinctive phonological patterns closely associated with village identity.
The most salient phonological feature is a chain shift involving (\textit{qāf} \Ar{ق}) and (\textit{kāf} \Ar{ك}). The historical ([q] \Ar{ق}) is frequently realized as ([k] \Ar{ك}), while ([k] \Ar{ك}) is in turn shifted to the affricate ([\textipa{\t{tS}}] ''ch'' \Ar{تِّش}). This creates a systematic redistribution of phonological contrasts: the word (\textit{qalb} \Ar{قلب}, `heart') surfaces as (\textit{kalb} \Ar{كلب}), while (\textit{kalb} \Ar{كلب}, `dog') surfaces as (\textit{chalb} \Ar{تَّشَلب}), with the original contrast preserved through affrication \cite{JHAZ14}.
The prominence of these features varies by subregion. In the Ramallah area, the affrication of (\Ar{ك}) into (``ch'') is more prominent, while in the Qalqilya region the realization of (\Ar{ق}) as [k] is prevalent. In Jenin, Nazareth, and the Galilee, speakers more frequently retain the uvular [q] realization, often with a stronger and more emphatic articulation.
In addition to phonological variation, these dialects employ distinctive lexical items. The temporal adverb `now' illustrates clear regional stratification: \textit{hassa} (\Ar{هسا}) is prevalent in the northern West Bank, while \textit{issa} (\Ar{إسا}) is characteristic of Nazareth and Galilee speech. Both forms contrast with \textit{halla'} (\Ar{هلّأ}) in urban dialects and \textit{halh\={\i}n} (\Ar{هلحين}) in Southern Rural Palestinian varieties.

\subsubsection{\textit{Falā\d{h}ī} - Southern Rural Palestinian}
In southern Palestine, the phonological and lexical profile of Palestinian Arabic shifts markedly. The Southern Rural Palestinian dialect is distributed across villages extending south of Jerusalem, through al-Ta'\=am\=irah and Ban\={\i} 'Ubayd in the Bethlehem countryside, and further into the Hebron highlands. Its distribution extends southwest toward Beer al-Saba', where Naqab Bedouin communities preserve strong tribal speech patterns, and westward into Gaza, where displaced southern families have contributed to the speech patterns of camps and coastal communities \cite{Cotter2022,Shahin2008}.

The most salient phonological characteristic of this variety is the realization of (\textit{qāf}  \Ar{ق}). Unlike Northern Rural Palestinian varieties, where [q] frequently shifts to [k], and urban varieties, where it commonly becomes a glottal stop [\textglotstop], Southern Rural Palestinian speech often realizes /q/ as ([g] \Ar{گ}). Thus, (\textit{qalb} \Ar{قلب}, `heart') may surface as (\textit{galb} \Ar{گلب}) \cite{JHRAZ17}.
Like other Palestinian dialect varieties, Southern Rural Palestinian extends beyond village and tribal boundaries, having been shaped by sustained contact with neighboring Bedouin dialects across southern Jordan, northern Saudi Arabia, and the Sinai Peninsula \cite{Palva2008}. Such contact has contributed to the variety's distinctive phonological profile and its continued role as a marker of rural and tribal identity across southern Palestine.

\subsection{Saudi Micro-dialects}

 The categorization of Saudi Arabian dialects into different regions is a conventional method used in modern sociolinguistic and phonetic research \cite{ingham1994najdi,Prochazka1988}. The dominant classification in current research encompasses the Najdi, Hijazi, Eastern, Northern, and Southern dialects. Each dialect reflects unique characteristics across the levels of linguistic analysis. The following is a summary of each dialect group.

\subsubsection{Najdi}
The Najdi dialect is spoken across central Saudi Arabia and can be divided into four main regional varieties \cite{ingham1994najdi}: Central Najdi (spoken in the heart of Najd), Northern Najdi (linked to the Jabal Shammar and ha'il area), Mixed Northern-Central Najdi (common in the Qassim region), and Southern Najdi (extending to southern areas such as Najran).

Northern Najdi is primarily spoken in the ha'il region and surrounding northern areas \cite{ingham1994najdi}. It retains many conservative and Bedouin features, including traditional vocabulary and expressions passed down through generations. The Mixed Northern-Central Najdi dialect, centered in the Qassim region, is characterized by distinctive pronunciation patterns, such as the affrication of certain consonants, and a unique melodic intonation. This variety reflects a blend of both northern and central Najdi, shaped by internal migration and trade \cite{ingham1994najdi}. Central Najdi, dominant in the capital city Riyadh and its surrounding towns, is considered the standard variety of Najdi Arabic. Due to Riyadh’s political and economic prominence, this dialect has become associated with formal communication \cite{ingham1994najdi, Prochazka1988}. Finally, Southern Najdi, spoken along the southern edges of Najd and extending toward Najran, preserves older features that are less common in urban centers and shows some influence from southern Arabian dialects \cite{ingham1994najdi}.

In some Najdi dialects (more dominantly Qassimi dialect), the sound \textipa{/k/} is often 
realized as \textipa{[ts]}; for example, a word pronounced as 
\textipa{/kaD."Da:b/} \Ar{كذاب} in MSA may be realized as 
\textipa{[tsaD."Da:b]} \Ar{تسذاب} in Najdi speech. Similarly, the greeting 
\textipa{/\v{s}.lo:.nak/} \Ar{شلونك} (when addressing a female) may be 
realized as \textipa{[\v{s}.lo:.nats]} \Ar{شلونتس} in some Najdi dialects.

\subsubsection{Hijazi}

Another major part of Saudi Arabia is the Hijaz region, which is a historically significant region in western Saudi Arabia that stretches along the Red Sea coast and encompasses the major cities of Mecca, Medina, Jeddah, Yanbu, and Taif, as well as the surrounding tribal areas \cite{destinationksa_hejaz}. Linguistically, the region is far from homogenous. The ``Hijazi dialect'' is not a single variety but rather a collection of dialects that can be grouped into two main categories \cite{sambas2016hijazi}: 1) The Urban Hijazi which is spoken in the large cities of Mecca, Medina, Jeddah, Taif, and Yanbu. These cities have long attracted Arabs and Muslims, whether to live in proximity to the holy sanctuaries of Mecca and Medina or for economic and social opportunities, creating a speech variety shaped by both native residents and long-term settlers. This is the dialect that most people typically identify as ``Hijazi Arabic.'' 2) The tribal Hijazi which is Associated with the speech of Bedouin tribes inhabiting the broader Hijaz region. These dialects preserve Bedouin phonological and morphological features and are spoken by tribal groups whether in rural areas or, in some cases, within urban settlements.

In the Hijazi dialect, the sound \textipa{/D/} is 
often realized as \textipa{[z]}; for example, a word pronounced as  
\textipa{/kaD."Da:b/} \Ar{كذاب} is frequently pronounced as 
\textipa{[kaz."za:b]} in Hijazi speech. Another characteristic feature is the substitution of \textipa{/T/} with 
\textipa{[t]}; for example, a word pronounced as \textipa{/Ta.la:.Ta/} 
\Ar{ثلاثة}  may be realized as \textipa{[ta.la:.ta]} 
in Hijazi speech.

\subsubsection{Eastern}
The Eastern dialect of Saudi Arabia is primarily spoken in the Eastern Province, including key cities and regions such as Dammam, Al-Ahsa (Hofuf), Qatif, and Khobar. The Eastern dialect is very similar to other countries in the Gulf region, such as Bahrain and Qatar \cite{gulfarabicresources}. This can be attributed to the boarder shared between Saudi and Bahrain and Saudi and Qatar. 

Socially, the Eastern dialect is often described as slow in tempo, contrasting with the faster, more clipped rhythm of other dialects such as Najdi or Southern Arabic. Vowel quality in Eastern Arabic is notably distinct. The /\textipa{a}/ sound in medial syllables can be rounded, while in final position it may appear as a more open /\textipa{a:}/, imparting a unique musicality to the dialect. Additionally, the Eastern dialect is known for preserving some conservative phonetic and lexical features that have evolved or disappeared in other Saudi dialects. For instance, certain traditional vocabulary items such as the word ``simat'' (meaning `dining mat'), are still used in the Eastern region but are rarely heard in other parts of Saudi Arabia. Additionally, a word such as \textipa{/kaDDab/} \Ar{كذاب} may be realized as \textipa{[tSaDDab]} \Ar{تشذاب}.

\subsubsection{Southern}
The Southern dialect of Saudi Arabia is spoken in the southern regions, notably in cities and areas such as Abha, Khamis Mushait, Jazan, Najran. This dialect is closest to the Yemeni dialect \cite{asiri2009}. The southern dialect can be further divided into Asiri (spoken in the highlands near Abha), Tihami (along the Red Sea coastal plain), Najrani (in Najran), and Jizani (in Jazan) \cite{alqahtani2015tihami}. The Southern dialect is often recognized for its fast and clipped rhythm. Phonological studies have shown that southern dialects may retain or develop consonantal and vowel features that are rare in other parts of Saudi Arabia, such as unique realizations of /q/.

The Asiri dialect is mostly spoken in cities such as Abha and Khamis Mushait. It maintains traditional Arabic features, including gender and plural distinctions in verbs and pronouns \cite{asiri2009}. The Tihami dialect is spoken along the Red Sea coastal plain, spanning the lowlands of western Asir and part of Jazan. It is notable for its rapid, clipped rhythm and simplified syllable structure. The Najrani and Jizani dialects have a similar grammar and vocabulary to Yemeni Arabic, a result of longstanding cultural and geographic proximity to Yemen.

In some Southern dialects, the greeting 
\textipa{/kay.fa \textsl{h}a:.lak/} \Ar{كيف حالك} (when addressing a female) 
is often pronounced as \textipa{[kayf \textsl{h}a:.la\v{s}]} \Ar{كيف حالش}. 
This reflects a common shift where the final ``k'' sound changes into a 
``sh'' sound. This feature, known as \textit{kashkasha}, is a well-known 
characteristic of speech of the Southern dialect.

\subsubsection{Northern}
The Northern dialect of Saudi Arabia is predominantly spoken in the northern regions, including cities and areas such as Tabuk, Arar, Al-Jawf, Rafha, and the surrounding rural communities. Al-Jawf dialect is  influenced by its proximity to Jordan, Syria, and Iraq \cite{north_Dialect}. However, dialects spoken in Arar, Tabuk, and Rafha are very similar to Ha'il dialect.  

Linguistically, the Northern dialect exhibits several salient features. There is often a rising intonation in speech, and certain words or pronunciations may be hard for speakers from other Saudi regions to fully grasp. The Northern dialect is often associated with Bedouin heritage and retains many traditional speech patterns reflective of the nomadic lifestyle prevalent in these regions. 

In some Northern dialects,  the sound \textipa{/k/} is often 
realized as  \textipa{[ts]}; for example, a word pronounced as 
\textipa{/kaD."Da:b/} \Ar{كذاب}  may be realized as 
\textipa{[tsaD."Da:b]} \Ar{تسذاب}> in Northern speech, similar to some Najdi dialect. Similarly, the greeting 
\textipa{/\v{s}ax.ba:.rak/} \Ar{شخبارك} (when addressing a female) may be 
realized as \textipa{[\v{s}ax.ba:.rats]} \Ar{شخبارتس}.

\section{Micro-Dialectal Overlap}
\label{appendix:microdialectal-overlap}

\subsection{Extended Corpus Statistics}
Table~\ref{tab:micro-dialect-detailed-full} extends Table~\ref{tab:micro-dialect-detailed} (Section~\ref{sec:corpus_stats}) with segment-duration range (\textit{Min}, \textit{Max}), average transcript length (\textit{AVT}), vocabulary size and density (\textit{U-Wds}, \textit{Avg U-Wds/hr}), and speaking-rate proxies (\textit{WPS}, \textit{CPS}).

\begin{table*}[t!]
\centering
\setlength{\tabcolsep}{1.5pt}
\resizebox{0.9\textwidth}{!}{%
\begin{tabular}{llcccccccccccc}
\hline

\textbf{Country} & \textbf{Micro-dialect} & \textbf{Segments} & \textbf{Total Hours} & \textbf{Mean} & \textbf{Min} & \textbf{Max} & \textbf{CS Hours} & \textbf{F/M} & \textbf{AVT} & \textbf{U-Wds} & \textbf{Avg U-Wds/hr} & \textbf{WPS} & \textbf{CPS} \\
\hline

\multirow{5}{*}{\textbf{\Egypt}}
& Cairene & 3,871 & 3.482 & 3.238 & 0.164 & 24.736 & 0.264 & 33.4/66.6 & 8.31 & 11,654 & 3,347.22 & 2.55 & 10.88 \\
& Saidi & 1,735 & 1.771 & 3.675 & 0.624 & 57.797 & 0.008 & 37.9/62.1 & 8.00 & 5,674 & 3,203.63 & 2.13 & 8.91 \\
& Port Saidi & 1,124 & 1.021 & 3.269 & 0.223 & 24.539 & 0.039 & 12.2/87.8 & 7.47 & 3,571 & 3,498.42 & 2.27 & 9.28 \\
& Alexandrian & 848 & 0.769 & 3.265 & 0.580 & 15.642 & 0.020 & 40.7/59.3 & 8.03 & 3,349 & 4354.29 & 2.44 & 10.31 \\
& \textbf{\textit{TOTAL}} & \textbf{7,364} & \textbf{6.889} & \textbf{3.368} & \textbf{0.164} & \textbf{57.797} & \textbf{0.327} & \textbf{32.1/67.9} & \textbf{8.13} & \textbf{18,805} & \textbf{2,729.69} & \textbf{2.39} & \textbf{10.08} \\
\hline

\multirow{4}{*}{\textbf{\Jordan}}
& Falahi & 3,276 & 4.540 & 4.989 & 0.523 & 33.065 & 0.119 & 18.6/81.4 & 10.82 & 10,743 & 2,366.27 & 2.15 & 8.52 \\
& Badawi & 2,899 & 3.968 & 4.928 & 0.249 & 30.946 & 0.005 & 26.6/73.4 & 8.97 & 8,057 & 2,030.37 & 1.80 & 7.02 \\
& Madani & 1,909 & 2.755 & 5.196 & 0.785 & 35.466 & 0.309 & 38.2/61.8 & 11.90 & 7,703 & 2,795.56 & 2.28 & 8.99 \\
& \textbf{\textit{TOTAL}} & \textbf{7,729} & \textbf{10.806} & \textbf{5.033} & \textbf{0.249} & \textbf{35.466} & \textbf{0.407} & \textbf{26.5/73.5} & \textbf{10.43} & \textbf{20,504} & \textbf{1,897.42} & \textbf{2.06} & \textbf{8.09} \\
\hline

\multirow{4}{*}{\textbf{\Lebanon}}
& Chamali & 1,105 & 2.071 & 6.746 & 0.059 & 31.316 & 0.560 & 13.6/86.4 & 16.07 & 5,575 & 2,692.38 & 2.38 & 12.30 \\
& Jnoubi & 1,464 & 3.123 & 7.679 & 0.058 & 41.439 & 0.212 & 19.4/80.6 & 14.35 & 7,705 & 2,467.43 & 1.87 & 9.16 \\
& B'albaki & 258 & 0.737 & 10.281 & 0.778 & 30.000 & 0.185 & 28.4/71.6 & 19.24 & 2,404 & 3,262.87 & 1.87 & 9.23 \\
& \textbf{\textit{TOTAL}} & \textbf{2,681} & \textbf{5.628} & \textbf{7.557} & \textbf{0.058} & \textbf{41.439} & \textbf{0.902} & \textbf{18.0/82.0} & \textbf{15.56} & \textbf{12,702} & \textbf{2,256.90} & \textbf{2.06} & \textbf{10.34} \\
\hline

\multirow{4}{*}{\textbf{\Mauritania}}
& Western & 1,210 & 3.780 & 11.247 & 0.013 & 76.181 & 0.464 & 8.3/91.7 & 25.78 & 9,256 & 2,448.60 & 2.21 & 9.16 \\
& Northern & 1,205 & 3.809 & 11.380 & 0.013 & 76.181 & 0.470 & 8.1/91.9 & 25.76 & 9,293 & 2,439.56 & 2.16 & 9.02 \\
& Southern & 1,126 & 3.666 & 11.720 & 0.013 & 76.181 & 0.467 & 8.7/91.3 & 26.80 & 9,087 & 2,478.92 & 2.20 & 9.18 \\
& \textbf{\textit{TOTAL}} & \textbf{1,314} & \textbf{3.998} & \textbf{10.953} & \textbf{0.013} & \textbf{76.181} & \textbf{0.473} & \textbf{7.7/92.3} & \textbf{24.73} & \textbf{9,563} & \textbf{2,392.07} & \textbf{2.16} & \textbf{8.96} \\
\hline

\multirow{4}{*}{\textbf{\Palestine}}
& Falahi & 1,564 & 2.720 & 6.260 & 0.446 & 87.218 & 0.028 & 9.0/91.0 & 9.91 & 5,782 & 2,125.96 & 1.56 & 6.47 \\
& Badawi & 864 & 1.461 & 6.088 & 0.947 & 46.522 & 0.001 & 3.1/96.9 & 8.31 & 3,166 & 2,166.83 & 1.36 & 5.81 \\
& Madani & 537 & 0.997 & 6.686 & 0.079 & 92.441 & 0.001 & 32.3/67.7 & 8.35 & 2,223 & 2,229.12 & 1.24 & 5.23 \\
& \textbf{\textit{TOTAL}} & \textbf{2,956} & \textbf{5.159} & \textbf{6.283} & \textbf{0.079} & \textbf{92.441} & \textbf{0.031} & \textbf{11.5/88.5} & \textbf{9.14} & \textbf{9,018} & \textbf{1,748.10} & \textbf{1.44} & \textbf{6.04} \\
\hline

\multirow{6}{*}{\textbf{\SaudiArabia}}
& Southern & 1,423 & 2.550 & 6.452 & 0.922 & 63.943 & 0.088 & 40.5/59.5 & 14.85 & 6,057 & 2,374.90 & 2.30 & 8.71 \\
& Hijazi & 1,413 & 2.661 & 6.779 & 0.946 & 27.251 & 0.250 & 95.0/5.0 & 15.65 & 5,937 & 2,231.39 & 2.31 & 9.17 \\
& Najdi & 1,089 & 2.234 & 7.384 & 0.982 & 29.618 & 0.156 & 3.4/96.6 & 14.88 & 5,196 & 2,326.32 & 2.01 & 7.87 \\
& Eastern & 956 & 2.023 & 7.618 & 0.918 & 51.223 & 0.044 & 17.9/82.1 & 13.26 & 4,183 & 2,067.83 & 1.70 & 6.67 \\
& Northern & 514 & 1.009 & 7.064 & 1.369 & 15.553 & 0.003 & 0.4/99.6 & 14.87 & 3,088 & 3,061.71 & 2.10 & 8.74 \\
& \textbf{\textit{TOTAL}} & \textbf{5,376} & \textbf{10.425} & \textbf{6.981} & \textbf{0.918} & \textbf{63.943} & \textbf{0.538} & \textbf{39.6/60.4} & \textbf{14.77} & \textbf{18,113} & \textbf{1,737.46} & \textbf{2.10} & \textbf{8.25} \\
\hline
\end{tabular}%
}
\caption{Extended \textsc{Cardamom} statistics by country and micro-dialect. \textit{Segments} gives the number of utterance segments, and \textit{Total Hours} gives the corresponding speech duration. \textit{Mean}, \textit{Min}, and \textit{Max} report segment duration statistics in seconds. \textit{CS Hours} gives the total duration of code-switched speech. \textit{F/M} reports the perceived speaker-gender distribution as female/male percentages. \textit{AVT} is the average transcript length. \textit{U-Wds} is the number of unique words, and \textit{Avg U-Wds/hr} is the number of unique words per hour. \textit{WPS} and \textit{CPS} denote words per second and characters per second, respectively. Country-level totals (\textbf{bold}) count each segment once; per-dialect rows count label memberships and can exceed the total when segments carry multiple micro-dialect labels.}
\label{tab:micro-dialect-detailed-full}
\end{table*}

To examine fluid boundaries between micro-dialect labels, we analyze segments assigned to more than one micro-dialect within the same country. Table~\ref{tab:dialect_intersections} reports the main pairwise overlaps. The highest overlap appears in Mauritania, where many segments are shared across Northern, Southern, and Western labels, reflecting the difficulty of separating highly related Hassaniya varieties in available public speech. Jordan also shows strong overlap, especially between Badawi--Falahi and Falahi--Madani, suggesting that Falahi functions as a bridge between Bedouin and urban varieties. In contrast, Saudi and Palestinian micro-dialects show much lower overlap, indicating more separated label assignments in the corpus.

\begin{table}[t!]

\centering
\small
\setlength{\tabcolsep}{4pt}

\begin{tabular}{@{}lllr@{}}
\toprule
\textbf{Country} & \textbf{Dialect A} & \textbf{Dialect B} & \textbf{Overlap Count} \\ 
\midrule

\rowcolor{orange!20}
\textbf{Egypt} & Alexandrian & Cairene & 130 \\
\rowcolor{orange!20}
               & Alexandrian & Port Saidi & 10 \\
\rowcolor{orange!20}
               & Alexandrian & Saidi & 3 \\
\rowcolor{orange!20}
               & Cairene & Port Saidi & 67 \\
\rowcolor{orange!20}
               & Cairene & Saidi & 9 \\
\rowcolor{orange!20}
               & Port Saidi & Saidi & 2 \\ 
\cmidrule(l){2-4}

\rowcolor{olive!30}
\textbf{Jordan} & Badawi & Falahi & 178 \\
\rowcolor{olive!30}
                & Badawi & Madani & 23 \\
\rowcolor{olive!30}
                & Falahi & Madani & 171 \\ 
\cmidrule(l){2-4}

\rowcolor{red!15}
\textbf{Lebanon} & B'albaki & Chamali & 31 \\
\rowcolor{red!15}
                 & B'albaki & Jnoubi & 30 \\
\rowcolor{red!15}
                 & Chamali & Jnoubi & 32 \\ 
\cmidrule(l){2-4}

\rowcolor{cyan!7}
\textbf{Mauritania} & Northern  & Southern  & 1114 \\
\rowcolor{cyan!7}
                    & Northern  & Western  & 1110 \\
\rowcolor{cyan!7}
                    & Southern  & Western & 1112 \\ 
\cmidrule(l){2-4}

\rowcolor{teal!20}
\textbf{Palestine} & Badawi & Madani & 6 \\
\rowcolor{teal!20}
                   & Falahi & Madani & 3 \\ 
\cmidrule(l){2-4}

\rowcolor{green!15}
\textbf{Saudi} & Eastern & Najdi & 3 \\
\rowcolor{green!15}
               & Eastern & Southern & 1 \\
\rowcolor{green!15}
               & Hijazi & Najdi & 14 \\
\rowcolor{green!15}
               & Northern & Southern & 1 \\ 
\bottomrule

\end{tabular}
\caption{Primary micro-dialectal intersections by region, reflecting the most fluid linguistic boundaries in the corpus.}
\label{tab:dialect_intersections}
\end{table}

\section{Speaker-Controlled Overlap Analysis}
\label{appendix:speaker-clustering}

To check whether the multi-label overlap patterns discussed in Section~\ref{subsec:quant_overlap} reflect genuine dialectal convergence rather than a small number of recurring speakers or channels, we cluster automatic speaker embeddings on the single-label segments of \cardamom{} (24,402 segments; multi-label segments are excluded to avoid conflating speaker overlap with label overlap). We extract embeddings with the off-the-shelf \texttt{pyannote/wespeaker-voxceleb-resnet34-LM} \footnote{\url{https://huggingface.co/pyannote/wespeaker-voxceleb-resnet34-LM}} model (6.63M parameters), excluding clips shorter than 1.0\,s. We group embeddings with average-linkage agglomerative clustering over cosine distance and restrict linkage to clips from the same YouTube video, so the clusters are \textit{video-local} estimates: a person appearing in two videos is counted twice, and no identity matching is attempted across unrelated videos. We set the distance threshold a priori to 0.25 (similarity 0.75) and did not validate clusters against human speaker labels; the resulting 19,770 estimated speakers should therefore be read as approximate. Table~\ref{tab:speaker-threshold-sensitivity} shows that totals change by 11--13\% under thresholds of 0.20 and 0.30, and that the larger cross-dialect intersections persist while some small ones disappear under the strict setting; we treat the latter as threshold-sensitive. Table~\ref{tab:speaker-per-microdialect} reports unique speaker counts per micro-dialect, and Table~\ref{tab:speaker-per-intersection} reports unique speaker counts within the multi-label intersections analyzed in Section~\ref{subsec:quant_overlap} and Appendix~\ref{appendix:microdialectal-overlap}; the intersection counts are drawn from multi-label utterances only and are not independent cross-dialect evidence on their own, but are informative alongside the single-label counts.

\begin{table}[t]
\centering
\small
\begin{tabular}{@{}llr@{}}
\toprule
\textbf{Country} & \textbf{Micro-dialect} & \textbf{Unique speakers} \\
\midrule
\multirow{4}{*}{\Egypt} & Alexandrian & 653 \\
 & Cairene & 2,938 \\
 & Port Saidi & 957 \\
 & Saidi & 1,566 \\
\midrule
\multirow{3}{*}{\Jordan} & Badawi & 2,271 \\
 & Falahi & 2,480 \\
 & Madani & 1,346 \\
\midrule
\multirow{3}{*}{\Lebanon} & B'albaki & 72 \\
 & Chamali & 590 \\
 & Jnoubi & 825 \\
\midrule
\multirow{3}{*}{\Mauritania} & Northern & 61 \\
 & Southern & 7 \\
 & Western & 53 \\
\midrule
\multirow{3}{*}{\Palestine} & Badawi & 785 \\
 & Falahi & 1,383 \\
 & Madani & 471 \\
\midrule
\multirow{5}{*}{\SaudiArabia} & Eastern & 717 \\
 & Hijazi & 408 \\
 & Najdi & 720 \\
 & Northern & 334 \\
 & Southern & 1,221 \\
\bottomrule
\end{tabular}
\caption{Estimated unique speakers per micro-dialect (single-label segments only), from automatic speaker-embedding clustering.}
\label{tab:speaker-per-microdialect}
\end{table}

\begin{table}[t]
\centering
\small
\setlength{\tabcolsep}{3pt}
\begin{tabular}{@{}llr@{}}
\toprule
\textbf{Country} & \textbf{Intersection} & \textbf{Unique speakers} \\
\midrule
\multirow{3}{*}{\Mauritania} & Northern--Southern & 336 \\
 & Northern--Western & 334 \\
 & Southern--Western & 333 \\
\midrule
\multirow{3}{*}{\Jordan} & Falahi--Madani & 160 \\
 & Badawi--Falahi & 157 \\
 & Badawi--Madani & 22 \\
\midrule
\multirow{4}{*}{\Egypt} & Alexandrian--Cairene & 127 \\
 & Cairene--Port Saidi & 58 \\
 & Alexandrian--Port Saidi & 10 \\
 & Cairene--Saidi & 9 \\
\midrule
\multirow{3}{*}{\Lebanon} & Chamali--Jnoubi & 55 \\
 & B'albaki--Chamali & 53 \\
 & B'albaki--Jnoubi & 53 \\
\midrule
\Palestine & Badawi--Madani & 6 \\
\midrule
\SaudiArabia & Hijazi--Najdi & 6 \\
\bottomrule
\end{tabular}
\caption{Estimated unique speakers within multi-label intersections (multi-label utterances only). Intersections with fewer than two estimated distinct speakers are omitted.}
\label{tab:speaker-per-intersection}
\end{table}

\begin{table}[t]
\centering
\small
\setlength{\tabcolsep}{3pt}
\begin{tabular}{@{}rrrrr@{}}
\toprule
\textbf{Distance} & \textbf{Speakers} & \textbf{JOR B--F} & \textbf{EGY A--C} & \textbf{LEB B--C} \\
\midrule
0.20 & 21,881 & 11 & 4 & 3 \\
0.25 & 19,770 & 30 & 8 & 3 \\
0.30 & 17,239 & 38 & 14 & 3 \\
\bottomrule
\end{tabular}
\caption{Sensitivity of video-local speaker clustering on single-label clips to the cosine-distance threshold (0.25 is the primary setting). Pair columns give estimated speakers shared by Jordanian Badawi--Falahi, Egyptian Alexandrian--Cairene, and Lebanese B'albaki--Chamali.}
\label{tab:speaker-threshold-sensitivity}
\end{table}

\paragraph{Split leakage.} Applying the same clustering to all clips, we estimate 719 speakers shared between the train and development partitions of the utterance-level split, 753 between train and test, and 725 between development and test. This motivates the source-disjoint MDI split (Appendix~\ref{sec:data-splits}).

\section{Annotation Schema}
\label{sec:annotation-schema}
\begin{figure}[t]
    \centering
    \includegraphics[width=\linewidth]{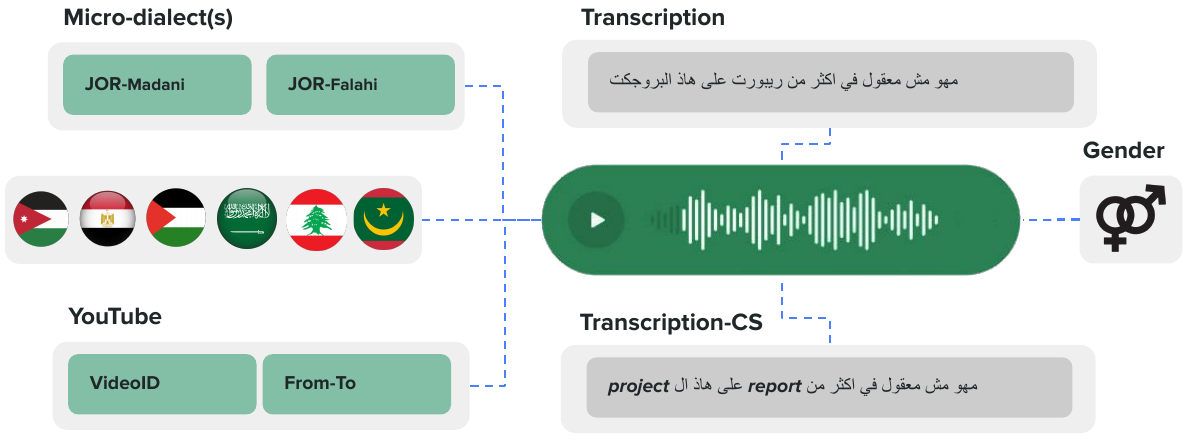}
    \caption{Annotation schema for each speech segment. Each speech segment is associated with: (1) transcription, (2) code-switching status, (3) country-level dialect, (4) micro-dialect label(s), (5) perceived speaker gender, (6) YouTube video ID, and (7) start/end timestamps in the original source video.}
    \label{fig:annotation}
\end{figure}

The complete annotation schema for each speech segment, including transcription, code-switching status, dialect and micro-dialect labels, speaker metadata, source video ID, and timestamp information, is shown in Figure~\ref{fig:annotation}.

\subsection{Annotators per Micro-dialect}
Table~\ref{tab:annotators-per-microdialect} reports the number of annotators covering each micro-dialect, together with the country-level totals reported in the main text. As noted in the Dataset Description, these per-micro-dialect counts are not mutually exclusive: several annotators were fluent in, and annotated, more than one micro-dialect within their country.

\begin{table}[t]
\centering
\small
\begin{tabular}{@{}llr@{}}
\toprule
\textbf{Country} & \textbf{Micro-dialect} & \textbf{Annotators} \\
\midrule
\multirow{4}{*}{\Egypt\ (5)} & Cairene & 3 \\
 & Saidi & 1 \\
 & Port Saidi & 1 \\
 & Alexandrian & 2 \\
\midrule
\multirow{3}{*}{\Jordan\ (8)} & Falahi & 4 \\
 & Madani & 2 \\
 & Badawi & 2 \\
\midrule
\multirow{3}{*}{\Lebanon\ (2)} & Chamali & 1 \\
 & Jnoubi & 2 \\
 & B'albaki & 1 \\
\midrule
\multirow{3}{*}{\Mauritania\ (4)} & Western & 2 \\
 & Northern & 2 \\
 & Southern & 2 \\
\midrule
\multirow{3}{*}{\Palestine\ (3)} & Falahi & 2 \\
 & Madani & 2 \\
 & Badawi & 1 \\
\midrule
\multirow{5}{*}{\SaudiArabia\ (9)} & Southern & 2 \\
 & Hijazi & 2 \\
 & Najdi & 3 \\
 & Northern & 2 \\
 & Eastern & 2 \\
\bottomrule
\end{tabular}
\caption{Number of annotators covering each micro-dialect, grouped by country (country-level participant totals in parentheses). Counts are not mutually exclusive: annotators fluent in more than one micro-dialect are counted under each variety they annotated.}
\label{tab:annotators-per-microdialect}
\end{table}

\section{Representative Multi-label Dialect Segments}
\label{appendix:multi-label-examples}
Table~\ref{tab:multi_label_examples} provides representative examples of utterances annotated with more than one micro-dialect label. These cases illustrate that multi-label segments often reflect dialectal convergence rather than annotation ambiguity: a single utterance may contain lexical, phonological, or pragmatic cues shared by neighboring or socially related varieties. The examples span several countries and show overlap both within closely related regional varieties, such as Alexandrian--Port Saidi and Chamali--Jnoubi, and across broader intra-country dialect groupings, such as Badawi--Madani or Hijazi--Najdi. This supports our treatment of multi-label annotations as valid label memberships rather than disjoint classes, since they preserve naturally shared dialectal features in spontaneous speech. We give an English translation and a Latin transliteration for each example.

\begin{table*}[t!]
\centering
\footnotesize
\setlength{\tabcolsep}{3pt}
\renewcommand{\arraystretch}{1.12}

\begin{tabularx}{\textwidth}{@{}
p{1.35cm}
>{\raggedleft\arraybackslash}p{5cm}
>{\raggedright\arraybackslash}X
>{\raggedright\arraybackslash}p{4.2cm}
p{1.75cm}
@{}}
\toprule
\textbf{Region} & \textbf{Arabic} & \textbf{English} & \textbf{Latin transliteration} & \textbf{Multi-label} \\
\midrule

\Jordan
& \setcode{utf8}\<يلا بخاطرك>
& Alright, see you!
& Bal\=a b-kh\=a\d{t}rak.
& Badawi, Falahi \\

\Jordan
& \setcode{utf8}\<اه والله مزبوط>
& Yes, that is correct.
& \=Ah wall\=ah mazb\=u\d{t}.
& Falahi, Madani \\

\Egypt
& \setcode{utf8}\<استنى يابا>
& Wait, Dad.
& Istanni y\=a b\=aba.
& Alex., Port Saidi \\

\Egypt
& \setcode{utf8}\<فلوس باباكي متلزمناش>
& Your father's money does not concern us.
& Ful\=us b\=ab\=aki ma tilzam-n\=ash.
& Alex., Cairene \\

\SaudiArabia
& \setcode{utf8}\<تماما ينطبق نفس الشي على الرجال>
& Exactly, the same thing applies to men.
& Tam\=aman yin\d{t}abiq nafs il-shay 'ala al-rij\=al.
& Hijazi, Najdi \\

\Palestine
& \setcode{utf8}\<قديش يعني؟ قد البحر. عاد ليش خايف؟>
& How much? As big as the sea. Why are you afraid?
& Qaddesh ya'ni? Add il-ba\d{h}r. 'Ad l\=esh kh\=ayef?
& Badawi, Madani \\

\Lebanon
& \setcode{utf8}\<بتزين السفرة بهالاشيا>
& It decorates the table with these things.
& Btzayyen is-sufra b-hal-ashya.
& Chamali, Jnoubi \\

\Mauritania
& \setcode{utf8}\<شوف انا اولدت ف ولاية انشيري >
& Look, I was born in Inchiri state; I am one of the people of Inchiri.
& Sh\=uf, ana wladt f wil\=ayat Inchiri; ana w\=ahed min ahl Inchiri.
& South., North. \\

\bottomrule
\end{tabularx}
\caption{Examples of multi-label segments from the \textsc{Cardamom} corpus showing how dialects can overlap in real use, with English translation and Latin transliteration.}
\label{tab:multi_label_examples}
\end{table*}

\section{Code-switching Examples}
\label{appendix:codeswitch-examples}

Table~\ref{tab:codeswitch-examples} illustrates several common types of Arabic--English/French code-switching observed in \cardamom{} across countries and micro-dialects. Some examples involve insertional or intra-sentential code-switching, where an English lexical item or phrase appears inside an otherwise Arabic utterance (e.g., \textit{queue}, \textit{social media}, and \textit{my life}). Others are closer to inter-sentential code-switching, where a larger English phrase or discourse unit is produced alongside Arabic material, as in \textit{it pays off}. We give an English translation and a Latin transliteration for each example.

\begin{table*}[t]
\centering
\small
\setlength{\tabcolsep}{4pt}
\begin{tabular}{@{}p{0.14\textwidth}
>{\RaggedLeft\arraybackslash}p{0.35\textwidth}
>{\RaggedRight\arraybackslash}p{0.28\textwidth}
>{\RaggedRight\arraybackslash}p{0.24\textwidth}@{}}
\toprule
\textbf{Dialect} & \textbf{Arabic} & \textbf{English translation} & \textbf{Latin transliteration} \\
\midrule
Egypt--Cairene &
\<انت كويس؟> Morning &
Morning, are you okay? &
Morning, inta kwayyis? \\[0.35em]

Jordan--Madani &
\<هلا لا هيه مش ست سنين أكيد بس> \<بالاخير> it pays off &
Well, no, it definitely isn't six years, but it pays off in the end. &
Halla la, hiyye mish sitt sn\={\i}n ak\={\i}d, bas it pays off bil-\=akhir. \\[0.35em]

Saudi--Najdi &
\<برا> queue \<امس كنت في محل عنده> &
Yesterday, I was at a store that had a queue outside. &
Ams kunt f\={\i} ma\d{h}all 'indah queue barra. \\[0.35em]

Palestine--Falahi &
\<انا بدي احس بحياتي انا بطلت احس ب> my life &
I want to feel alive; I no longer feel connected to my life. &
Ana biddi a\d{h}iss bi-\d{h}ay\=at\={\i}, ana ba\d{t}allt a\d{h}iss bi-my life. \\[0.35em]

Lebanon--B'albaki &
\<كتير يعني عطول فأنا نقلت عال> social media &
A lot, basically all the time, so I moved to social media. &
Kt\={\i}r ya'ni 'a-\d{t}\=ul, fa-ana na'alt 'al-social media. \\[0.35em]

Mauritania--Northern &
Instagram \<وابدأت نعدلها على> \<بعد ذاك كانو معاي شباب> &
I started editing it on Instagram; after that, some young men were with me. &
W-ibd\={\i}t n'addilha 'al\=a Instagram, ba'd dh\=ak k\=an\=u ma'\={\i}ya shab\=ab. \\
\bottomrule
\end{tabular}
\caption{Representative Arabic--English code-switched utterances with English translation and Latin transliteration.}
\label{tab:codeswitch-examples}
\end{table*}

\section{Train/Development/Test Splits}
\label{sec:data-splits}

We divide the data of each country independently into train, development, and
test sets.  The three partitions are kept as close as possible in size, so that
each contains approximately one third of the utterances from that country.  An
utterance is assigned to only one partition; it is never copied across train,
development, and test.  This country-wise design preserves the coverage of each
country in all partitions while preventing larger countries from determining
the split sizes of smaller ones. 

Micro-dialects are taken into account within each country.  A split based only
on the total number of utterances could leave a low-resource micro-dialect
underrepresented in one partition, even when the country-level split sizes are
balanced.  We therefore balance the micro-dialect label memberships alongside
the utterance counts.  For each country, the target is to place approximately
one third of the examples associated with each micro-dialect in train, one
third in development, and one third in test.  The resulting label counts are
shown in Table~\ref{tab:micro-split-counts}.

This treatment also accounts for overlap between micro-dialects.  Some
utterances are annotated with more than one micro-dialect label, for example an
utterance belonging to both Badawi and Falahi.  Such an utterance remains a
single example and is assigned to a single partition, but it contributes to the
balance of both micro-dialect labels.  We do not convert overlapping labels into
a separate class, and we do not duplicate the utterance once per label.  This
preserves the original annotation structure while keeping the marginal
distribution of each micro-dialect similar across the three partitions.



\begin{table}[t]
\centering
\small
\resizebox{0.48\textwidth}{!}{%
\begin{tabular}{@{}llrrrr@{}}
\toprule
\textbf{Country} & \textbf{Micro-dialect} & \textbf{All} &
\textbf{Train} & \textbf{Dev.} & \textbf{Test} \\
\midrule
\Egypt      & Alexandrian &   848 & 282 & 283 & 283 \\
           & Cairene     & 3,871 & 1,290 & 1,291 & 1,290 \\
           & Port Saidi  & 1,124 & 375 & 375 & 374 \\
           & Saidi       & 1,735 & 578 & 578 & 579 \\
\midrule
\Jordan     & Badawi      & 2,899 & 967 & 966 & 966 \\
           & Falahi      & 3,276 & 1,092 & 1,092 & 1,092 \\
           & Madani      & 1,909 & 637 & 636 & 636 \\
\midrule
\Lebanon    & B'albaki    &   258 & 86 & 86 & 86 \\
           & Chamali     & 1,105 & 368 & 369 & 368 \\
           & Jnoubi      & 1,464 & 487 & 488 & 489 \\
\midrule
\Mauritania & Northern    & 1,205 & 402 & 401 & 402 \\
           & Southern    & 1,126 & 376 & 375 & 375 \\
           & Western     & 1,210 & 402 & 404 & 404 \\
\midrule
\Palestine  & Badawi      &   864 & 288 & 288 & 288 \\
           & Falahi      & 1,564 & 521 & 522 & 521 \\
           & Madani      &   537 & 179 & 179 & 179 \\
\midrule
\SaudiArabia      & Eastern     &   956 & 319 & 318 & 319 \\
           & Hijazi      & 1,413 & 471 & 471 & 471 \\
           & Najdi       & 1,089 & 363 & 363 & 363 \\
           & Northern    &   514 & 172 & 171 & 171 \\
           & Southern    & 1,423 & 473 & 475 & 475 \\
\bottomrule
\end{tabular}
}
\caption{Micro-dialect label memberships in the country-wise splits.  Counts
are label memberships rather than disjoint utterance counts:
an utterance with two micro-dialect labels contributes to both corresponding
rows but remains a single utterance in one partition.}
\label{tab:micro-split-counts}
\end{table}

\paragraph{Source-disjoint split for MDI.} The utterance-level split above can place segments from the same YouTube video, and therefore the same speakers and recording conditions, in different partitions (Appendix~\ref{appendix:speaker-clustering}). For micro-dialect identification, we therefore assign complete videos to train, development, or test with a deterministic mixed-integer optimization (fixed seed) that targets an 80/10/10 allocation of country-qualified micro-dialect memberships. After removing three exact duplicate waveforms, the split contains 22,271/2,492/3,062 segments from 320/47/46 videos, with no video in more than one partition. Because some labels are concentrated in few videos, balance is necessarily imperfect: Table~\ref{tab:mdi-source-split} reports label memberships together with the number of contributing videos. For example, all 683 Cairene and all 291 Port Saidi test memberships come from one video each, and Port Saidi has a single development segment.

\begin{table*}[t]
\centering
\small
\begin{tabular}{@{}llrrrrrr@{}}
\toprule
& & \multicolumn{2}{c}{\textbf{Train}} & \multicolumn{2}{c}{\textbf{Development}} & \multicolumn{2}{c}{\textbf{Test}} \\
\cmidrule(lr){3-4}\cmidrule(lr){5-6}\cmidrule(lr){7-8}
\textbf{Country} & \textbf{Micro-dialect} & Segs. & Videos & Segs. & Videos & Segs. & Videos \\
\midrule
\multirow{4}{*}{Egypt} & Alexandrian & 678 & 27 & 85 & 4 & 85 & 4 \\
 & Cairene & 2,877 & 69 & 311 & 3 & 683 & 1 \\
 & Port Saidi & 832 & 3 & 1 & 1 & 291 & 1 \\
 & Saidi & 1,388 & 45 & 174 & 3 & 173 & 3 \\
\midrule
\multirow{3}{*}{Jordan} & Badawi & 2,319 & 40 & 290 & 6 & 290 & 4 \\
 & Falahi & 2,619 & 61 & 329 & 9 & 328 & 4 \\
 & Madani & 1,528 & 51 & 190 & 8 & 191 & 4 \\
\midrule
\multirow{3}{*}{Lebanon} & B'albaki & 205 & 12 & 22 & 5 & 31 & 1 \\
 & Chamali & 959 & 7 & 58 & 3 & 88 & 3 \\
 & Jnoubi & 1,171 & 16 & 148 & 4 & 145 & 4 \\
\midrule
\multirow{3}{*}{Mauritania} & Northern & 1,254 & 26 & 156 & 2 & 155 & 5 \\
 & Southern & 1,167 & 26 & 156 & 2 & 152 & 5 \\
 & Western & 1,279 & 26 & 160 & 2 & 157 & 5 \\
\midrule
\multirow{3}{*}{Palestine} & Badawi & 692 & 13 & 86 & 4 & 86 & 5 \\
 & Falahi & 1,251 & 26 & 158 & 5 & 155 & 9 \\
 & Madani & 430 & 21 & 53 & 6 & 54 & 6 \\
\midrule
\multirow{5}{*}{Saudi Arabia} & Eastern & 765 & 23 & 96 & 3 & 95 & 3 \\
 & Hijazi & 1,259 & 3 & 83 & 7 & 71 & 2 \\
 & Najdi & 871 & 8 & 109 & 5 & 109 & 6 \\
 & Northern & 411 & 12 & 52 & 2 & 51 & 1 \\
 & Southern & 1,116 & 18 & 165 & 1 & 142 & 4 \\
\bottomrule
\end{tabular}
\caption{Source-disjoint MDI split: micro-dialect label memberships (\textit{Segs.}) and number of distinct source videos per partition. Multi-label segments contribute to each of their labels.}
\label{tab:mdi-source-split}
\end{table*}

\section{Text Preprocessing}
\label{appendix:preprocessing}

We apply the same text preprocessing to both reference transcripts and system predictions before computing WER and CER. Specifically, we: \textit{(a)} remove all types of punctuations; \textit{(b)} remove Arabic diacritics, hamzas, and maddas; and \textit{(c)} convert Eastern Arabic numerals to Western Arabic numerals (e.g., \<٢٩> becomes 29). Unlike pipelines designed for monolingual Arabic, we preserve Latin characters because \cardamom{} contains code-switching, and deleting Latin text would remove valid lexical material from both references and hypotheses.

\section{ASR Adaptation Details}
\label{sec:asr-adaptation-details}

Table~\ref{tab:asr-adaptation-overall} reports aggregate WER/CER for all four systems, zero-shot vs.\ adapted (summarized in Section~\ref{sec:asr-adaptation}).

\paragraph{Evaluation population.}
Table~\ref{tab:micro-split-counts} counts all test-split label memberships. ASR scoring additionally excludes 125 test clips that have no reference transcription, as well as MSA-only clips, leaving 9,152 unique test clips; this is why the per-dialect counts in Tables~\ref{tab:asr-zeroshot-ci} and \ref{tab:asr-adapted-ci} are slightly lower than in Table~\ref{tab:micro-split-counts}. Both ASR tables use exactly the same clips. Aggregate metrics count each clip once, whereas a multi-label clip contributes to every applicable row of the per-dialect tables. Whisper-v3 processes at most 30\,s of audio per input; 64 scored test clips (0.7\%) are longer.

\paragraph{Oracle dialect prompt.} To test a simple form of dialect conditioning, we supplied zero-shot Whisper-v3 with an Arabic initial prompt naming each clip's gold country and micro-dialect label(s), with otherwise identical decoding, normalization, and scoring. On the same 9,152 clips, WER increased from 55.38\% to 61.43\% and CER from 29.40\% to 36.70\%. This oracle setting is a diagnostic rather than a deployable system, and it does not show that dialect conditioning is ineffective in general; dialect conditioning was not tested for the other systems.

\begin{table}[t]
\centering
\small
\setlength{\tabcolsep}{3pt}
\resizebox{0.48\textwidth}{!}{
\begin{tabular}{@{}llll@{}}
\toprule
\textbf{Model} & \textbf{Setting} & \textbf{WER (95\% CI)} & \textbf{CER (95\% CI)} \\
\midrule
\multirow{2}{*}{Whisper-v3} & Zero-shot & 55.38 [53.75, 57.07] & 29.40 [28.07, 30.90] \\
 & Adapted & 46.03 [44.70, 47.42] & 21.53 [20.44, 22.78] \\
\multirow{2}{*}{Seamless-v2} & Zero-shot & 53.13 [52.06, 54.26] & 28.84 [27.78, 30.00] \\
 & Adapted & 52.48 [50.57, 54.58] & 25.38 [23.93, 26.90] \\
\multirow{2}{*}{OmniASR-LLM} & Zero-shot & 43.47 [42.67, 44.29] & 20.67 [19.88, 21.52] \\
 & Adapted & \textbf{35.21} [34.40, 36.09] & \textbf{16.04} [15.27, 16.90] \\
\multirow{2}{*}{OmniASR-CTC} & Zero-shot & 58.17 [57.25, 59.12] & 33.78 [32.56, 35.19] \\
 & Adapted & 40.83 [40.08, 41.61] & 16.08 [15.46, 16.77] \\
\bottomrule
\end{tabular}
}
\caption{Aggregate WER and CER (\%; lower is better) over the 9{,}152 unique test utterances  under zero-shot and adapted settings. Multi-label utterances are counted once. Bracketed values are 95\% percentile bootstrap confidence intervals obtained from 2{,}000 utterance-level resamples of the test set. Adapted models were trained once; therefore, the intervals quantify test-sample uncertainty but not variation across training runs. The best adapted result in each column is shown in \textbf{bold}.}
\label{tab:asr-adaptation-overall}
\end{table}

\subsection{Source-Video-Disjoint OmniASR-LLM Adaptation}
\label{sec:asr-source-disjoint}

Because the utterance-level ASR split can place clips from the same source video in different partitions, we repeat the strongest-model experiment with complete YouTube videos assigned disjointly to train, development, and test. After excluding clips without references, MSA-only clips, clips longer than 30 seconds, and clips shorter than 1,000 samples at 16~kHz, the split contains 21,863/2,467/2,952 utterances (32.34/4.37/4.19 hours) from 309/46/46 source videos, with no video overlap. We fully fine-tune \textsc{OmniASR-LLM-7B-v2} for 1,200 steps using the setup in Table~\ref{tab:model-details}, select without using the test set, and compare zero-shot and adapted decoding on the identical 2,952-utterance test set. Aggregate 95\% intervals use 2,000 bootstrap resamples of the 46 held-out source videos, preserving within-video dependence.

\begin{table}[t]
\centering
\small
\setlength{\tabcolsep}{3pt}
\resizebox{0.48\textwidth}{!}{
\begin{tabular}{@{}lll@{}}
\toprule
\textbf{Setting} & \textbf{WER (95\% CI)} & \textbf{CER (95\% CI)} \\
\midrule
Zero-shot & 43.51 [37.45, 48.67] & 21.84 [16.56, 27.01] \\
Source-disjoint adaptation & \textbf{35.24} [29.95, 40.48] & \textbf{15.83} [12.32, 19.50] \\
\bottomrule
\end{tabular}
}
\caption{Aggregate source-video-disjoint results for \textsc{OmniASR-LLM} (\%; lower is better). Intervals are 95\% percentile intervals from 2,000 source-video-clustered bootstrap resamples.}
\label{tab:asr-source-disjoint-overall}
\end{table}

Table~\ref{tab:asr-source-disjoint-dialects} places the \textsc{OmniASR-LLM} results from Table~\ref{tab:wer-cer-results} beside the source-video-disjoint zero-shot and adapted results. The Table~\ref{tab:wer-cer-results} and source-disjoint columns use different test subsets and therefore provide context rather than a paired comparison. On the source-disjoint test set, adaptation lowers WER for 20 of 21 varieties; Saudi Eastern shows a small 0.78-point WER increase while its CER decreases. Multi-label clips contribute to every applicable row, so row counts do not sum to the number of unique test clips. The smallest subsets, especially Lebanese B'albaki and Saudi Northern, remain descriptive rather than reliable rankings.

\begin{table*}[t]
\centering
\small
\setlength{\tabcolsep}{20pt}
\resizebox{\textwidth}{!}{
\begin{tabular}{@{}llr@{\hspace{15pt}}r@{\hspace{15pt}}r@{\hspace{15pt}}r@{\hspace{18pt}}r@{\hspace{15pt}}r@{\hspace{15pt}}}
\toprule
\textbf{Cnt.} & \textbf{M-Dia} & \multicolumn{2}{c}{\textbf{Table~\ref{tab:wer-cer-results} \textsc{O}-LLM}} & \multicolumn{2}{c}{\textbf{S-disjoint ZS}} & \multicolumn{2}{c}{\textbf{S-disjoint adapted}} \\
\cmidrule(lr){3-4} \cmidrule(lr){5-6} \cmidrule(lr){7-8}
 & & \scriptsize WER $\downarrow$ & \scriptsize CER $\downarrow$ & \scriptsize WER $\downarrow$ & \scriptsize CER $\downarrow$ & \scriptsize WER $\downarrow$ & \scriptsize CER $\downarrow$ \\
\hline
\multirow{4}{*}{\textbf{\EGY}}
 & Alex. & 41.4 & 20.3 & 36.35 & 13.25 & \textbf{22.46} & \textbf{8.45} \\
 & Cairene & 36.1 & 17.9 & 47.45 & 28.20 & \textbf{36.10} & \textbf{16.53} \\
 & Port Said & 46.1 & 25.2 & 54.12 & 26.42 & \textbf{41.15} & \textbf{17.58} \\
 & Saidi & 44.1 & 20.8 & 40.20 & 15.72 & \textbf{30.12} & \textbf{11.14} \\
\hline
\multirow{3}{*}{\textbf{\JOR}}
 & Badawi & 33.6 & 12.2 & 34.67 & 13.59 & \textbf{23.58} & \textbf{8.48} \\
 & Falahi & 33.0 & 13.7 & 29.96 & 10.82 & \textbf{21.57} & \textbf{7.53} \\
 & Madani & \textbf{28.9} & 11.9 & 40.42 & 14.65 & 32.67 & \textbf{11.39} \\
\midrule
\multirow{3}{*}{\textbf{\LEB}}
 & B'albaki & 64.1 & 40.0 & 46.15 & 24.00 & \textbf{41.44} & \textbf{20.79} \\
 & Chamali & \textbf{38.9} & 21.6 & 54.51 & 25.30 & 43.47 & \textbf{19.28} \\
 & Jnoubi & 45.2 & 23.0 & 49.45 & 26.44 & \textbf{42.67} & \textbf{22.77} \\
\midrule
\multirow{3}{*}{\textbf{\MAU}}
 & Northern & 76.0 & 48.5 & 74.67 & 39.29 & \textbf{65.55} & \textbf{30.89} \\
 & Southern & 75.9 & 48.5 & 74.30 & 39.01 & \textbf{64.88} & \textbf{30.41} \\
 & Western & 76.2 & 49.2 & 75.46 & 42.62 & \textbf{68.46} & \textbf{33.74} \\
\midrule
\multirow{3}{*}{\textbf{\PAL}}
 & Badawi & 45.0 & 21.1 & 48.71 & 21.11 & \textbf{34.16} & \textbf{18.48} \\
 & Falahi & 46.4 & 22.9 & 32.64 & 12.50 & \textbf{22.93} & \textbf{9.27} \\
 & Madani & 44.5 & 26.6 & 40.67 & 16.62 & \textbf{28.87} & \textbf{12.27} \\
\midrule
\multirow{5}{*}{\textbf{\KSA}}
 & Eastern & 35.9 & 17.1 & \textbf{17.46} & 7.38 & 18.24 & \textbf{6.66} \\
 & Hijazi & 36.0 & 12.9 & 35.30 & 14.25 & \textbf{25.81} & \textbf{8.89} \\
 & Najdi & 39.4 & 21.1 & 33.41 & 13.27 & \textbf{27.05} & \textbf{9.23} \\
 & Northern & \textbf{67.5} & \textbf{44.9} & 80.26 & 79.63 & 77.90 & 50.45 \\
 & Southern & \textbf{36.5} & \textbf{15.6} & 42.04 & 20.92 & 39.25 & 17.71 \\
\bottomrule
\end{tabular}
}
\caption{Country-grouped \textsc{OmniASR-LLM} WER and CER (\%; lower is better). The Table~\ref{tab:wer-cer-results} columns use the utterance-level split; the source-disjoint columns use 2,952 test clips from 46 held-out videos. Multi-label clips contribute to every applicable row. ZS denotes zero-shot. Bold indicates the lowest value per micro-dialect and metric.}
\label{tab:asr-source-disjoint-dialects}
\end{table*}

\subsection{Checkpoints, Inference, and Fine-Tuning Setup}
Table~\ref{tab:model-details} reports checkpoint names, parameter counts, zero-shot inference settings, and adaptation hyperparameters for all four ASR systems (Section~\ref{sec:asr-benchmark}, \ref{sec:asr-adaptation}). All zero-shot experiments use the original released checkpoints without additional training or in-context examples. Whisper-v3 and Seamless-v2 inference and adaptation use one NVIDIA A100 80GB GPU; \textsc{OmniASR} adaptation uses four A100 80GB GPUs. Adaptation uses only the training split, with validation on the development split; the test split is untouched. No external language model, rescoring, metadata, or test-time adaptation is used.

\begin{table*}[t]
\centering
\small
\setlength{\tabcolsep}{4pt}
\resizebox{\textwidth}{!}{
\begin{tabular}{@{}p{3.2cm}p{5cm}p{4.3cm}p{6cm}@{}}
\toprule
\textbf{System} & \textbf{Checkpoint / Parameters} & \textbf{Zero-shot Inference} & \textbf{Adaptation} \\
\midrule
Whisper-v3 & \texttt{openai/whisper-large-v3}, 1.55B & Greedy Arabic transcription; batch 16 & LoRA on Q/V projections: rank 16, $\alpha{=}32$, dropout 0.05; 3 epochs; batch 8, 4-step accumulation; LR 1e-4 \\
SeamlessM4T-v2 & \texttt{facebook/seamless-m4t-v2-large}, 1.50B S2T parameters & Greedy speech-to-text with Arabic target; batch 8 & LoRA on Q/V projections: rank 16, $\alpha{=}32$, dropout 0.05; 3 epochs; batch 4, 4-step accumulation; LR 1e-4 \\
OmniASR-LLM-v2 & \texttt{omniASR\_LLM\_7B\_v2}, $\approx$7.80B & Autoregressive one-best decoding, \texttt{arb\_Arab}; batch 2 & Full fine-tuning, 1{,}200 steps; LR 1e-5; BF16 FSDP on 4$\times$A100; 8-step accumulation \\
OmniASR-CTC-v2 & \texttt{omniASR\_CTC\_7B\_v2}, $\approx$6.50B & Greedy CTC decoding, \texttt{arb\_Arab}; batch 2 & Full fine-tuning, 1{,}200 steps; LR 1e-5; BF16 FSDP on 4$\times$A100; 8-step accumulation \\
\bottomrule
\end{tabular}
}
\caption{Checkpoints, parameter counts, zero-shot inference settings, and adaptation hyperparameters for the four ASR systems evaluated in Sections~\ref{sec:asr-benchmark} and \ref{sec:asr-adaptation}.}
\label{tab:model-details}
\end{table*}

\subsection{Per-Micro-Dialect Confidence Intervals}
Table~\ref{tab:asr-zeroshot-ci} reports zero-shot WER with 95\% bootstrap confidence intervals (2,000 resamples) per micro-dialect, on the test split described in Section~\ref{sec:asr-adaptation}. Because Jordan and Palestine share the label names Badawi, Falahi, and Madani, and Mauritania and Saudi Arabia both use Northern and Southern, we mark each country explicitly (e.g., \textit{(JOR)} vs.\ \textit{(PAL)}) and distinguish the Mauritanian rows with the \textit{-Mau} suffix; all rows report country-specific micro-dialects. Confidence intervals reflect test-sample uncertainty, not training-seed variance, since each system is evaluated (or, for the adapted setting, fine-tuned) once. The broad ordering across dialects is stable, but narrow rankings on the smallest subsets (B'albaki, Saudi Northern) fall within overlapping intervals and should not be treated as conclusive.

\begin{table*}[t]
\centering
\small
\setlength{\tabcolsep}{3pt}
\begin{tabular}{@{}lrllll@{}}
\toprule
\textbf{Micro-dialect} & \textbf{N} & \textbf{Whisper-v3} & \textbf{Seamless-v2} & \textbf{OmniASR-LLM} & \textbf{OmniASR-CTC} \\
\midrule
Alexandrian (EGY) & 281 & 53.43 [50.10, 56.93] & 46.73 [42.37, 52.02] & 41.33 [38.36, 44.43] & 49.45 [46.34, 52.63] \\
Cairene (EGY) & 1,277 & 50.79 [44.97, 58.72] & 38.22 [36.35, 40.48] & 36.17 [34.78, 37.57] & 44.84 [43.35, 46.25] \\
Port Saidi (EGY) & 374 & 60.70 [54.03, 71.02] & 50.42 [47.14, 53.97] & 45.99 [43.39, 48.85] & 56.26 [53.41, 59.29] \\
Saidi (EGY) & 562 & 58.47 [50.71, 68.09] & 48.22 [44.34, 54.26] & 42.60 [40.50, 44.74] & 50.67 [48.38, 53.10] \\
Badawi (JOR) & 950 & 43.27 [41.25, 45.43] & 45.80 [42.16, 50.30] & 33.69 [31.60, 36.11] & 44.14 [42.38, 45.94] \\
Falahi (JOR) & 1,080 & 40.13 [38.54, 41.75] & 41.97 [40.08, 43.99] & 33.24 [31.73, 34.78] & 44.72 [42.81, 46.65] \\
Madani (JOR) & 632 & 39.58 [34.40, 47.86] & 40.98 [38.36, 43.82] & 28.87 [27.19, 30.62] & 49.38 [46.24, 52.93] \\
B'albaki (LEB) & 81 & 70.32 [58.54, 85.04] & 74.42 [67.67, 80.70] & 58.08 [48.38, 71.99] & 88.16 [82.45, 92.18] \\
Chamali (LEB) & 362 & 60.28 [56.62, 64.72] & 66.23 [62.37, 69.60] & 42.30 [40.19, 44.56] & 81.25 [77.67, 84.36] \\
Jnoubi (LEB) & 478 & 55.28 [49.94, 61.64] & 57.43 [52.24, 63.52] & 45.30 [41.96, 49.33] & 61.45 [57.60, 65.06] \\
Northern-Mau (MAU) & 500 & 86.40 [81.02, 92.96] & 80.82 [78.38, 83.32] & 70.90 [68.90, 72.88] & 85.44 [83.51, 87.19] \\
Southern-Mau (MAU) & 475 & 86.68 [81.38, 93.07] & 81.93 [79.11, 85.00] & 71.14 [69.30, 73.27] & 85.86 [84.08, 87.62] \\
Western-Mau (MAU) & 514 & 86.34 [81.19, 92.47] & 82.40 [79.29, 86.14] & 71.06 [69.09, 72.93] & 85.80 [84.09, 87.49] \\
Badawi (PAL) & 287 & 46.88 [43.46, 50.80] & 44.91 [41.53, 48.72] & 45.94 [41.31, 51.32] & 49.38 [46.21, 52.55] \\
Falahi (PAL) & 510 & 52.40 [48.27, 57.98] & 48.01 [45.46, 50.62] & 46.51 [43.24, 50.21] & 51.99 [49.46, 54.67] \\
Madani (PAL) & 179 & 48.43 [43.83, 53.38] & 47.82 [41.07, 55.82] & 51.51 [38.84, 68.95] & 49.66 [44.78, 54.90] \\
Eastern (KSA) & 312 & 48.39 [42.85, 56.23] & 43.69 [40.82, 47.02] & 35.48 [32.68, 38.59] & 58.06 [54.03, 61.90] \\
Hijazi (KSA) & 471 & 41.72 [40.10, 43.27] & 40.12 [37.79, 42.69] & 36.04 [34.45, 37.66] & 43.87 [41.76, 46.09] \\
Najdi (KSA) & 363 & 53.77 [47.95, 61.70] & 52.29 [49.02, 55.88] & 39.59 [37.30, 42.02] & 61.79 [57.62, 65.67] \\
Northern (KSA) & 170 & 105.73 [82.43, 135.58] & 89.03 [80.28, 100.59] & 68.06 [63.84, 72.52] & 80.31 [76.95, 83.42] \\
Southern (KSA) & 474 & 56.26 [50.76, 62.95] & 58.35 [53.13, 63.11] & 42.17 [38.52, 45.71] & 70.10 [64.40, 74.68] \\
\bottomrule
\end{tabular}
\caption{Zero-shot WER (\%) with 95\% bootstrap confidence intervals per micro-dialect. N is the number of test utterances evaluated for that micro-dialect (single- and multi-label).}
\label{tab:asr-zeroshot-ci}
\end{table*}

Table~\ref{tab:asr-adapted-ci} reports the corresponding WER after adaptation (Section~\ref{sec:asr-adaptation}), using the same micro-dialect grouping and test sets as Table~\ref{tab:asr-zeroshot-ci}.

\begin{table*}[t]
\centering
\small
\setlength{\tabcolsep}{3pt}
\begin{tabular}{@{}lrllll@{}}
\toprule
\textbf{Micro-dialect} & \textbf{N} & \textbf{Whisper-v3} & \textbf{Seamless-v2} & \textbf{OmniASR-LLM} & \textbf{OmniASR-CTC} \\
\midrule
Alexandrian (EGY) & 281 & 45.68 [42.61, 48.97] & 47.02 [40.55, 57.34] & 30.94 [28.11, 34.02] & 41.58 [38.63, 44.74] \\
Cairene (EGY) & 1,277 & 42.39 [38.35, 48.14] & 43.97 [37.59, 52.22] & 29.50 [28.24, 30.86] & 37.49 [36.11, 38.83] \\
Port Saidi (EGY) & 374 & 49.24 [45.85, 53.28] & 48.07 [45.28, 50.94] & 40.14 [36.49, 44.65] & 47.77 [45.18, 50.45] \\
Saidi (EGY) & 562 & 43.38 [41.27, 45.35] & 48.90 [42.25, 60.92] & 33.53 [31.30, 35.88] & 41.19 [39.28, 43.24] \\
Badawi (JOR) & 950 & 33.33 [31.75, 34.87] & 42.24 [39.15, 46.15] & 24.77 [23.51, 26.04] & 31.45 [30.07, 32.87] \\
Falahi (JOR) & 1,080 & 33.08 [31.32, 35.24] & 40.21 [37.11, 43.94] & 24.19 [22.99, 25.40] & 30.61 [29.26, 31.92] \\
Madani (JOR) & 632 & 28.65 [26.91, 30.51] & 34.11 [31.98, 36.42] & 23.33 [21.81, 25.01] & 28.40 [26.69, 30.30] \\
B'albaki (LEB) & 81 & 63.28 [52.45, 77.81] & 78.81 [58.70, 106.86] & 47.81 [43.49, 52.09] & 56.29 [51.62, 60.96] \\
Chamali (LEB) & 362 & 36.05 [33.98, 38.43] & 46.88 [44.50, 49.32] & 31.09 [28.72, 34.04] & 35.00 [32.85, 37.53] \\
Jnoubi (LEB) & 478 & 44.60 [41.81, 47.71] & 63.19 [52.46, 76.32] & 36.38 [34.01, 38.79] & 43.13 [40.99, 45.33] \\
Northern-Mau (MAU) & 500 & 74.64 [71.70, 77.68] & 78.84 [73.93, 84.53] & 59.03 [55.74, 62.11] & 61.04 [58.28, 63.74] \\
Southern-Mau (MAU) & 475 & 75.71 [72.47, 79.35] & 79.09 [74.30, 85.37] & 59.02 [55.76, 62.35] & 61.17 [58.41, 63.80] \\
Western-Mau (MAU) & 514 & 74.88 [72.05, 78.02] & 78.76 [74.14, 84.20] & 58.70 [55.64, 61.90] & 61.17 [58.66, 63.78] \\
Badawi (PAL) & 287 & 38.27 [35.51, 41.39] & 61.16 [43.87, 83.10] & 30.68 [28.10, 33.42] & 36.63 [33.76, 39.69] \\
Falahi (PAL) & 510 & 48.16 [43.54, 54.57] & 50.47 [45.13, 57.85] & 33.89 [31.62, 36.31] & 40.52 [38.37, 42.73] \\
Madani (PAL) & 179 & 39.09 [34.79, 43.79] & 95.70 [51.12, 154.07] & 33.25 [29.41, 37.35] & 39.40 [35.44, 43.75] \\
Eastern (KSA) & 312 & 41.76 [36.80, 49.47] & 44.59 [38.90, 53.68] & 33.64 [30.92, 36.68] & 37.42 [34.55, 40.57] \\
Hijazi (KSA) & 471 & 36.68 [35.11, 38.31] & 36.52 [34.86, 38.05] & 24.77 [23.40, 26.10] & 27.23 [25.72, 28.79] \\
Najdi (KSA) & 363 & 50.44 [42.26, 62.30] & 56.26 [46.16, 70.22] & 34.20 [32.04, 36.40] & 41.01 [38.56, 43.43] \\
Northern (KSA) & 170 & 88.26 [69.01, 116.40] & 93.58 [74.18, 121.38] & 56.32 [52.84, 59.84] & 64.06 [60.93, 67.26] \\
Southern (KSA) & 474 & 49.36 [43.76, 55.66] & 50.41 [45.67, 55.63] & 49.03 [42.58, 54.39] & 52.02 [46.80, 56.19] \\
\bottomrule
\end{tabular}
\caption{WER (\%) after adaptation, with 95\% bootstrap confidence intervals per micro-dialect, matching the test sets and grouping of Table~\ref{tab:asr-zeroshot-ci}.}
\label{tab:asr-adapted-ci}
\end{table*}

\section{Micro-Dialect Identification Details}
\label{sec:dialect-id-appendix}

This section provides task definitions, training details, and full results for the micro-dialect identification (MDI) experiments summarized in Section~\ref{sec:dialect-id}.

\subsection{Tasks and Scoring}
We define three audio-based tasks: \textit{(i)} 6-way country identification over Egypt, Jordan, Lebanon, Mauritania, Palestine, and Saudi Arabia; \textit{(ii)} global micro-dialect identification over all micro-dialect label names pooled across countries; and \textit{(iii)} country-conditioned micro-dialect identification, which predicts among the labels of the gold country. Because a clip may carry several gold labels while the classifier outputs one label, a prediction $\hat{y}_i$ is counted as correct when it belongs to the gold set $G_i$:
\begin{equation}
\mathrm{Acc}_{\mathrm{any}}=\frac{1}{N}\sum_{i=1}^{N}\mathbf{1}[\hat{y}_i\in G_i].
\end{equation}
Matching one gold label does not require recovering the full set. Macro-F1 is the unweighted mean of per-label F1 scores, comparing the one-hot prediction with the multi-hot gold vector.

\subsection{Source-Disjoint Classifier}
The primary evaluation uses the source-disjoint split described in Appendix~\ref{sec:data-splits} (Table~\ref{tab:mdi-source-split}). We train a classifier with a shared \textsc{Whisper-large-v3} audio encoder and three jointly trained heads (country, global micro-dialect, and country-conditioned micro-dialect) for five epochs with learning rate $10^{-5}$ and 5\% linear warm-up, select the checkpoint with the lowest development loss, and evaluate on complete test clips. We repeat training with three seeds while keeping the split, architecture, hyperparameters, and selection rule fixed. Majority baselines predict the most frequent training label (for the conditioned task, the most frequent label within each country).

Table~\ref{tab:dialect-id-full} reports every run. For the first seed, 95\% intervals from 2,000 bootstrap resamples of the 46 held-out videos are 61.10--83.49 (accuracy) and 52.40--77.73 (macro-F1) for country identification, 30.98--68.68 and 21.35--39.18 for global MDI, and 41.30--80.14 and 33.61--51.25 for conditioned MDI. These intervals quantify test-set uncertainty for one checkpoint; the seed standard deviations quantify training variability. The global majority baseline (15.77\%) predicts Falahi, which is the most frequent training label partly because the Jordanian and Palestinian Falahi categories share a prediction label.

\begin{table*}[t]
\centering
\small
\setlength{\tabcolsep}{4pt}
\begin{tabular}{@{}lcccccc@{}}
\toprule
& \multicolumn{2}{c}{\textbf{Country}} & \multicolumn{2}{c}{\textbf{Global M-D}} & \multicolumn{2}{c}{\textbf{Cond.\ M-D}} \\
\cmidrule(lr){2-3} \cmidrule(lr){4-5} \cmidrule(lr){6-7}
\textbf{System} & Acc. & F1 & Acc. & F1 & Acc. & F1 \\
\midrule
Majority & 24.20 & 6.49 & 15.77 & 1.51 & 50.26 & 19.05 \\
Seed 1 & 71.65 & 65.83 & 45.89 & 33.24 & 57.67 & 45.36 \\
Seed 2 & 71.16 & 60.65 & 48.11 & 30.79 & 62.67 & 43.80 \\
Seed 3 & 76.09 & 67.21 & 46.57 & 31.81 & 58.46 & 44.17 \\
Mean $\pm$ SD & 72.97$\pm$2.72 & 64.56$\pm$3.46 & 46.85$\pm$1.14 & 31.95$\pm$1.23 & 59.60$\pm$2.69 & 44.45$\pm$0.82 \\
Mean $\pm$ SD, w/o Mauritania & 74.54$\pm$3.18 & 69.68$\pm$3.93 & 47.00$\pm$1.66 & 34.85$\pm$0.67 & 57.55$\pm$2.86 & 44.08$\pm$1.07 \\
\bottomrule
\end{tabular}
\caption{Source-disjoint MDI accuracy and macro-F1 (\%) on 3,062 test clips from 46 held-out videos (2,900 clips without Mauritania). SD is the sample standard deviation across three training seeds.}
\label{tab:dialect-id-full}
\end{table*}

\subsection{External Checkpoints}
\label{sec:external-baselines}
We evaluate the released ADI-20~\citep{elleuch2025adi}\footnote{\url{https://huggingface.co/Elyadata/ADI-whisper-ADI20}} and VoxLect~\citep{feng2026voxlect}\footnote{\url{https://huggingface.co/tiantiaf/voxlect-arabic-dialect-whisper-large-v3}} checkpoints, without any training on \cardamom{}, on the same 3,062 source-disjoint test clips (Tables~\ref{tab:external-adi} and~\ref{tab:external-adi-country}). Their label spaces are coarser than ours, so neither is a micro-dialect baseline. For ADI-20, we report its native 20-way output and a closed-set evaluation restricted to the six countries in \cardamom{}. For VoxLect, we map Egypt to Egyptian; Jordan, Lebanon, and Palestine to Levantine; Mauritania to Maghrebi; and Saudi Arabia to Peninsular, and report the native five-way output (which includes MSA) and a closed-set evaluation over the four non-MSA regions. The restricted settings assume that the correct label lies in the candidate set. Following VoxLect's documented 15-second input limit, longer clips are center-cropped to 15\,s for that model. Because the training data and label granularity differ from those of our in-domain classifier, these are contextual rather than controlled comparisons. Performance also varies sharply by country, so aggregate scores do not indicate uniform regional coverage.

\begin{table*}[t]
\centering
\small
\begin{tabular}{@{}llcc@{}}
\toprule
\textbf{System} & \textbf{Output space} & \textbf{Accuracy [95\% CI]} & \textbf{Macro-F1 [95\% CI]} \\
\midrule
ADI-20 & Native 20-way & 13.29 [7.04, 19.09] & 17.89 [12.02, 24.90] \\
ADI-20 & Restricted to six countries & 53.14 [42.03, 59.43] & 49.75 [38.70, 54.77] \\
VoxLect & Native 5-way & 35.53 [19.37, 47.08] & 29.73 [18.26, 34.44] \\
VoxLect & Restricted to four non-MSA regions & 38.96 [21.39, 51.80] & 30.99 [19.05, 35.27] \\
\cardamom{} country head (seed 1) & 6-way & 71.65 [61.10, 83.49] & 65.83 [52.40, 77.73] \\
\bottomrule
\end{tabular}
\caption{External checkpoints and our in-domain country classifier on the same source-disjoint test clips. Intervals use 2,000 bootstrap resamples of the 46 held-out videos. Label spaces differ as described in the text.}
\label{tab:external-adi}
\end{table*}

\begin{table}[t]
\centering
\small
\setlength{\tabcolsep}{3pt}
\begin{tabular}{@{}lrrrrr@{}}
\toprule
& & \multicolumn{2}{c}{\textbf{ADI-20}} & \multicolumn{2}{c}{\textbf{VoxLect}} \\
\cmidrule(lr){3-4}\cmidrule(lr){5-6}
\textbf{Country} & \textbf{N} & \textbf{Native} & \textbf{Restr.} & \textbf{Native} & \textbf{Restr.} \\
\midrule
Egypt & 1,196 & 19.15 & 65.13 & 50.08 & 55.85 \\
Jordan & 741 & 2.56 & 45.07 & 4.72 & 5.80 \\
Lebanon & 212 & 4.72 & 33.96 & 7.08 & 8.02 \\
Mauritania & 162 & 6.17 & 51.85 & 6.79 & 8.02 \\
Palestine & 293 & 6.14 & 40.96 & 10.92 & 14.68 \\
Saudi Arabia & 458 & 26.42 & 51.97 & 86.46 & 89.30 \\
\bottomrule
\end{tabular}
\caption{Per-country accuracy (\%) of the external checkpoints in Table~\ref{tab:external-adi}.}
\label{tab:external-adi-country}
\end{table}

\subsection{Mauritania as a Multi-label Case}
\label{sec:mauritania-mdi}
Native speakers describe the three Mauritanian regional varieties as closely related, and annotators frequently assigned more than one of them to the same segment (Sections~\ref{subsec:quant_overlap} and~\ref{sec:agreement-audit}). We present these labels to document the variation reported by native-speaker annotators, not to claim that the varieties are cleanly separable. The source-disjoint test set contains 162 Mauritanian clips from only five videos; 151 carry all three labels and 11 are single-label. Any-match accuracy is therefore highly permissive for this subset, and we do not compare it directly with predominantly single-label subsets. Table~\ref{tab:mauritania-mdi} adds stricter measures for the first seed: example-level Jaccard (one-third credit when one of three gold labels is predicted), exact-set accuracy (which every three-label clip necessarily fails), and macro label F1. Recall is also uneven across labels: 56.13\% (Northern), 9.21\% (Southern), and 35.03\% (Western) in the conditioned task.

\begin{table}[t]
\centering
\small
\setlength{\tabcolsep}{3pt}
\begin{tabular}{@{}lcccc@{}}
\toprule
\textbf{Task} & \textbf{Any-match} & \textbf{Jaccard} & \textbf{Exact set} & \textbf{Macro-F1} \\
\midrule
Global M-D & 54.32 & 19.34 & 1.85 & 29.60 \\
Cond.\ M-D & 96.30 & 34.16 & 3.09 & 46.38 \\
\bottomrule
\end{tabular}
\caption{Mauritanian source-disjoint test clips (162 clips, five videos; seed 1): any-match accuracy versus stricter multi-label measures (\%).}
\label{tab:mauritania-mdi}
\end{table}

\subsection{Utterance-Level Split (Comparison)}
For comparison, we also trained and evaluated MDI systems on the utterance-level ASR split, where an estimated 753 speakers are shared between train and test (Appendix~\ref{appendix:speaker-clustering}); Tables~\ref{tab:dialect-id-utterance} and~\ref{tab:dialect-id-country} report these results as a leakage-sensitive comparison. On that split, we compare zero-shot \textsc{Qwen3-Omni} (prompted with the audio and a closed label set), the same model after multitask LoRA fine-tuning (8.65M trainable parameters), and the Whisper classifier. Global accuracy falls from 85.57\% on this split to 46.85\% on the source-disjoint split, indicating that these figures largely reflect shared speakers, sources, or recording conditions; they should not be read as estimates of generalization. We did not evaluate \textsc{Qwen3-Omni} on the source-disjoint split.

\begin{table}[t]
\centering
\small
\setlength{\tabcolsep}{3pt}
\resizebox{\columnwidth}{!}{%
\begin{tabular}{@{}lcccccc@{}}
\toprule
& \multicolumn{2}{c}{\textbf{Country}} & \multicolumn{2}{c}{\textbf{Global M-D}} & \multicolumn{2}{c}{\textbf{Cond.\ M-D}} \\
\cmidrule(lr){2-3} \cmidrule(lr){4-5} \cmidrule(lr){6-7}
\textbf{System} & Acc. & F1 & Acc. & F1 & Acc. & F1 \\
\midrule
Qwen3-Omni (ZS) & 41.55 & 28.60 & 19.20 & 6.97 & 40.72 & 23.05 \\
Qwen3-Omni (LoRA) & 89.27 & 89.21 & 77.65 & 70.14 & 84.37 & 75.49 \\
Whisper-v3 clf. & \textbf{95.67} & \textbf{95.72} & \textbf{85.57} & \textbf{73.76} & \textbf{88.20} & \textbf{77.11} \\
\bottomrule
\end{tabular}}
\caption{Leakage-sensitive diagnostic: accuracy and macro-F1 (\%) on the utterance-level ASR split, where speakers and source videos are shared across partitions. ZS: zero-shot. These results are not estimates of generalization to unseen sources.}
\label{tab:dialect-id-utterance}
\end{table}

\begin{table}[t]
\centering
\small
\begin{tabular}{@{}lcc@{}}
\toprule
\textbf{Country} & \textbf{Qwen3-Omni (LoRA)} & \textbf{Whisper-v3 clf.} \\
\midrule
Egypt & 77.52 & 82.16 \\
Jordan & 83.42 & 87.54 \\
Lebanon & 94.30 & 94.41 \\
Mauritania & 90.09 & 93.22 \\
Palestine & 74.72 & 79.09 \\
Saudi Arabia & 93.64 & 97.71 \\
\bottomrule
\end{tabular}
\caption{Country-conditioned micro-dialect identification accuracy (\%) by country on the utterance-level split (leakage-sensitive diagnostic).}
\label{tab:dialect-id-country}
\end{table}

\section{Dialectness Result}
\label{app:dialectness}
\begin{figure*}[t]
    \centering
    \includegraphics[width=0.8\linewidth]{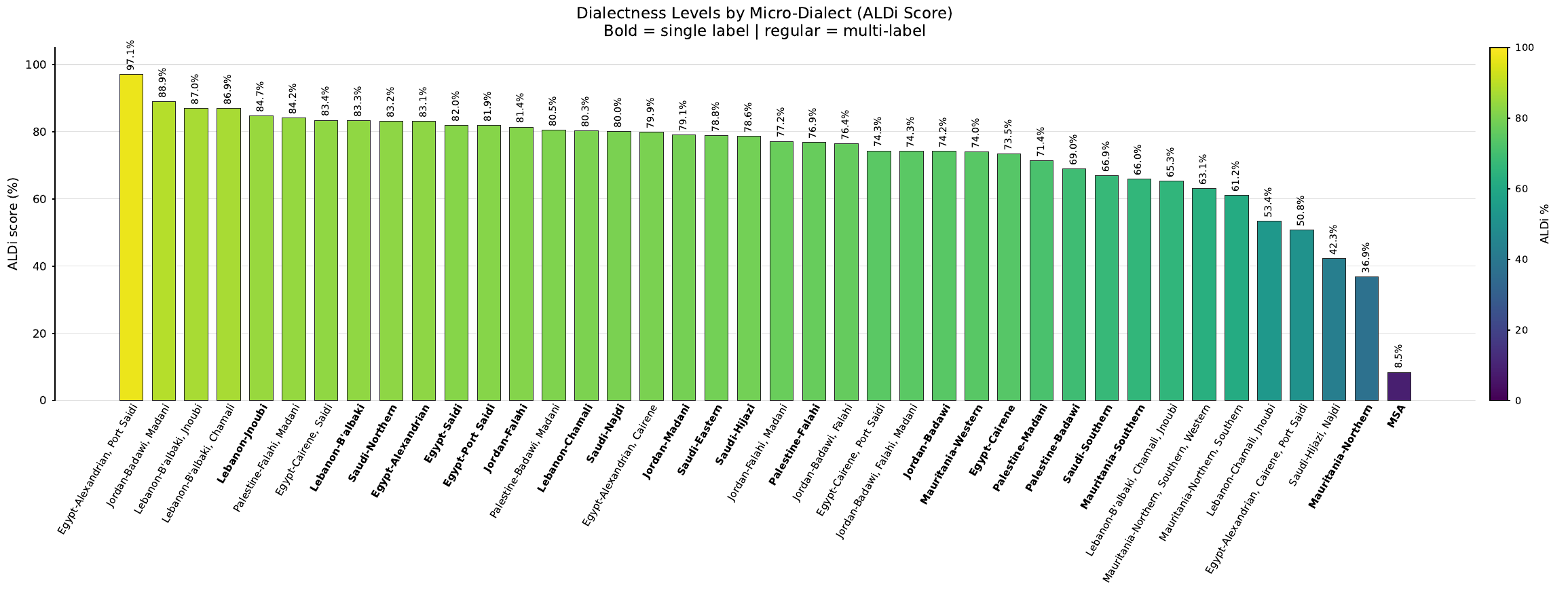}
    \caption{Dialectness scores across micro-dialects; higher indicates more dialectal language.}
    \label{fig:dialectness}
\end{figure*}

To assess how strongly the transcriptions exhibit dialectal characteristics rather than MSA, we score each segment with the \textsc{ALDi} model~\cite{keleg2023aldi} and treat dialectness (ALDi score) and recognition difficulty (ASR error rate) as separate axes. MSA-only segments score lowest (8.5\%) and multi-label intersections highest (up to 97.1\%), confirming a strong dialectal signal in the transcripts; critically, dialectness alone does not explain ASR difficulty, as capital-associated varieties such as Jordan-Madani and Egypt-Cairene remain comparatively easy to recognize despite scoring highly on dialectness. Figure~\ref{fig:dialectness} reports the full per-micro-dialect breakdown.

Figure~\ref{fig:dialectness} shows ALDi dialectness scores~\cite{keleg2023aldi} per micro-dialect group; MSA segments score lowest (8.5\%), multi-label intersections highest (up to 97.1\%), and Mauritanian varieties occupy the lower end (36.9--53.4\%), consistent with their MSA-proximate broadcast sources.

Many of the highest dialectness scores correspond to multi-label dialect--dialect intersections, such as Egypt-Alexandrian--Port Saidi (97.1\%) and Jordan-Badawi--Madani (88.9\%), suggesting that overlapping labels often capture strongly dialectal utterances shared across neighboring or socially related varieties rather than neutral or MSA-like speech. Dialectness and ASR difficulty largely agree for highly dialectal, high-WER varieties such as Egypt-Port Saidi, Egypt-Saidi, and Saudi-Northern (Table~\ref{tab:wer-cer-results}), but they diverge for capital-associated varieties: Jordan-Madani and Egypt-Cairene score highly on dialectness yet remain comparatively easy to recognize, showing that a transcript can be strongly dialectal by ALDi's measure while still being easy for current ASR systems. Better representation of capital-city varieties in pretraining data is one plausible explanation, but we cannot verify it because the training data of these models is not documented at this granularity.

\end{document}